\documentclass[10.0pt,journal,compsoc]{IEEEtran}
\usepackage{booktabs}
\usepackage[svgnames,table]{xcolor}
\usepackage{microtype}
\usepackage{graphicx}
\usepackage{subcaption}
\usepackage{multirow}
\usepackage{booktabs} 
\usepackage{amsmath,amssymb,amsthm,epsfig,dsfont,color,graphicx,epstopdf,hhline,pgfplots}
\usepackage{algorithm,algpseudocode}
\usepackage{tabularx}

\usetikzlibrary{pgfplots.groupplots}
\usetikzlibrary{patterns}
\pgfplotsset{compat=1.11}

\ifCLASSOPTIONcompsoc

  \usepackage[nocompress]{cite}
\else
  \usepackage{cite}
\fi

\usepackage{url}

\newlength\figH
\newlength\figW

\usepackage{pgfplots}

\definecolor{color1}{rgb}{1,0.498039215686275,0.0549019607843137}
\definecolor{color0}{rgb}{0.12156862745098,0.466666666666667,0.705882352941177}
\definecolor{color3}{rgb}{0.83921568627451,0.152941176470588,0.156862745098039}
\definecolor{color2}{rgb}{0.172549019607843,0.627450980392157,0.172549019607843}
\definecolor{color4}{rgb}{0.580392156862745,0.403921568627451,0.741176470588235}
\definecolor{color5}{rgb}{0.549019607843137,0.337254901960784,0.294117647058824}
\definecolor{color6}{rgb}{0.749019607843137,0.137254901960784,0.594117647058824}

\begin{document}
	\title{Node Embedding with Adaptive Similarities\\ for Scalable Learning over Graphs}	
	\author{Dimitris Berberidis,~\IEEEmembership{Student Member,~IEEE,}
		and Georgios B. Giannakis,~\IEEEmembership{Fellow,~IEEE,}
		\IEEEcompsocitemizethanks{\IEEEcompsocthanksitem The authors are with the Department
			of Electrical and Computer Engineering, and Digital Technology Center, University of Minnesota
			MN, 55455.\protect\\
		E-mails: \{bermp001,anikolak,georgios\}@umn.edu
			\IEEEcompsocthanksitem Work was supported by NSF 171141, 1514056 and 1500713.}
		\thanks{Manuscript received XXX.}}
	
	\markboth{IEEE Transactions on Knowledge and Data Engineering (Submitted \today)}%
	{Berberidis \MakeLowercase{\textit{et al.}}: Adaptive-similarity node embedding for scalable learning over graphs}

\IEEEtitleabstractindextext{%
\begin{abstract}
Node embedding is the task of extracting informative and descriptive features over the nodes of a graph. The importance of node embedding for graph analytics as well as learning tasks, such as node classification, link prediction, and community detection, has led to a growing interest and a number of recent advances. Nonetheless, node embedding faces several major challenges. Practical embedding methods have to deal with real-world graphs that arise from different domains, with inherently diverse underlying processes as well as similarity structures and metrics. On the other hand, similar to principal component analysis in feature vector spaces, node embedding is an inherently \emph{unsupervised} task. Lacking metadata for validation, practical schemes motivate standardization and limited use of tunable hyperparameters. Finally, node embedding methods must be scalable in order to cope with large-scale real-world graphs of networks with ever-increasing size. The present work puts forth an adaptive node embedding framework that adjusts the embedding process to a given underlying graph, in a fully unsupervised manner. This is achieved by leveraging the notion of a tunable node similarity matrix that assigns weights on multihop paths. The design of multihop similarities ensures that the resultant embeddings also inherit interpretable spectral properties. The proposed model is thoroughly investigated, interpreted, and numerically evaluated using stochastic block models. Moreover, an unsupervised algorithm is developed for training the model parameters effieciently. Extensive node classification, link prediction, and clustering experiments are carried out on many real-world graphs from various domains, along with comparisons with state-of-the-art scalable and unsupervised node embedding alternatives. The proposed method enjoys superior performance in many cases, while also yielding interpretable information on the underlying graph structure.   

\end{abstract}
	}

\maketitle	

\IEEEdisplaynontitleabstractindextext

\IEEEpeerreviewmaketitle

\IEEEraisesectionheading{\section{Introduction}}
\label{sec:intro}

\IEEEPARstart{U}supervised node embedding is an exciting field, in which  a significant amount of progress has been made in recent years \cite{survey_tkde}. The task consists of mapping each node of a graph to a vector in a low-dimensional Euclidean space. The main goal is to \emph{extract features} that can be utilized downstream in order to perform a variety of unsupervised or (semi-)supervised learning tasks, such as node classification, link prediction, or clustering \cite{survey2}. Ideally, it is desired for the embedded nodal vectors to convey at least as much information as the original graph. Nevertheless, an appropriate embedding can boost the performance of certain learning tasks because they allow one to work with the more ``friendly'' and intuitive Euclidean representation, and deploy mature and widely implemented feature-based algorithms such as (kernel) support vector machines (SVMs), logistic regression, and K-means.   

Early embedding works mostly focused on a structure-preserving dimensionality reduction of feature vectors (instead of nodes); see for instance \cite{mds, isomap, lle, lpp, partially}. In this context, graphs are constructed from pairwise feature vector relations and are treated as representations of the manifold that data lie on; embedded vectors are then generated so that they preserve the corresponding pair-wise proximities on the manifold. More recently, nodal vector embedding of a graph has attracted considerable attention in different fields, and is often posed as the factorization of a properly defined node similarity matrix
\cite{large_scale_fact,text_fact,mf,hope,netmf,grarep,factor2,factor3}. Efforts in this direction mostly focus on designing meaningful similarity metrics to factorize. While some methods (e.g. \cite{large_scale_fact,hope}) maintain scalability by factorizing similarity matrices in an implicit manner (without explicitly forming them), others such as \cite{netmf,grarep} form and/or factorize dense similarity matrices that scale poorly to large graphs. Another line of work opts to gradually fit pairs of embedded vectors to existing edges using stochastic optimization tools \cite{LINE,PTE}. Such approaches are naturally scalable and entail a high degree of locality. Recently, stochastic edge-fitting has been generalized to implicitly accommodate long-range node similarities \cite{VERSE}. Meanwhile, other works have approached node embeddings using random-walk-based tools and concepts originating from natural language processing \cite{deepwalk,node2vec,attention}; see also related works on embedding of knowledge graphs \cite{know1,know2,know3}. Methods that rely on graph convolutional neural networks and autoencoders have also been proposed for node embedding \cite{neural1,neural2,neural3}. Moreover, a gamut of related embedding tasks are gaining traction, such as embedding based on structural roles of nodes \cite{struct2vec,graphwave}, supervised embeddings for classification \cite{planetoid}, and inductive embedding methods that utilize multiple graphs \cite{inductive} 			 
\\ 

We identify the following \emph{challenges} that need to be addressed in order to design embedding methods that are applicable in practice:
\begin{itemize}
	\item \textbf{Diversity}. Since graphs that arise from different domains are generally characterized by a diverse set of properties, there may not be a ``one-size-fits-all'' node embedding approach. 	
	
	\item \textbf{No supervision}. At the same time, node embedding may need to be performed in a \emph{fully unsupervised} manner, that is, without extra information (node attributes, labels, or groundtruth communities) to guide the parameter tuning process with cross-validation.  
	
	\item \textbf{Scalability}. While some real-world networks are of moderate size, others may contain massive numbers of nodes and edges. Specifically, graphs encountered with social networks, transportation networks, knowledge graphs and others, typically scale to millions of nodes and tens of millions of edges. Thus, strict computational constraints must be accounted by the design of node embedding methods.
\end{itemize}
In response to these challenges, we propose a scalable node embedding framework that is based on factorizing an adaptive node similarity matrix. The first challenge is addressed by utilizing a large family of node similarity metrics, parametrized by placing different weights on node proximities of different orders; see also our precursor work \cite{adadif}. Experiments indicate that the proposed model for similarity metrics is expressive enough to describe real-world graphs from diverse domains and with different structures. To address the second challenge (lack of supervision), we put forth a self-supervised parameter learning scheme based on predicting randomly removed edges. Finally, we accommodate scalability by constraining the parametrization of similarity matrices such that the proximity order parameters carry over to the embedded vectors in a smooth manner. This allows for learning proximity order parameters directly on the feature vectors. Consequently, dense similarity matrices do not need to be explicitly formed and factorized, thus endowing the proposed method with the desired level of scalability.

The rest of the paper is organized as follows. Section 2 introduces the problem and the proposed similarity model. Section 3 presents a numerical study on model properties, while Section 4 deals with learning the model parameters in an unsupervised manner. Finally, Section 5 discusses related methods, and Section 6 contains experiments on real graphs, comparisons with competing alternatives, and interpretation of the results. While notation is defined wherever it is introduced, we also summarize the most important symbols that appear throughout the paper in Table 1.   \\
 
\begin{table}[t]\caption{Important Notation}
	\centering 
	\begin{tabular}{r c p{6cm} }
		\toprule
		$\mathcal{V}$ & $\triangleq$ & Set of nodes\\
		$\mathcal{E}$ & $\triangleq$ & Set of edges\\		
		$\mathbf{A}$ & $\triangleq$ & $N\times N$ adjacency matrix\\	
		$\mathbf{D}$ & $\triangleq$ & $\mathrm{diag}(\mathbf{1}^T\mathbf{A})$ diagonal degree matrix \\
		$\mathbf{E}$ & $\triangleq$ & $N\times d$ matrix of embeddings\\
		$\mathbf{e}_i$ & $\triangleq$ & Embedding vector of node $v_i$ \\
		$s_\mathcal{G}(\cdot,\cdot)$ & $\triangleq$ & Node -- to -- node similarity  \\								
		$s_k (\cdot,\cdot)$	& $\triangleq$ & $k-$hop node -- to -- node similarity  \\					
		$s_\mathcal{E}(\cdot,\cdot)$ & $\triangleq$ & Embedding -- to -- embedding similarity \\						
		$\ell(\cdot,\cdot)$ & $\triangleq$ & Distance (loss) between similarities \\									
		$\mathbf{S}_{\mathcal{G}}$& $\triangleq$ & Final node similarity matrix \\					
		$\mathbf{S}$ & $\triangleq$ & Basic sparse (single-hop) and symmetric node similarity matrix \\										
		$\theta_k$& $\triangleq$ & Coefficient of $k$-hop paths \\			
		$\boldsymbol{\theta}$& $\triangleq$ & $[\theta_1, \ldots, \theta_K]^T$  vector of coefficients \\								
		$\mathcal{S}^K$ & $\triangleq$ & $K-$dimensional probability simplex \\					
		$\mathcal{S}^+$ & $\triangleq$ & Set of sampled positive edges\\				
		$\mathcal{S}^-$ & $\triangleq$ & Set of all sampled negative edges\\						
		$\mathcal{S}$ & $\triangleq$ & $\mathcal{S}^+\cup\mathcal{S}^-$ all sampled edges\\				
		$N_s$ & $\triangleq$ & Number of sampled edges\\
		$\boldsymbol{\theta}^\ast_\mathcal{S}$ & $\triangleq$ & Optimal coefficients that fit sample $\mathcal{S}$\\
				$T_s$ & $\triangleq$ & Number of different edge samples\\
		\bottomrule
	\end{tabular}
	\label{tab:TableOfNotationForMyResearch}
\end{table}

\section{Problem Statement and Modeling}\label{sec:problem}

Given an undirected graph $\mathcal{G}:=\{ \mathcal{V},\mathcal{E}\}$,  where $\mathcal{V}$ is the set of $N$ nodes, and $\mathcal{E} \subseteq \mathcal{V}\times \mathcal{V}$ is the set of edges, the task of node embedding boils down to determining  $f(\cdot):\mathcal{V}\rightarrow\mathbb{R}^d$, where $d\ll N$. In other works, a function is sought to map every node of $\mathcal{G}$ to a vector in the $d-$dimensional Euclidean space. Typically, the embedding is low dimensional with $d$ much smaller than the number of nodes. Given $f(\cdot)$, the low-dimensional vector representation of each node $v_i$ is
\begin{equation*}
\mathbf{e}_i  =f(v_i) ~~\forall v_i \in \mathcal{V}\;.
\end{equation*}
Since the number of nodes is finite, instead of finding a general $f(\cdot)$ (induction), one may pose the embedding task in its most general form as the following minimization problem over the embedded vectors
\begin{equation} \label{most_general}
\{\mathbf{e}^\ast_i\}_{i=1}^N = \arg\underset{\{\mathbf{e}_i\}_{i=1}^N}{\min} \sum_{v_i,v_j\in\mathcal{V}} \ell\left(s_{\mathcal{G}}(v_i,v_j), s_{\mathcal{E}}(\mathbf{e}_i,\mathbf{e}_j)\right)
\end{equation}
where $\ell(\cdot,\cdot): \mathbb{R}\times\mathbb{R}\rightarrow\mathbb{R}$ is a loss function; $s_{\mathcal{G}}(\cdot,\cdot): \mathcal{V}\times\mathcal{V}\rightarrow\mathbb{R}$ is a similarity metric over pairs of graph \emph{nodes}; and $s_{\mathcal{E}}(\cdot,\cdot): \mathbb{R}^d\times\mathbb{R}^d\rightarrow\mathbb{R}$ a similarity metric over pairs of \emph{vectors} in the $d-$dimensional Euclidean space. 

In par with \eqref{most_general}, node embedding can be viewed as the design of nodal vectors $\{\mathbf{e}_i\}_{i=1}^N$ that successfully ``encode'' a certain notion of pairwise similarities among graph nodes. 

\subsection{Embedding as matrix factorization}

Starting from the generalized framework in \eqref{most_general}, one may arrive at concrete approaches by specifying choices of $s_{\mathcal{G}}(\cdot,\cdot)$, $s_{\mathcal{E}}(\cdot,\cdot)$, and $\ell(\cdot,\cdot)$. To start, suppose that the node similarity metric is symmetric; that is,  $s_{\mathcal{G}}(v_i,v_j)=s_{\mathcal{G}}(v_j,v_i)~\forall v_i,v_j\in\mathcal{V}$. Furthermore, let the loss function be quadratic
\begin{equation*}
\ell(x,x^\prime) = \left( x - x^\prime \right)^2
\end{equation*}	
and the nodal vector similarity be the inner product
\begin{equation*}
s_{\mathcal{E}}(\mathbf{e}_i,\mathbf{e}_j)  = \mathbf{e}_i^\top\mathbf{e}_j. 
\end{equation*}	
Using these specifications, \eqref{most_general} reduces to the following symmetric matrix factorization problem
	\begin{equation} \label{general}
	\mathbf{E}^\ast = \arg\underset{\mathbf{E}\in\mathbb{R}^{N\times d}}{\min} 
	\|\mathbf{S}_{\mathcal{G}}-\mathbf{E}\mathbf{E}^\top\|_F^2
	\end{equation}
where $\mathbf{S}_{\mathcal{G}}\in\mathbb{R}^{N\times N}$ is the symmetric similarity matrix with  $\left[\mathbf{S}_{\mathcal{G}}\right]_{i,j}=\left[\mathbf{S}_{\mathcal{G}}\right]_{j,i}=s_{\mathcal{G}}(v_i,v_j)$, and matrix $\mathbf{E} := \left[\mathbf{e}_1 \ldots \mathbf{e}_N\right]^\top$ concatenates all node embeddings as rows.
%
A well-known analytical solution to \eqref{general} relies on the singular value decomposition (SVD) of the similarity matrix, that is $\mathbf{S}_{\mathcal{G}} = \mathbf{U} \boldsymbol{\Sigma} \mathbf{V}^T$, where $\mathbf{U}$ and $\mathbf{V}$ are the $N\times N$ unitary matrices formed by the left and right singular vectors, and $\boldsymbol{\Sigma}$ is diagonal with non-negative singular values sorted in decreasing order; in our case, $\mathbf{U}=\mathbf{V}$ since $\mathbf{S}_{\mathcal{G}}$ is symmetric. Given the SVD of $\mathbf{S}_{\mathcal{G}}$, the low-rank ($d\ll N$) solver in \eqref{general} is   
$\mathbf{E}^\ast = \mathbf{U}_d \boldsymbol{\Sigma}_d^{1/2}$, where  $\boldsymbol{\Sigma}_d$ contains the $d$ largest singular values, and $\mathbf{U}_d$ the corresponding singular vectors~\cite{svd}. Matrices $\mathbf{U}_d$ and $\boldsymbol{\Sigma}_d$ can be obtained directly using the reduced-complexity scheme known as \emph{truncated} SVD. 

If in addition $\mathbf{S}_{\mathcal{G}}$ is \emph{sparse}, \eqref{general} can be solved even more efficiently, with complexity that scales with the number of edges. One such example with sparse similarities is when  $\mathbf{S}_{\mathcal{G}}=\mathbf{A}$, where $\mathbf{A}$ is the graph 
adjacency matrix. Embeddings generally gain scalability by avoiding the explicit construction of a \emph{dense} $\mathbf{S}_{\mathcal{G}}$. In fact, simply storing $\mathbf{S}_{\mathcal{G}}$ in the working memory becomes prohibitive even for graphs of moderate sizes (say $N > 10^5$). 

In the ensuing section, we will design a family of dense similarity matrices that (among other properties) can be decomposed implicitly, at the cost of input sparsity.  

\subsection{Multihop graph node similarities}
Having reduced the node embedding problem to the one in \eqref{general}, it remains to specify the graph similarity metric that gives rise to $\mathbf{S}_{\mathcal{G}}$. Towards this end, and in order to maintain expressibility, we will design a parametric model for $\mathbf{S}_{\mathcal{G}}$, with each pairwise node similarity metric expressed as
\begin{equation} \label{multipath_similarity}
s_{\mathcal{G}}(v_i,v_j;\boldsymbol{\theta}) = \sum_{k=1}^K \theta_k s_k(v_i,v_j), ~~~\mathrm{s.t.}~~~ \boldsymbol{\theta} \in \mathcal{S}^K
\end{equation} 
where $\mathcal{S}^K:=\{ \boldsymbol{\theta} \in \mathbb{R}^K : \boldsymbol{\theta}\geq \mathbf{0}, \boldsymbol{\theta}^\top \mathbf{1}=1 \}$ is the $K$-dimensional probability simplex, and $s_k (v_i,v_j)$ is a similarity metric that depends on all $k$-hop paths of possibly repeated nodes that start from $v_i$ and end at $v_j$ (or vice-versa). Thus, $s_{\mathcal{G}}(\cdot,\cdot;\boldsymbol{\theta})$ contains all $k$-hop interactions between two nodes, each weighted by a non-negative importance score $\theta_k$ with $k=1,\ldots,K$. 

Let $\mathbf{S}$ be any similarity matrix that is characterized by the same sparsity pattern as the adjacency matrix, that is 
\begin{align} \label{sparse_pattern}		
S_{i,j}=\left\{ \begin{array}{cc}
s_{i,j},~&~ (i,j) \in \mathcal{E} \\
0,~&~(i,j) \notin \mathcal{E}
\end{array} \right., 
\end{align}
where $\{s_{i,j}\}$s denote the generic non-negative values of entries that correspond to edges of $\mathcal{G}$. Maintaining the same sparsity pattern as $\mathbf{A}$ allows for the $(i,j)$ entry of $\mathbf{S}^k$ to be interpreted as a measure of influence between $v_i$ and $v_j$ that depends on all $k$-hop paths that connect them; that is, $\left[\mathbf{S}^k\right]_{i,j}=s_k(v_i,v_j)$. For instance, selecting $\mathbf{S}=\mathbf{A}$ is equivalent to using the $k$-step similarity 
$s_k(v_i,v_j) = | \{k-\mathrm{length~paths~connecting}~ v_i~ \mathrm{to}~ v_j \} |$ \cite{arope}. Likewise, if $\mathbf{S}=\mathbf{A}\mathbf{D}^{-1}$ where $\mathbf{D} = \mathrm{diag}(\mathbf{1}^T\mathbf{A})$, then  $s_k(v_i,v_j)$ can be interpreted as the probability that a random walk starting from $v_j$ lands on $v_i$ after exactly $k$ steps, e.g., \cite{grarep}. Thus, for a properly selected $\mathbf{S}$ with entries as in \eqref{sparse_pattern}, tunable multihop similarity metrics in \eqref{multipath_similarity} can be collected as entries of the power series matrix
\begin{equation} \label{our_similarity}
\mathbf{S}_{\mathcal{G}}(\boldsymbol{\theta}) = \sum_{k=1}^K \theta_k \mathbf{S}^k, ~~~\mathrm{s.t.}~~~ \boldsymbol{\theta} \in \mathcal{S}^K\;.
\end{equation} 
Upon substituting \eqref{our_similarity} into \eqref{general} yields the tunable embeddings $\mathbf{E}^\ast(\boldsymbol{\theta})$ that depend on the choice of parameters $\boldsymbol{\theta}$. From the eigen-decomposition $\mathbf{S} = \mathbf{U}\boldsymbol{\Sigma} \mathbf{U}^\top $, and given that $\mathbf{U}^\top\mathbf{U}=\mathbf{I}$, we readily arrive at
\begin{equation} \label{power}
\mathbf{S}^k = \mathbf{U}\boldsymbol{\Sigma}^k \mathbf{U}^\top
\end{equation} 
and after plugging \eqref{power} into \eqref{our_similarity}, we obtain
\begin{equation} \label{nice_form}
\mathbf{S}_{\mathcal{G}}(\boldsymbol{\theta}) = \mathbf{U} \left( \sum_{k=1}^K \theta_k \boldsymbol{\Sigma}^k \right) \mathbf{U}^\top , ~~~\mathrm{s.t.}~~~ \boldsymbol{\theta} \in \mathcal{S}^K\:.
\end{equation} 
Furthermore, the truncated singular pairs  of $\mathbf{S}_{\mathcal{G}}(\boldsymbol{\theta})$ conveniently follow from those of $\mathbf{S}$, and they have to be computed once. Specifically, the truncated singular vectors and singular values are $\mathbf{U}_d(\boldsymbol{\theta})=\mathbf{U}_d$ and $\boldsymbol{\Sigma}_d(\boldsymbol{\theta})=\sum_{k=1}^K \theta_k \boldsymbol{\Sigma}_d^k$, respectively. Thus, if $\mathbf{S}\in \mathrm{Sym}_N$ the solution to \eqref{general} with $\mathbf{S}_{\mathcal{G}}$ parametrized by $\boldsymbol{\theta}$ is simply given as
\begin{equation}\label{solution}
\mathbf{E}^\ast(\boldsymbol{\theta}) = \mathbf{U}_d \sqrt{\boldsymbol{\Sigma}_d(\boldsymbol{\theta})}\;.
\end{equation}
Note that this holds only for non-negative parameters $\theta_k\geq 0~\forall~k$. If $\theta_k<0$ for at least one $k\in \{1,\ldots,K\}$, then the diagonal entries of $\boldsymbol{\Sigma}_d(\boldsymbol{\theta})$ cannot be guaranteed to be non-negative and sorted in decreasing order, which would cause $\left(\mathbf{U}_d(\boldsymbol{\theta}), \boldsymbol{\Sigma}_d(\boldsymbol{\theta})\right)$ to \emph{not} be a valid SVD pair. 

Having narrowed down $\mathbf{S}_{\mathcal{G}}$ to belong to the parametrized family in \eqref{our_similarity}, we proceed to select an appropriate sparsity-preserving $\mathbf{S}$ in order to obtain a solid model.

\subsection{Spectral multihop embeddings }

While any symmetric $\mathbf{S}$ that obeys \eqref{sparse_pattern} can be used for constructing multihop similarities (cf. \eqref{our_similarity}), judicious designs of 
$\mathbf{S}$ can effect certain desirable properties. Bearing this in mind, consider the following identity
\begin{equation}\label{equiv}
\mathbf{S}\in \mathcal{P}^+_N ~\iff~ \mathbf{S} = \mathbf{U}\boldsymbol{\Sigma} \mathbf{U}^\top = \mathbf{U}\boldsymbol{\Lambda} \mathbf{U}^\top
\end{equation}
where $\mathcal{P}^+_N$ denotes the space of $N\times N$ symmetric positive definite (SPD) matrices, and $\boldsymbol{\Lambda}$ is the diagonal matrix that contains the eigenvalues of $\mathbf{S}$ sorted in decreasing order. For SPD matrices as in \eqref{equiv}, the SVD is identical to the eigenvalue decomposition (EVD). Thus, if $\mathbf{S}\in \mathcal{P}^+_N$, the solution to \eqref{general} is also given as (cf. \eqref{solution}) 
 \begin{equation}\label{solution2}
\mathbf{E}^\ast(\boldsymbol{\theta}) = \mathbf{U}_d \sqrt{\boldsymbol{\Lambda}_d(\boldsymbol{\theta})}
\end{equation}
where $\mathbf{U}_d$ are also the first $d$ \emph{eigenvectors} of $\mathbf{S}$, and $\boldsymbol{\Lambda}_d(\boldsymbol{\theta})=\sum_{k=1}^K \theta_k \boldsymbol{\Lambda}_d^k$ is the $K$th order polynomial of its eigenvalues defined by $\boldsymbol{\theta}$.

Consider now specifying $\mathbf{S}$ as
\begin{equation} \label{my_matrix}
\mathbf{S} = \frac{1}{2}\left( \mathbf{I} + \mathbf{D}^{-1/2} \mathbf{A} \mathbf{D}^{-1/2} \right).
\end{equation} 
Recalling that $\lambda_i\left( \mathbf{D}^{-1/2} \mathbf{A} \mathbf{D}^{-1/2} \right)\in [-1,1]~\forall~i$, and after using the identity shifting and scaling, we deduce that $\lambda_i(\mathbf{S})\in [0,1]~\forall~i$; hence, matrix $\mathbf{S}$ in \eqref{my_matrix} is SPD. It can also be readily verified that the first $d$ eigenvectors of $\mathbf{S}$ coincide with the eigenvectors corresponding to the $d$ smallest eigenvalues of the symmetric normalized Laplacian matrix
\begin{equation}\label{lsym}
\mathbf{L}_{\mathrm{sym}}:= \mathbf{I} - \mathbf{D}^{-1/2} \mathbf{A} \mathbf{D}^{-1/2}.
\end{equation}
These smallest eigenvalues are known to contain useful information on cluster structures of different resolution levels, a key property that has been successfully employed by spectral clustering \cite{spectral}. Intuitively, assigning weight $\boldsymbol{\theta}_k$ to $k$-hop paths in the node similarity of \eqref{our_similarity}, is equivalent to shrinking the $d$-dimensional spectral node embeddings (rows of $\mathbf{U}_d$) coordinates according to $\boldsymbol{\Lambda}_d(\boldsymbol{\theta})$. Interestingly, assigning large weights to longer paths ($K\gg 1$) is equivalent to fast shrinking the coordinates that correspond to small eigenvalues and capture the fine-grained structures and local relations, what leads to a coarse, high-level cluster description of the graph.  

\subsection{Relation to random walks}
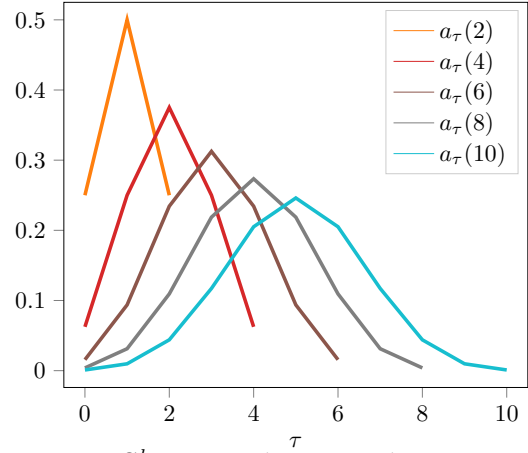
\begin{figure}[t!]  
	\hspace{0.25in}
\begin{tikzpicture}[scale=0.9]

\definecolor{color0}{rgb}{0.12156862745098,0.466666666666667,0.705882352941177}
\definecolor{color1}{rgb}{1,0.498039215686275,0.0549019607843137}
\definecolor{color2}{rgb}{0.172549019607843,0.627450980392157,0.172549019607843}
\definecolor{color3}{rgb}{0.83921568627451,0.152941176470588,0.156862745098039}
\definecolor{color7}{rgb}{0.737254901960784,0.741176470588235,0.133333333333333}
\definecolor{color8}{rgb}{0.0901960784313725,0.745098039215686,0.811764705882353}
\definecolor{color5}{rgb}{0.549019607843137,0.337254901960784,0.294117647058824}
\definecolor{color4}{rgb}{0.580392156862745,0.403921568627451,0.741176470588235}
\definecolor{color6}{rgb}{0.890196078431372,0.466666666666667,0.76078431372549}

\begin{axis}[
legend cell align={left},
legend entries={$a_\tau(2)$,$a_\tau(4)$,$a_\tau(6)$, $a_\tau(8)$,$a_\tau(10)$},
legend style={draw=white!80.0!black},
tick align=outside,
tick pos=left,
x grid style={lightgray!92.02614379084967!black},
xlabel={$\tau$},
xmin=-0.5, xmax=10.5,
y grid style={lightgray!92.02614379084967!black},
ymin=-0.023974609375, ymax=0.524951171875,
]

\addlegendimage{no markers, color1}
\addlegendimage{no markers, color3}
\addlegendimage{no markers, color5}
\addlegendimage{no markers, gray!99.6078431372549!black}
\addlegendimage{no markers, color8}

\addplot [line width = 1.5pt, color1]
table [row sep=\\]{%
0	0.25 \\
1	0.5 \\
2	0.25 \\
};

\addplot [line width = 1.5pt, color3]
table [row sep=\\]{%
0	0.0625 \\
1	0.25 \\
2	0.375 \\
3	0.25 \\
4	0.0625 \\
};

\addplot [line width = 1.5pt, color5]
table [row sep=\\]{%
0	0.015625 \\
1	0.09375 \\
2	0.234375 \\
3	0.3125 \\
4	0.234375 \\
5	0.09375 \\
6	0.015625 \\
};

\addplot [line width = 1.5pt, white!49.80392156862745!black]
table [row sep=\\]{%
0	0.00390625 \\
1	0.03125 \\
2	0.109375 \\
3	0.21875 \\
4	0.2734375 \\
5	0.21875 \\
6	0.109375 \\
7	0.03125 \\
8	0.00390625 \\
};

\addplot [line width = 1.5pt, color8]
table [row sep=\\]{%
0	0.0009765625 \\
1	0.009765625 \\
2	0.0439453125 \\
3	0.1171875 \\
4	0.205078125 \\
5	0.24609375 \\
6	0.205078125 \\
7	0.1171875 \\
8	0.0439453125 \\
9	0.009765625 \\
10	0.0009765625 \\
};

\end{axis}

\end{tikzpicture}
	\vspace{-0.15in}
	\caption{Matrix $\mathbf{S}^k$ is equivalent to applying ``wavelet''-type weights $\alpha_\tau(k)$ over walks with hops $\leq k$.}
	\label{fig:Sk}
\end{figure}

Apart from the spectral embedding interpretation discussed in the last subsection, using powers of $\eqref{my_matrix}$ to capture multihop similarities also admits an interesting random walk interpretation. We begin by expressing the $k$th power of $\mathbf{S}$ as 
\begin{align}\nonumber
\mathbf{S}^k &= \frac{1}{2^k}\left( \mathbf{I} + \mathbf{D}^{-1/2} \mathbf{A} \mathbf{D}^{-1/2} \right)^k \\ \label{expand1}
 &=\sum_{\tau=0}^k \alpha_{\tau}(k) \left(\mathbf{D}^{-1/2} \mathbf{A} \mathbf{D}^{-1/2}\right)^\tau
\end{align}
where the sequence
\begin{equation} \label{binom}
\alpha_{\tau}(k) :=\left\{ \begin{array}{cc}
\frac{1}{2^k}\binom{k}{\tau},~&~ 0 \leq \tau \leq k \\
0,~&~ \mathrm{else}
\end{array} \right.  
\end{equation}
can be interpreted as nonzero weights that $\mathbf{S}^k$ assigns to all paths with the number of hops \emph{up to}  $k$ (see Fig. \ref{fig:Sk}). 

Using \eqref{expand1} and \eqref{binom}, the multihop similarity in \eqref{our_similarity} becomes
\begin{align} \nonumber
\mathbf{S}_{\mathcal{G}}(\boldsymbol{\theta}) &= \sum_{\tau=0}^K c_\tau(\boldsymbol{\theta}) \left(\mathbf{D}^{-1/2} \mathbf{A} \mathbf{D}^{-1/2}\right)^\tau \\ \label{expand2}  
& = \mathbf{D}^{-1/2} \left( \sum_{\tau=0}^K c_\tau(\boldsymbol{\theta}) \mathbf{P}^\tau \right) \mathbf{D}^{1/2}
\end{align} 
where 
\begin{equation}\label{c}
c_\tau(\boldsymbol{\theta}) := \sum_{k=1}^K \theta_k \alpha_{\tau}(k)
\end{equation}
and $\mathbf{P} = \mathbf{AD}^{-1}$ is the probability transition matrix of a simple random walk defined over $\mathcal{G}$; that is, $P_{i,j}$ is the probabiity that a random walker positioned on node (state) $j$ transitions to node $i$ in one step. Thus, the $k$-hop similarity function defined in \eqref{multipath_similarity} is expressed as 
\begin{equation}
 s_{\mathcal{G}}(v_i,v_j,\boldsymbol{\theta}) = \sqrt{\frac{d_j}{d_i}} \sum_{\tau=0}^K c_\tau(\boldsymbol{\theta}) \Pr \{ X_\tau = v_i | X_0 = v_j\}
\end{equation} 
where $\Pr \{ X_\tau = v_i | X_0 = v_j\}:=\left[\mathbf{P}^\tau\right]_{ij}$ is the probability that a random walk starting from $v_j$ lands on $v_i$ after $\tau$ steps.

Interestingly,  $\mathbf{S}_{\mathcal{G}}(\boldsymbol{\theta})$ does not weigh landing probabilities of different lengths independently. Instead, it accumulates the latter as weighted combinations (cf. \eqref{c}) in a basis of ``wavelet''-type functions of different resolution (see Fig. 1). 

Having established links to spectral clustering and random walks, our novel  $\mathbf{S}_{\mathcal{G}}(\boldsymbol{\theta})$ is well motivated as a family of node similarity matrices. Nevertheless, before devising an algorithm for learning $\boldsymbol{\theta}$ and testing it on real graphs, we will evaluate how well the basis $\{ \mathbf{S}^k\}_{k=1}^K$, on which $\mathbf{S}_{\mathcal{G}}(\boldsymbol{\theta})$ is built, can capture underlying node similarities. 


\setlength\figW{0.99\columnwidth}
\setlength\figH{0.53\columnwidth}

\section{Model expressiveness}

\begin{figure*}[t!]  
	\centering
	\begin{subfigure}[b]{0.28\textwidth}
		\centering
		{True SBM similarities ($\mathbf{S}^\ast$)}
		\vspace{-0.3cm}
		\includegraphics[width=\textwidth]{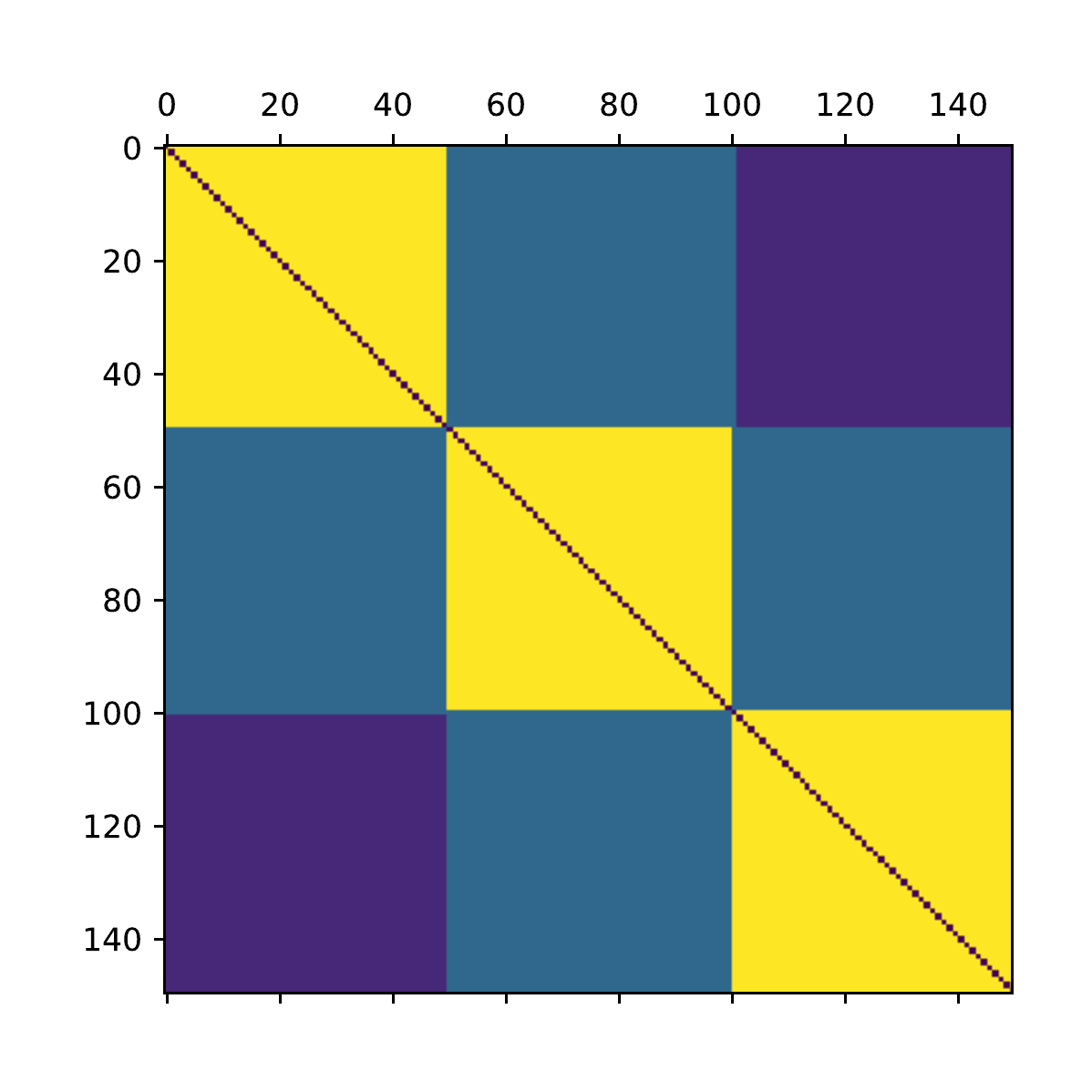}
		\label{fig:gull}
	\end{subfigure}
	~
	\begin{subfigure}[b]{0.28\textwidth}
		\centering
		{PageRank ($\hat{\mathbf{S}}_{PPR}$)}
		\vspace{-0.25cm}
		\includegraphics[width=\textwidth]{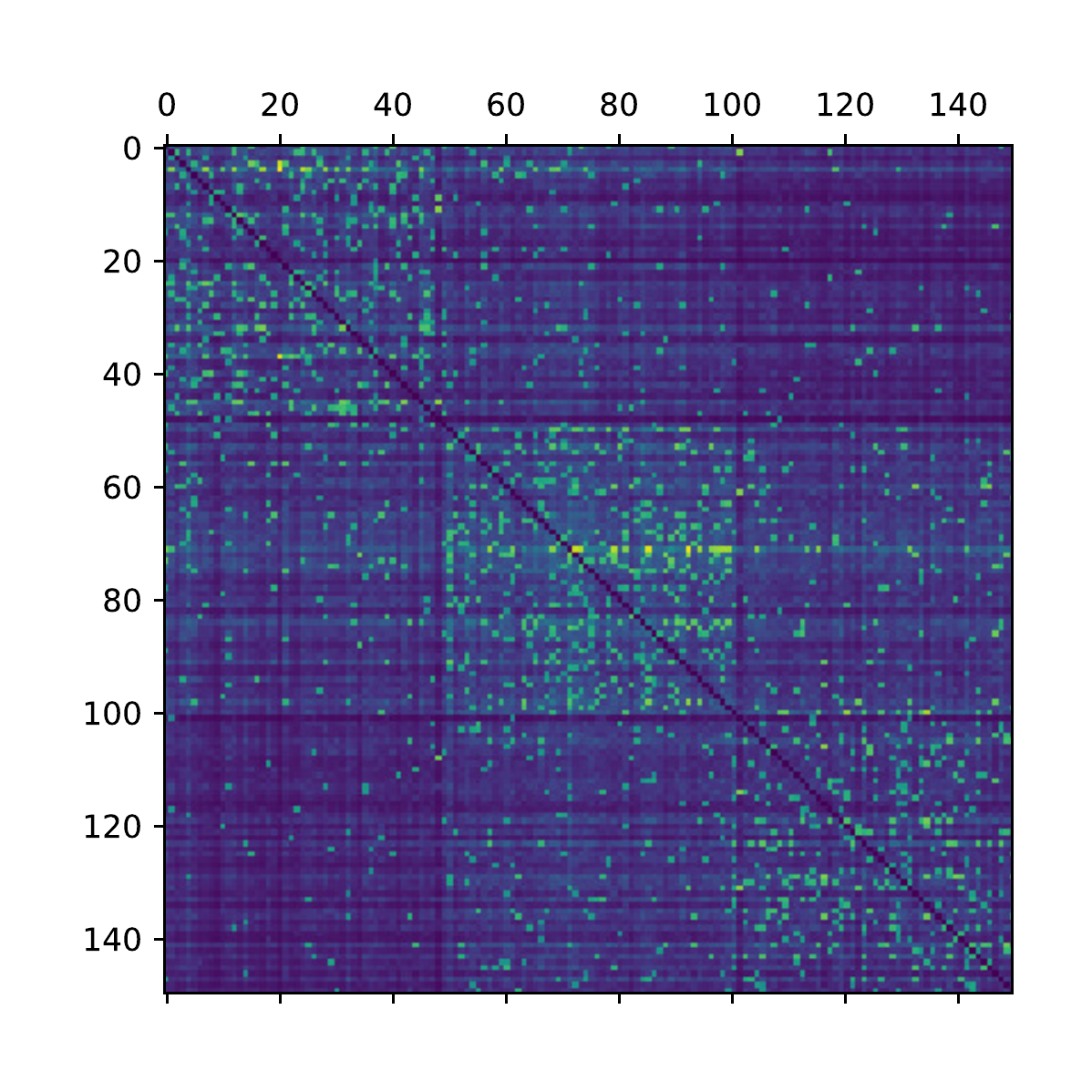}
		\label{fig:gull}
	\end{subfigure}
	~
	\begin{subfigure}[b]{0.28\textwidth}
		\centering
		{Adamic-Adar ($\hat{\mathbf{S}}_{AA}$)}
		\vspace{-0.25cm}
		\includegraphics[width=\textwidth]{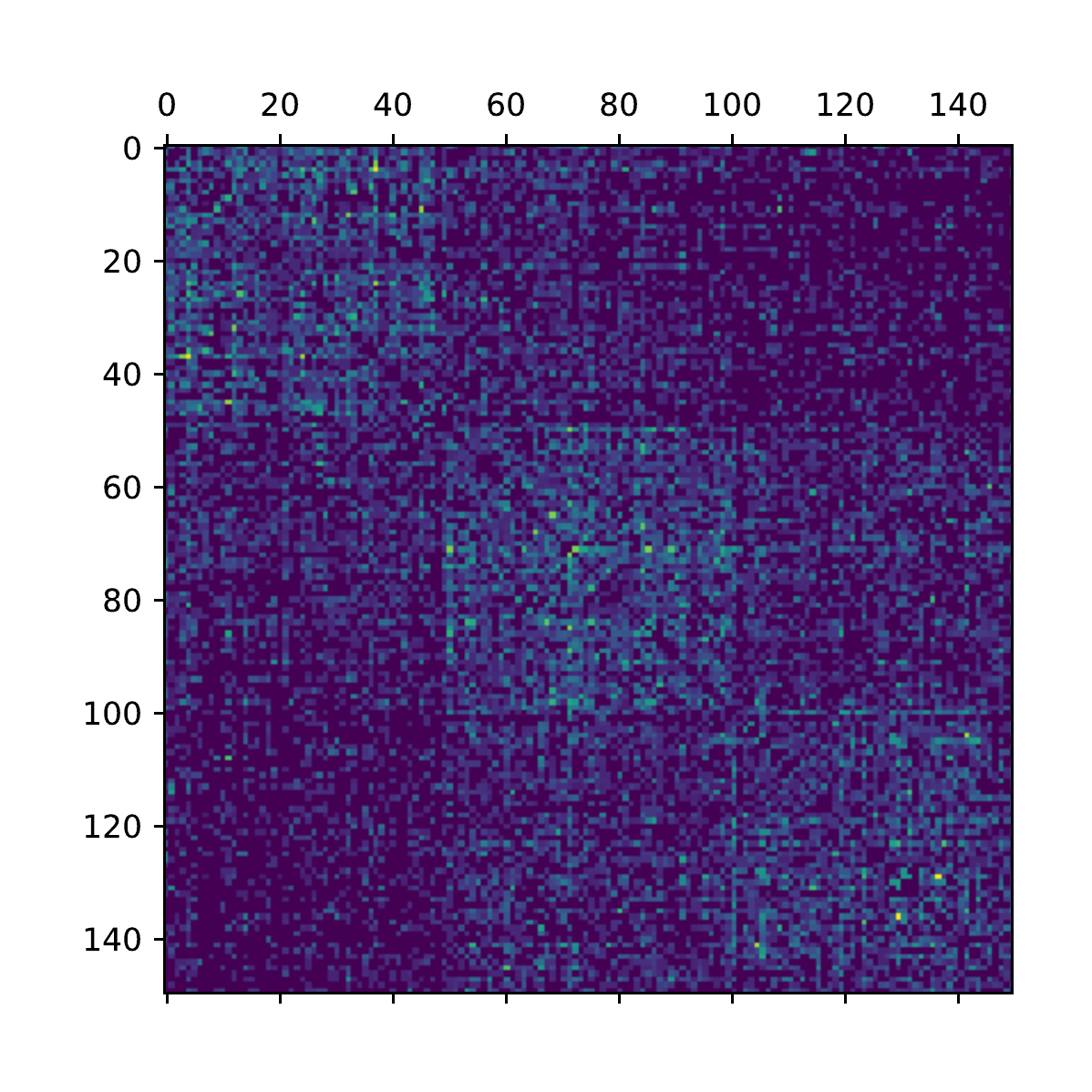}
		\label{fig:gull}
	\end{subfigure}
	
	\begin{subfigure}[b]{0.28\textwidth}
		\centering
		{Proposed ($\mathbf{S}^1$)}
		\vspace{-0.25cm}
		\includegraphics[width=\textwidth]{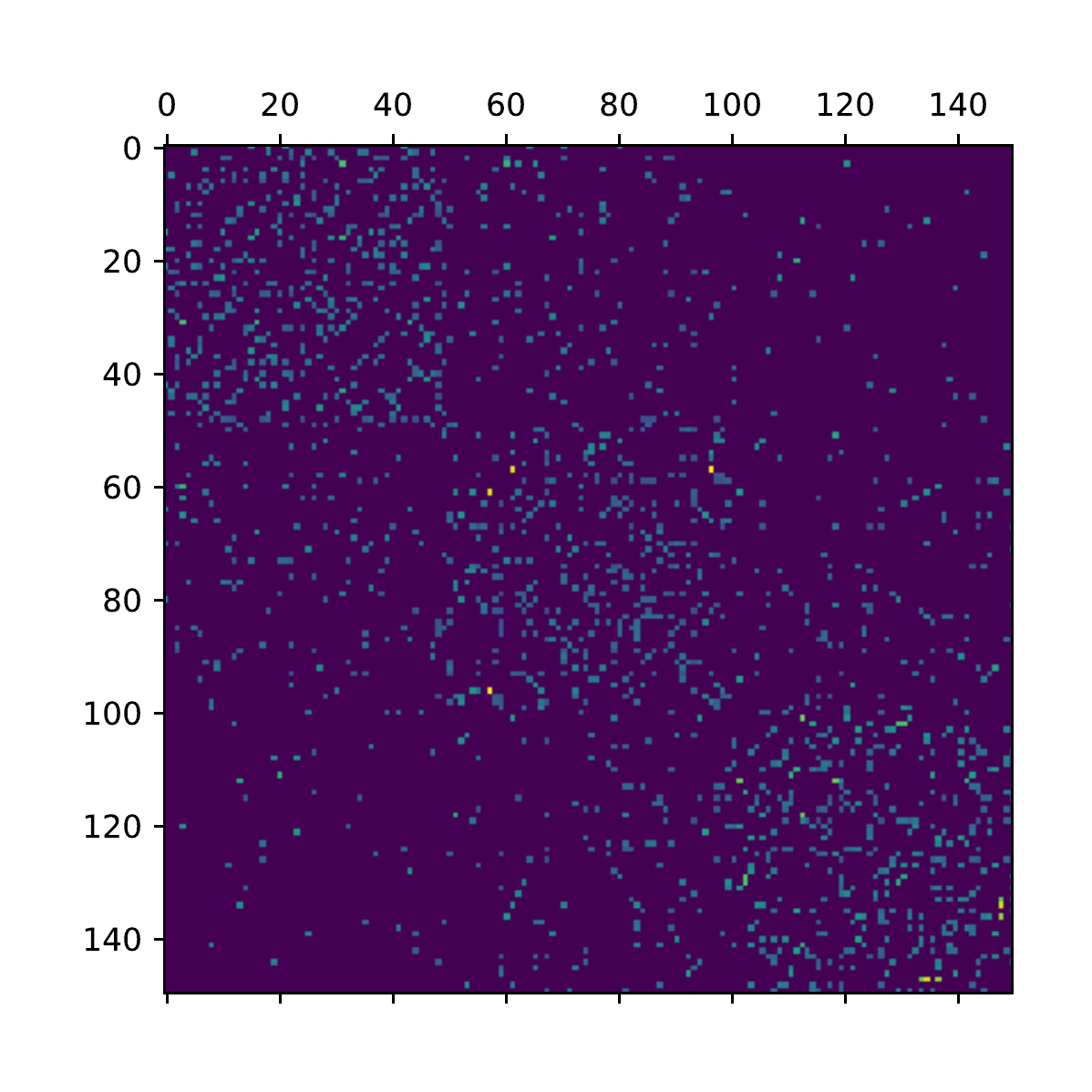}
		\label{fig:gull}
	\end{subfigure}
	~
	\begin{subfigure}[b]{0.28\textwidth}
		\centering
		{Proposed ($\mathbf{S}^6$)}
		\vspace{-0.25cm}
		\includegraphics[width=\textwidth]{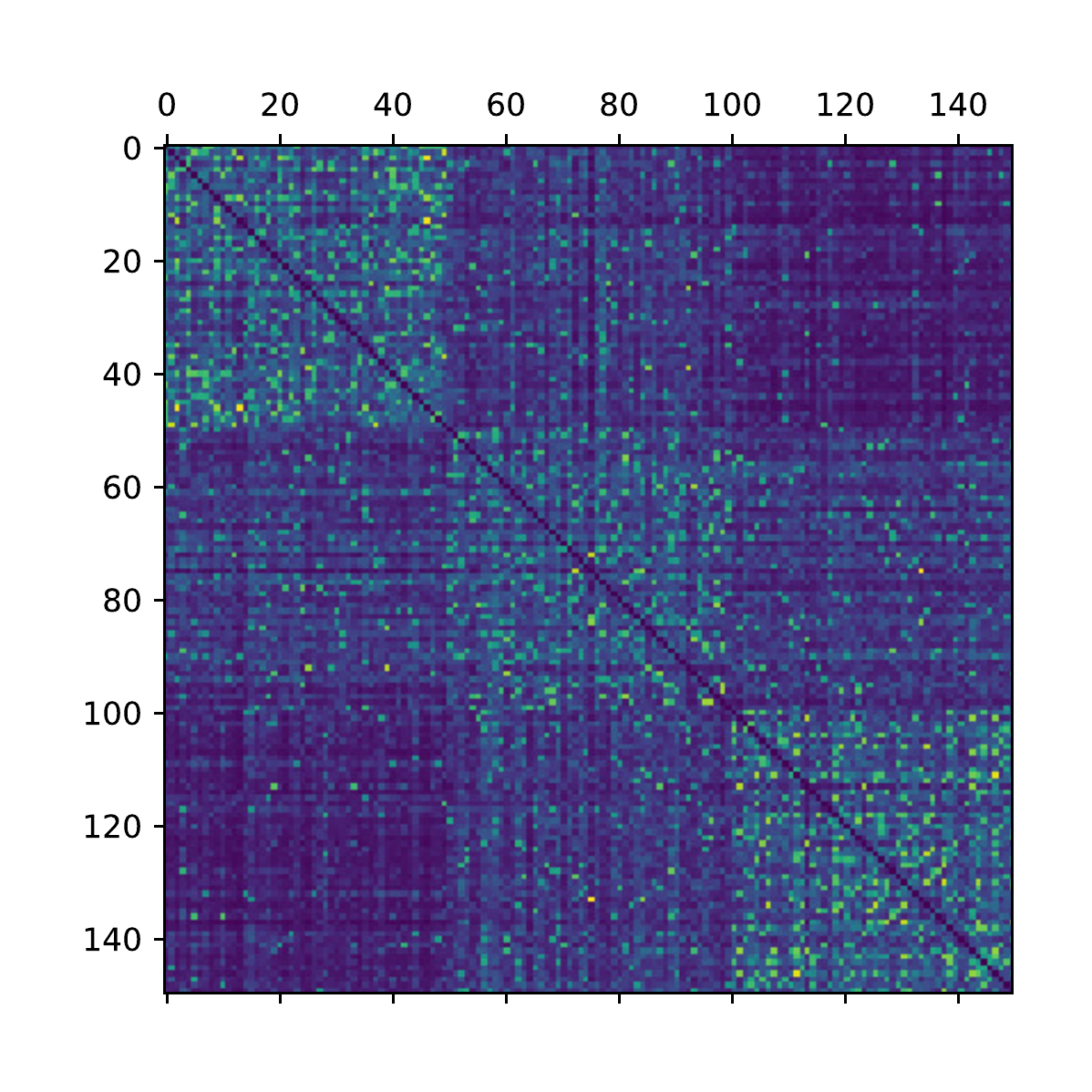}
		\label{fig:gull}
	\end{subfigure}
	~
	\begin{subfigure}[b]{0.28\textwidth}
		\centering
		{Proposed ($\mathbf{S}^{15}$)}
		\vspace{-0.25cm}
		\includegraphics[width=\textwidth]{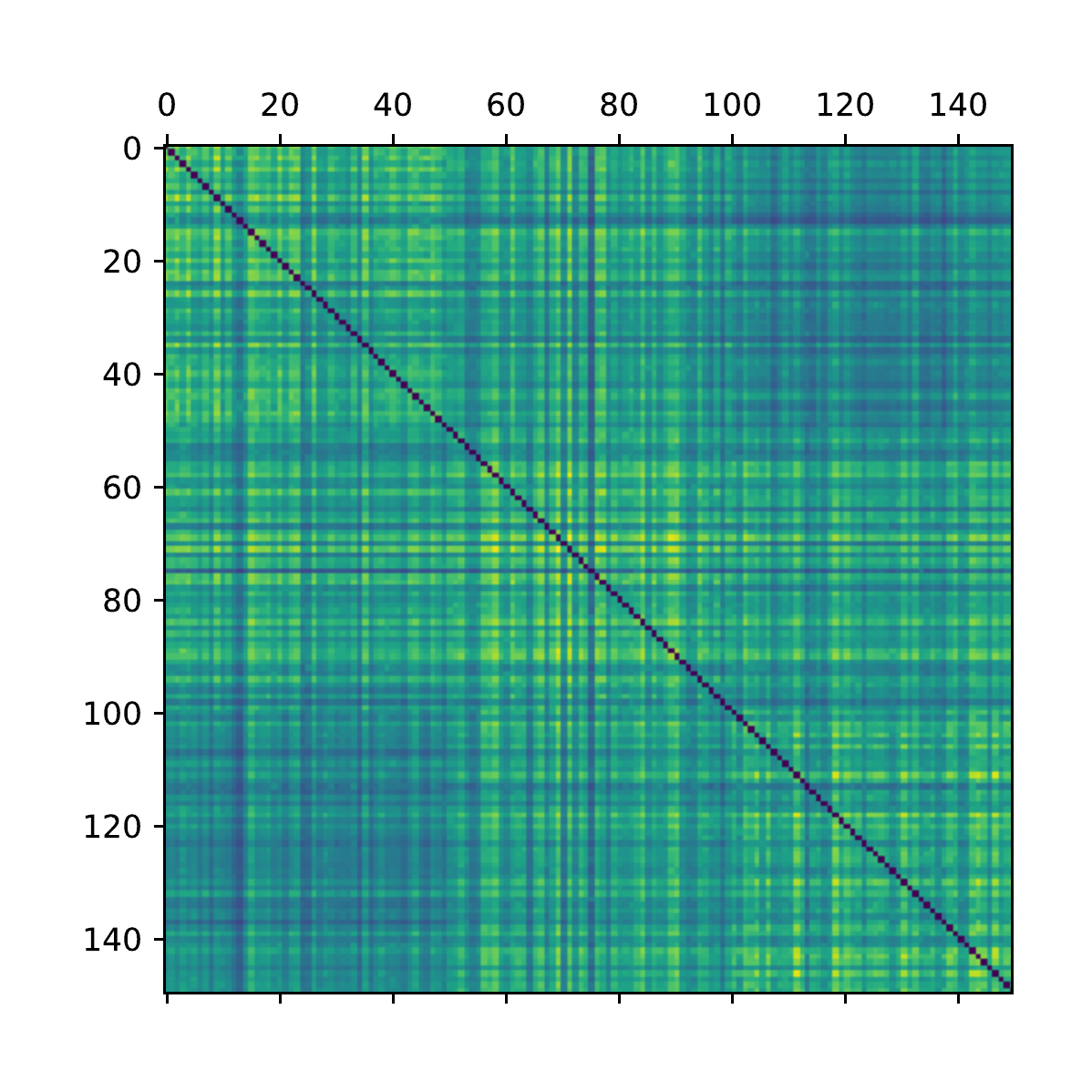}
		\label{fig:gull}
	\end{subfigure}
	\vspace{-0.2in}
	\caption{ Depiction of groundtruth and estimated similarity matrices, as yielded from an instance of the  numerical experiments described in Section 3.1.} \label{matplot}
	\vspace{-0.2in}
\end{figure*}
\setlength\figW{0.99\columnwidth}
\setlength\figH{0.6\columnwidth}
\begin{figure*}[t!]  
	\vspace{0.2in}
	\centering
\begin{tikzpicture}[scale=0.69]

\definecolor{color3}{rgb}{0.83921568627451,0.152941176470588,0.156862745098039}
\definecolor{color1}{rgb}{1,0.498039215686275,0.0549019607843137}
\definecolor{color2}{rgb}{0.172549019607843,0.627450980392157,0.172549019607843}
\definecolor{color4}{rgb}{0.580392156862745,0.403921568627451,0.741176470588235}
\definecolor{color0}{rgb}{0.12156862745098,0.466666666666667,0.705882352941177}
\definecolor{color5}{rgb}{0.549019607843137,0.337254901960784,0.294117647058824}

\begin{axis}[
legend cell align={left},
legend entries={{$\mathbf{S}^k$},{$\mathbf{A}^k$},{$\mathbf{S}_{PGR}$},{$\mathbf{S}_{KATZ}$},{$\mathbf{S}_{NEIGH}$},{$\mathbf{S}_{AA}$}},
legend style={at={(0.97,0.03)}, anchor=south east, draw=white!80.0!black},
legend columns = 6,
legend to name = {named2},
tick align=outside,
tick pos=left,
x grid style={lightgray!92.02614379084967!black},
xlabel={k},
ylabel = {QoM},
xmin=-1.45, xmax=30.45,
y grid style={lightgray!92.02614379084967!black},
ymin=0.45170935933847, ymax=0.919633448732827,
title =  {a) Clustered - dense ($p=0.3,~q=0.1$) }
]
\addplot [line width=2.0pt, color0]
table [row sep=\\]{%
0	0.472978636129123 \\
1	0.600996245552759 \\
2	0.725475530054492 \\
3	0.820444388424821 \\
4	0.87478110524039 \\
5	0.896194290927143 \\
6	0.898364171942174 \\
7	0.891637257356226 \\
8	0.881771526920018 \\
9	0.871502094791518 \\
10	0.861979343723982 \\
11	0.853601432526093 \\
12	0.846429819374151 \\
13	0.84038519024526 \\
14	0.835337608041813 \\
15	0.831147129867714 \\
16	0.827681328468145 \\
17	0.824822023776264 \\
18	0.822467000503656 \\
19	0.820529462255401 \\
20	0.818936548144356 \\
21	0.81762755242873 \\
22	0.816552150688366 \\
23	0.815668767662343 \\
24	0.814943137278194 \\
25	0.814347063447955 \\
26	0.813857370311059 \\
27	0.813455022196044 \\
28	0.813124391095146 \\
29	0.812852649931263 \\
};
\addplot [line width=1.0pt, color1, mark=*, mark size=2, mark options={solid}]
table [row sep=\\]{%
0	0.475932193012983 \\
1	0.861527311521903 \\
2	0.874157213397282 \\
3	0.848521107778989 \\
4	0.827647894475634 \\
5	0.813561147131587 \\
6	0.805126009284301 \\
7	0.799956419871489 \\
8	0.796864061209494 \\
9	0.794995314844841 \\
10	0.793871197242409 \\
11	0.793191872139456 \\
12	0.792781299203606 \\
13	0.792532511881177 \\
14	0.792381589888707 \\
15	0.792289875683167 \\
16	0.792234073482005 \\
17	0.792200075546196 \\
18	0.792179338382122 \\
19	0.792166675164255 \\
20	0.792158934158623 \\
21	0.792154197171475 \\
22	0.792151295550772 \\
23	0.792149516424712 \\
24	0.792148424500978 \\
25	0.792147753699458 \\
26	0.792147341214063 \\
27	0.792147087330465 \\
28	0.792146930917995 \\
29	0.792146834464303 \\
};
\addplot [line width=1.0pt, color2, mark=+, mark size=3, mark options={solid}]
table [row sep=\\]{%
0	0.839571502252236 \\
1	0.839571502252236 \\
2	0.839571502252236 \\
3	0.839571502252236 \\
4	0.839571502252236 \\
5	0.839571502252236 \\
6	0.839571502252236 \\
7	0.839571502252236 \\
8	0.839571502252236 \\
9	0.839571502252236 \\
10	0.839571502252236 \\
11	0.839571502252236 \\
12	0.839571502252236 \\
13	0.839571502252236 \\
14	0.839571502252236 \\
15	0.839571502252236 \\
16	0.839571502252236 \\
17	0.839571502252236 \\
18	0.839571502252236 \\
19	0.839571502252236 \\
20	0.839571502252236 \\
21	0.839571502252236 \\
22	0.839571502252236 \\
23	0.839571502252236 \\
24	0.839571502252236 \\
25	0.839571502252236 \\
26	0.839571502252236 \\
27	0.839571502252236 \\
28	0.839571502252236 \\
29	0.839571502252236 \\
};
\addplot [line width=1.0pt, color3, mark=diamond, mark size=3, mark options={solid}]
table [row sep=\\]{%
0	0.770384743492101 \\
1	0.770384743492101 \\
2	0.770384743492101 \\
3	0.770384743492101 \\
4	0.770384743492101 \\
5	0.770384743492101 \\
6	0.770384743492101 \\
7	0.770384743492101 \\
8	0.770384743492101 \\
9	0.770384743492101 \\
10	0.770384743492101 \\
11	0.770384743492101 \\
12	0.770384743492101 \\
13	0.770384743492101 \\
14	0.770384743492101 \\
15	0.770384743492101 \\
16	0.770384743492101 \\
17	0.770384743492101 \\
18	0.770384743492101 \\
19	0.770384743492101 \\
20	0.770384743492101 \\
21	0.770384743492101 \\
22	0.770384743492101 \\
23	0.770384743492101 \\
24	0.770384743492101 \\
25	0.770384743492101 \\
26	0.770384743492101 \\
27	0.770384743492101 \\
28	0.770384743492101 \\
29	0.770384743492101 \\
};
\addplot [line width=1.0pt, color4, mark=triangle*, mark size=3, mark options={solid,rotate=90}]
table [row sep=\\]{%
0	0.861527311521903 \\
1	0.861527311521903 \\
2	0.861527311521903 \\
3	0.861527311521903 \\
4	0.861527311521903 \\
5	0.861527311521903 \\
6	0.861527311521903 \\
7	0.861527311521903 \\
8	0.861527311521903 \\
9	0.861527311521903 \\
10	0.861527311521903 \\
11	0.861527311521903 \\
12	0.861527311521903 \\
13	0.861527311521903 \\
14	0.861527311521903 \\
15	0.861527311521903 \\
16	0.861527311521903 \\
17	0.861527311521903 \\
18	0.861527311521903 \\
19	0.861527311521903 \\
20	0.861527311521903 \\
21	0.861527311521903 \\
22	0.861527311521903 \\
23	0.861527311521903 \\
24	0.861527311521903 \\
25	0.861527311521903 \\
26	0.861527311521903 \\
27	0.861527311521903 \\
28	0.861527311521903 \\
29	0.861527311521903 \\
};
\addplot [ color5,line width=1.0pt, mark=triangle*, mark size=3, mark options={solid}]
table [row sep=\\]{%
0	0.86079525389901 \\
1	0.86079525389901 \\
2	0.86079525389901 \\
3	0.86079525389901 \\
4	0.86079525389901 \\
5	0.86079525389901 \\
6	0.86079525389901 \\
7	0.86079525389901 \\
8	0.86079525389901 \\
9	0.86079525389901 \\
10	0.86079525389901 \\
11	0.86079525389901 \\
12	0.86079525389901 \\
13	0.86079525389901 \\
14	0.86079525389901 \\
15	0.86079525389901 \\
16	0.86079525389901 \\
17	0.86079525389901 \\
18	0.86079525389901 \\
19	0.86079525389901 \\
20	0.86079525389901 \\
21	0.86079525389901 \\
22	0.86079525389901 \\
23	0.86079525389901 \\
24	0.86079525389901 \\
25	0.86079525389901 \\
26	0.86079525389901 \\
27	0.86079525389901 \\
28	0.86079525389901 \\
29	0.86079525389901 \\
};
\end{axis}

\end{tikzpicture}		
\begin{tikzpicture}[scale=0.69]

\definecolor{color3}{rgb}{0.83921568627451,0.152941176470588,0.156862745098039}
\definecolor{color5}{rgb}{0.549019607843137,0.337254901960784,0.294117647058824}
\definecolor{color2}{rgb}{0.172549019607843,0.627450980392157,0.172549019607843}
\definecolor{color4}{rgb}{0.580392156862745,0.403921568627451,0.741176470588235}
\definecolor{color1}{rgb}{1,0.498039215686275,0.0549019607843137}
\definecolor{color0}{rgb}{0.12156862745098,0.466666666666667,0.705882352941177}

\begin{axis}[
tick align=outside,
tick pos=left,
x grid style={white!69.01960784313725!black},
xlabel={k},
ylabel={QoM},
xmin=-1.45, xmax=30.45,
y grid style={white!69.01960784313725!black},
ymin=0.303289152651276, ymax=0.88446079352587,
title =  {b) Clustered - sparse ($p=0.15,~q=0.05$) }
]

\addplot [line width=2.0pt, color0]
table [row sep=\\]{%
0	0.329887194774063 \\
1	0.433514077679538 \\
2	0.547062114463455 \\
3	0.655872138508178 \\
4	0.744683801359657 \\
5	0.805702275559329 \\
6	0.840826003499224 \\
7	0.857126587976927 \\
8	0.861848037229801 \\
9	0.860282216487654 \\
10	0.855709617706899 \\
11	0.849982277244079 \\
12	0.844078475172287 \\
13	0.838479640472917 \\
14	0.833396158834865 \\
15	0.828895222245812 \\
16	0.824971265057437 \\
17	0.821584170543652 \\
18	0.818679709557766 \\
19	0.816200250633704 \\
20	0.814090158765246 \\
21	0.812298305743441 \\
22	0.810779024582359 \\
23	0.809492243574516 \\
24	0.808403206782042 \\
25	0.807482005405871 \\
26	0.806703042570798 \\
27	0.806044496801072 \\
28	0.805487817211939 \\
29	0.805017265342108 \\
};
\addplot [line width=1.0pt, color1, mark=*, mark size=2, mark options={solid}]
table [row sep=\\]{%
0	0.335853985926413 \\
1	0.676256735651958 \\
2	0.767271115794041 \\
3	0.800167885576622 \\
4	0.790513264500882 \\
5	0.77811252418394 \\
6	0.769455477219473 \\
7	0.763015264187243 \\
8	0.759009719567815 \\
9	0.756271585165045 \\
10	0.754548991493225 \\
11	0.753401772791162 \\
12	0.752668756464188 \\
13	0.752184812159231 \\
14	0.751872005002264 \\
15	0.751665908573656 \\
16	0.751531583023079 \\
17	0.751443016854092 \\
18	0.751384935543698 \\
19	0.751346568282107 \\
20	0.751321286173563 \\
21	0.75130454662945 \\
22	0.751293472974753 \\
23	0.751286123092333 \\
24	0.751281244860984 \\
25	0.751277999198081 \\
26	0.751275838767432 \\
27	0.751274397980005 \\
28	0.751273436450265 \\
29	0.751272793765262 \\
};
\addplot [line width=1.0pt, color2, mark=+, mark size=3, mark options={solid}]
table [row sep=\\]{%
0	0.813892826877329 \\
1	0.813892826877329 \\
2	0.813892826877329 \\
3	0.813892826877329 \\
4	0.813892826877329 \\
5	0.813892826877329 \\
6	0.813892826877329 \\
7	0.813892826877329 \\
8	0.813892826877329 \\
9	0.813892826877329 \\
10	0.813892826877329 \\
11	0.813892826877329 \\
12	0.813892826877329 \\
13	0.813892826877329 \\
14	0.813892826877329 \\
15	0.813892826877329 \\
16	0.813892826877329 \\
17	0.813892826877329 \\
18	0.813892826877329 \\
19	0.813892826877329 \\
20	0.813892826877329 \\
21	0.813892826877329 \\
22	0.813892826877329 \\
23	0.813892826877329 \\
24	0.813892826877329 \\
25	0.813892826877329 \\
26	0.813892826877329 \\
27	0.813892826877329 \\
28	0.813892826877329 \\
29	0.813892826877329 \\
};
\addplot [line width=1.0pt, color3, mark=diamond, mark size=3, mark options={solid}]
table [row sep=\\]{%
0	0.502370567451528 \\
1	0.502370567451528 \\
2	0.502370567451528 \\
3	0.502370567451528 \\
4	0.502370567451528 \\
5	0.502370567451528 \\
6	0.502370567451528 \\
7	0.502370567451528 \\
8	0.502370567451528 \\
9	0.502370567451528 \\
10	0.502370567451528 \\
11	0.502370567451528 \\
12	0.502370567451528 \\
13	0.502370567451528 \\
14	0.502370567451528 \\
15	0.502370567451528 \\
16	0.502370567451528 \\
17	0.502370567451528 \\
18	0.502370567451528 \\
19	0.502370567451528 \\
20	0.502370567451528 \\
21	0.502370567451528 \\
22	0.502370567451528 \\
23	0.502370567451528 \\
24	0.502370567451528 \\
25	0.502370567451528 \\
26	0.502370567451528 \\
27	0.502370567451528 \\
28	0.502370567451528 \\
29	0.502370567451528 \\
};
\addplot [line width=1.0pt, color4, mark=triangle*, mark size=3, mark options={solid,rotate=90}]
table [row sep=\\]{%
0	0.676256735651958 \\
1	0.676256735651958 \\
2	0.676256735651958 \\
3	0.676256735651958 \\
4	0.676256735651958 \\
5	0.676256735651958 \\
6	0.676256735651958 \\
7	0.676256735651958 \\
8	0.676256735651958 \\
9	0.676256735651958 \\
10	0.676256735651958 \\
11	0.676256735651958 \\
12	0.676256735651958 \\
13	0.676256735651958 \\
14	0.676256735651958 \\
15	0.676256735651958 \\
16	0.676256735651958 \\
17	0.676256735651958 \\
18	0.676256735651958 \\
19	0.676256735651958 \\
20	0.676256735651958 \\
21	0.676256735651958 \\
22	0.676256735651958 \\
23	0.676256735651958 \\
24	0.676256735651958 \\
25	0.676256735651958 \\
26	0.676256735651958 \\
27	0.676256735651958 \\
28	0.676256735651958 \\
29	0.676256735651958 \\
};
\addplot [line width=1.0pt, color5, mark=triangle*, mark size=3, mark options={solid}]
table [row sep=\\]{%
0	0.665134599477324 \\
1	0.665134599477324 \\
2	0.665134599477324 \\
3	0.665134599477324 \\
4	0.665134599477324 \\
5	0.665134599477324 \\
6	0.665134599477324 \\
7	0.665134599477324 \\
8	0.665134599477324 \\
9	0.665134599477324 \\
10	0.665134599477324 \\
11	0.665134599477324 \\
12	0.665134599477324 \\
13	0.665134599477324 \\
14	0.665134599477324 \\
15	0.665134599477324 \\
16	0.665134599477324 \\
17	0.665134599477324 \\
18	0.665134599477324 \\
19	0.665134599477324 \\
20	0.665134599477324 \\
21	0.665134599477324 \\
22	0.665134599477324 \\
23	0.665134599477324 \\
24	0.665134599477324 \\
25	0.665134599477324 \\
26	0.665134599477324 \\
27	0.665134599477324 \\
28	0.665134599477324 \\
29	0.665134599477324 \\
};
\end{axis}

\end{tikzpicture}
\begin{tikzpicture}[scale=0.69]

\definecolor{color5}{rgb}{0.549019607843137,0.337254901960784,0.294117647058824}
\definecolor{color3}{rgb}{0.83921568627451,0.152941176470588,0.156862745098039}
\definecolor{color1}{rgb}{1,0.498039215686275,0.0549019607843137}
\definecolor{color0}{rgb}{0.12156862745098,0.466666666666667,0.705882352941177}
\definecolor{color2}{rgb}{0.172549019607843,0.627450980392157,0.172549019607843}
\definecolor{color4}{rgb}{0.580392156862745,0.403921568627451,0.741176470588235}

\begin{axis}[
tick align=outside,
tick pos=left,
x grid style={white!69.01960784313725!black},
xlabel={k},
ylabel = {QoM},
xmin=-1.45, xmax=30.45,
y grid style={white!69.01960784313725!black},
ymin=0.265374787861692, ymax=0.977579905526989,
title =  {c) Unstructured ($p=0.1,~q=0.1$) }
]

\addplot [line width=2.0pt, color0]
table [row sep=\\]{%
0	0.29774774775557 \\
1	0.406665262306347 \\
2	0.534119640098366 \\
3	0.662733031607305 \\
4	0.771217920888058 \\
5	0.847788176416681 \\
6	0.894542692522315 \\
7	0.920351450880444 \\
8	0.933679801353575 \\
9	0.940229139939164 \\
10	0.943286824647859 \\
11	0.944609481903331 \\
12	0.945099013228865 \\
13	0.945206945633112 \\
14	0.945154214698327 \\
15	0.945045701043127 \\
16	0.944928746646226 \\
17	0.944822720264469 \\
18	0.944733815086294 \\
19	0.944662363059464 \\
20	0.944606377592332 \\
21	0.944563209719205 \\
22	0.944530272783148 \\
23	0.944505317511653 \\
24	0.944486498685311 \\
25	0.944472352380104 \\
26	0.944461740976812 \\
27	0.944453792182189 \\
28	0.944447843062394 \\
29	0.944443392799152 \\
};
\addplot [line width=1.0pt, color1, mark=*, mark size=2, mark options={solid}]
table [row sep=\\]{%
0	0.30606562700778 \\
1	0.710679839572812 \\
2	0.829210770703123 \\
3	0.883199830465284 \\
4	0.888745973870826 \\
5	0.88986104420691 \\
6	0.889814023203863 \\
7	0.889684702718186 \\
8	0.889619462729622 \\
9	0.889578833665517 \\
10	0.889561917963888 \\
11	0.889552774321015 \\
12	0.889548884158413 \\
13	0.889546895251697 \\
14	0.889546010294777 \\
15	0.889545573787009 \\
16	0.889545370424489 \\
17	0.889545273074229 \\
18	0.88954522567511 \\
19	0.889545203611867 \\
20	0.889545192404526 \\
21	0.889545187329122 \\
22	0.889545184642687 \\
23	0.889545183459024 \\
24	0.889545182806737 \\
25	0.889545182527141 \\
26	0.889545182366836 \\
27	0.889545182299992 \\
28	0.889545182260147 \\
29	0.889545182243981 \\
};
\addplot [line width=1.0pt, color2, mark=+, mark size=3, mark options={solid}]
table [row sep=\\]{%
0	0.914286753922571 \\
1	0.914286753922571 \\
2	0.914286753922571 \\
3	0.914286753922571 \\
4	0.914286753922571 \\
5	0.914286753922571 \\
6	0.914286753922571 \\
7	0.914286753922571 \\
8	0.914286753922571 \\
9	0.914286753922571 \\
10	0.914286753922571 \\
11	0.914286753922571 \\
12	0.914286753922571 \\
13	0.914286753922571 \\
14	0.914286753922571 \\
15	0.914286753922571 \\
16	0.914286753922571 \\
17	0.914286753922571 \\
18	0.914286753922571 \\
19	0.914286753922571 \\
20	0.914286753922571 \\
21	0.914286753922571 \\
22	0.914286753922571 \\
23	0.914286753922571 \\
24	0.914286753922571 \\
25	0.914286753922571 \\
26	0.914286753922571 \\
27	0.914286753922571 \\
28	0.914286753922571 \\
29	0.914286753922571 \\
};
\addplot [line width=1.0pt, color3, mark=diamond, mark size=3, mark options={solid}]
table [row sep=\\]{%
0	0.534675280589581 \\
1	0.534675280589581 \\
2	0.534675280589581 \\
3	0.534675280589581 \\
4	0.534675280589581 \\
5	0.534675280589581 \\
6	0.534675280589581 \\
7	0.534675280589581 \\
8	0.534675280589581 \\
9	0.534675280589581 \\
10	0.534675280589581 \\
11	0.534675280589581 \\
12	0.534675280589581 \\
13	0.534675280589581 \\
14	0.534675280589581 \\
15	0.534675280589581 \\
16	0.534675280589581 \\
17	0.534675280589581 \\
18	0.534675280589581 \\
19	0.534675280589581 \\
20	0.534675280589581 \\
21	0.534675280589581 \\
22	0.534675280589581 \\
23	0.534675280589581 \\
24	0.534675280589581 \\
25	0.534675280589581 \\
26	0.534675280589581 \\
27	0.534675280589581 \\
28	0.534675280589581 \\
29	0.534675280589581 \\
};
\addplot [line width=1.0pt, color4, mark=triangle*, mark size=3, mark options={solid,rotate=90}]
table [row sep=\\]{%
0	0.710679839572812 \\
1	0.710679839572812 \\
2	0.710679839572812 \\
3	0.710679839572812 \\
4	0.710679839572812 \\
5	0.710679839572812 \\
6	0.710679839572812 \\
7	0.710679839572812 \\
8	0.710679839572812 \\
9	0.710679839572812 \\
10	0.710679839572812 \\
11	0.710679839572812 \\
12	0.710679839572812 \\
13	0.710679839572812 \\
14	0.710679839572812 \\
15	0.710679839572812 \\
16	0.710679839572812 \\
17	0.710679839572812 \\
18	0.710679839572812 \\
19	0.710679839572812 \\
20	0.710679839572812 \\
21	0.710679839572812 \\
22	0.710679839572812 \\
23	0.710679839572812 \\
24	0.710679839572812 \\
25	0.710679839572812 \\
26	0.710679839572812 \\
27	0.710679839572812 \\
28	0.710679839572812 \\
29	0.710679839572812 \\
};
\addplot [line width=1.0pt, color5, mark=triangle*, mark size=3, mark options={solid}]
table [row sep=\\]{%
0	0.697974511624638 \\
1	0.697974511624638 \\
2	0.697974511624638 \\
3	0.697974511624638 \\
4	0.697974511624638 \\
5	0.697974511624638 \\
6	0.697974511624638 \\
7	0.697974511624638 \\
8	0.697974511624638 \\
9	0.697974511624638 \\
10	0.697974511624638 \\
11	0.697974511624638 \\
12	0.697974511624638 \\
13	0.697974511624638 \\
14	0.697974511624638 \\
15	0.697974511624638 \\
16	0.697974511624638 \\
17	0.697974511624638 \\
18	0.697974511624638 \\
19	0.697974511624638 \\
20	0.697974511624638 \\
21	0.697974511624638 \\
22	0.697974511624638 \\
23	0.697974511624638 \\
24	0.697974511624638 \\
25	0.697974511624638 \\
26	0.697974511624638 \\
27	0.697974511624638 \\
28	0.697974511624638 \\
29	0.697974511624638 \\
};
\end{axis}

\end{tikzpicture}
	\phantom{ppppp}\ref{named2}
	\caption{ Quality of match between true SBM similarity and various estimates, as yielded from experiments of Section 3.1.} \label{curves}
\end{figure*}
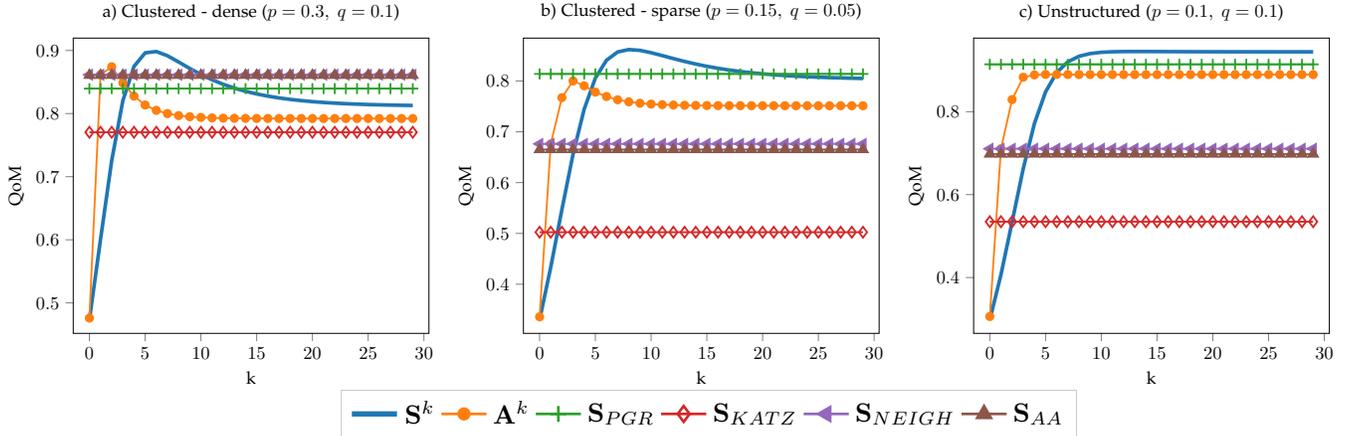

This section introduces a performance metric that quantifies how well a node similarity matrix derived from the graph itself matches the ``true'' underlying similarity structure between nodes. The discussion is followed by numerical evaluation of the performance of different similarity matrices (including the one in \eqref{expand1}) on graphs that are generated according to the stochastic block model \cite{sbm}.

To begin, suppose that for a given set of nodes, an adjacency matrix $\mathbf{A}$ is generated as
\begin{equation*}
\mathbf{A} \sim f_A(\mathbf{A})
\end{equation*}
where $f_A(\mathbf{A})$ is a probability density function defined over the space of all possible adjacency matrices. Let the ``true'' underlying similarity between nodes $v_i$ and $v_j$ be   
\begin{equation*}
s^\ast(v_i,v_j) := \Pr\{ (i,j) \in \mathcal{E} \} = \mathbb{E}_{f_A}\left[ A_{i,j} \right]
\end{equation*}
which is the probability that the two nodes are connected. The ``true'' similarity matrix is thus given as the expected adjacency matrix 
\begin{equation*}
\mathbf{S}^\ast := \mathbb{E}_{f_A}\left[ \mathbf{A} \right].
\end{equation*}
We define the quality-of-match (QoM) between the underlying $\mathbf{S}^\ast$ and any similarity $\hat{\mathbf{S}} = F(\mathbf{A})$ estimated from the adjacency matrix as
\begin{equation}\label{quality}
\mathrm{QoM} := \mathbb{E}_{f_A}\left[\mathrm{PC}\left(\mathbf{S}^\ast,F(\mathbf{A})\right)\right]
\end{equation}
where
\begin{equation}\label{pearson}
\mathrm{PC}\left(\mathbf{X}_1,\mathbf{X}_2\right):= \frac{\left(\mathrm{vec}\left(\mathbf{X}_1\right)\right)^\top \mathrm{vec}\left(\mathbf{X}_2\right)}{ \|\mathbf{X}_1\|_F \|\mathbf{X}_2\|_F }
\end{equation}
is the Pearson correlation between two matrices $\mathbf{X}_1$ and $\mathbf{X}_2$, with $\mathrm{vec}\left(\mathbf{X}\right)$ denoting matrix vectorization. The latter is used for appropriate rescaling of the ``true'' similarity matrix in order for the comparison with $\mathbf{S}_{\mathcal{G}}$ to be meaningful. Intuitively, \eqref{quality} measures how well the estimated node similarities in $\hat{\mathbf{S}}$ are expected to match the pattern of true underlying similarities in $\mathbf{S}^\ast$, when edges are generated according to the known $f_A(\cdot)$.

\subsection{Numerical experiments and observations}	
We numerically evaluate the QoM achieved by different similarity matrices, on a set of $N$ nodes whose interconnections are generated according to a stochastic block model (SBM). For this set of experiments, we divided the nodes into three clusters of equal size
\begin{equation*}\label{clusters}
\mathcal{C}_l = \{ i: (l-1)N/3 \leq i \leq lN/3 \},~l\in \{ 1,2,3 \}
\end{equation*}
with inter- and intra-connection probabilities 
\begin{equation}\label{probs}
\Pr\{ (i,j) \in \mathcal{E} \} = \left \{ \begin{array}{cc}
p~&~, (i,j) \mathrm{~in~the~same~}\mathcal{C}_l   \\
cq~&~, i\in\mathcal{C}_1\mathrm{~and~}j\in\mathcal{C}_3 \\
q~&~\mathrm{else}
\end{array} \right. 
\end{equation}
where $p$ is the probability of connection when two nodes belong to the same cluster, and $c<1$ introduces asymmetry and a hierarchical clustering organization (see Fig. 2-top left), by making two of the clusters less likely to connect; we have related Python scripts available.\footnote{https://github.com/DimBer/ASE-project/tree/master/sim\textunderscore tests} The SBM probability matrix \cite{sbm} is given as 
\begin{equation} \label{sbm_mat}
\mathbf{W}_\mathrm{sbm} =\left[ \begin{array}{ccc}
p~&~ q~&~ c q \\
q~&~p~&~q \\
c q~&~q~&~p
\end{array} \right] 
\end{equation}
and the underlying similarity can be expressed as  
\begin{equation}\label{true_similarity}
\mathbf{S}^\ast = \mathbb{E}\left[\mathbf{A}\right] = \mathbf{W}_\mathrm{sbm} \otimes \left( \mathbf{1}_{N/3}\mathbf{1}_{N/3}^T \right) - \mathrm{diag}(p\mathbf{1}_N)
\end{equation}
where $\otimes$ denotes the Kronecker product. 

For each experiment, we set $N=150$ and generated a graph according to 
\eqref{probs}. We then compared the QoM between \eqref{true_similarity} and the $k$th power of the proposed \eqref{my_matrix}, the $k$th power of the adjacency ($\mathbf{A}^k$), as well as each of the following well known similarity metrics:
 \begin{itemize}
	\item $\hat{\mathbf{S}}_{PPR}:= (1-\alpha)(\mathbf{I} - \alpha \mathbf{AD}^{-1})^{-1}$: the steady state probability that a random walk  restarting at $v_j$ with probability $1-\alpha$ at every step is located at $v_i$. Essentially a personalized PageRank (PPR) computed for every node of the graph, inheriting the properties of the celebrated centrality measure \cite{brin2012reprint,GleichBeyond,kloumann2017block}.  
	\item $\hat{\mathbf{S}}_{KATZ}:= (1-\beta)(\mathbf{I} - \beta \mathbf{A})^{-1}\mathbf{A}$ : the Katz index \cite{arope}, an exponentially weighted summation over paths of all possible hops between
	two nodes.
	\item $\hat{\mathbf{S}}_{NEIGH}:=\mathbf{A}^2$: the number of common neighbors that every pair of nodes shares.
	\item $\hat{\mathbf{S}}_{AA}:=\mathbf{A}\mathbf{D}^{-1}\mathbf{A}$:	Adamic-Adar \cite{adamic} is a variant of common neighbors 
	where each set of neighbors is weighted inversely proportional to its cardinality.
\end{itemize}
The resulting QoM was averaged over 200 experiments. Parameters $\alpha$ in $\hat{\mathbf{S}}_{PPR}$ and $\beta$ in $\hat{\mathbf{S}}_{KATZ}$ were tuned to maximize the performance of the metrics. Figure 3 depicts QoM as a function of $k$, for three different scenarios.

In the first scenario (Fig. 3-a), with graphs being dense and clustered ($p=0.3$, $q=0.1$), the proposed $\mathbf{S}^k$ improves sharply in the first few steps, reaching maximum QoM after 4 or 5 steps, and gradually decreases as $k$ continues to increase. The $k$th order proximities that are given as entries of $\mathbf{A}^k$ follow a similar trend, however their QoM peaks shortly after 2 or 3 steps and declines fast for larger $k$. The matrix plots of a randomly selected experiment depicted in Fig. 2 can aid in understanding the underlying mechanism that gives rise to this highly step-dependent behavior. Specifically, $\mathbf{S}^1$ (bottom left) that has the same sparsity pattern as the adjacency is a poor match to the dense block-structure of $\mathbf{S}^\ast$. On the other side of the spectrum, $\mathbf{S}^{15}$ (bottom right) is too ``flat'' and also a poor similarity metric. Meanwhile, taking $k=6$ promotes enough mixing without ``dissipating.'' As a result, $\mathbf{S}^{6}$ (bottom center) visibly matches the structure of $\mathbf{S}^\ast$.
Interestingly, for $k\in[4,10]$ the proposed $\mathbf{S}^k$ surpasses in QoM all other similarity metrics that were tested. Nevertheless, the simple $2-$hop Adamic-adar, common-neighbors similarities perform reasonably well by exploiting the relatively dense structure of the graphs.    

Results were markedly different in the second scenario shown in Fig. 3-b. Here, graphs were generated with the same clustering structure but significantly sparser, with edge probability parameters $p=0.15$ and $q=0.05$. For sparser graphs, $\mathbf{A}^k$ and $\mathbf{S}^k$ require more steps to reach peak QoM (4 and 9 respectively). Similarly, PPR which relies on long paths performs much better than the short-reaching Adamic-Adar. This behavior is intuitively reasonable because the sparser a graph is, the longer become the paths that need to be explored around each node, in order for the latter to ``gauge'' its position on the graph. 

Finally, a third scenario (Fig. 3-c) was examined, where each graph was generated without a clustering structure ($p=q=0.1$ and $c=1$); essentially an Erdos-Renyi graph. For this degenerate case that is of no real practical interest, all pairs of nodes are equally similar; this type of similarity requires infinitely long paths to be described.

In a nutshell, the presented numerical study hints at the two following facts. First, $\mathbf{S}^k$ can successfully model similarities that are based on grouping nodes in arbitrary and multilevel sets with variable degrees of homophily and heterophily. The second fact, is that the performance of $\mathbf{S}^k$ varies significantly with $k$. Moreover, the way that $k$ affects performance may also vary from graph to graph, depending on the underlying properties -- what suggests viewing this way as a  graph ``signature'' that is also validated by the real graphs in Section 6. Thus, a principled means of specifying  $\mathbf{S}_{\mathcal{G}}(\boldsymbol{\theta})$ by learning the parameters that match this graph ``signature'' in an unsupervised mode, is highly motivated. 

\section{ Unsupervised similarity learning}\label{sec:main}	

We have arrived at the point where for a given graph, it is prudent to select a specific $\boldsymbol{\theta}\in\mathcal{S}^K$ without supervision. Following the discussion in Section 3, it would be ideal to fit $\mathbf{S}_{\mathcal{G}}(\boldsymbol{\theta})$ to a true $\mathbf{S}^\ast$  
by minimizing an expected cost
\begin{equation}\label{ideal}
\boldsymbol{\theta}^\ast = \arg\min_{\boldsymbol{\theta}\in \mathcal{S}^K}\mathbb{E}_{f_A}\left[\ell\left(\mathbf{S}^\ast,\mathbf{S}_\mathcal{G}(\mathbf{A};\boldsymbol{\theta})\right)\right] \:.
\end{equation}
Unfortunately, we only have one realization $\mathbf{A}$ of $f_A(\cdot)$, which means that without prior knowledge, the best approximation of $\mathbf{S}^\ast$ that we can obtain is the adjacency matrix itself, that is $\mathbf{S}^\ast \approx \mathbf{A}$. Using this approximation yields
\begin{equation}\label{wrong}
\min_{\boldsymbol{\theta}\in \mathcal{S}^K}\ell\left(\mathbf{A},\mathbf{S}_\mathcal{G}(\mathbf{A};\boldsymbol{\theta})\right).
\end{equation}
While straightforward, \eqref{wrong} yields embeddings with limited generalization capability. Simply put, regardless of the choice of $\ell(\cdot)$, solving \eqref{wrong} amounts to predicting a set of edges by tuning a similarity metric that is generated by the \emph{same} set of edges.

To mitigate overfitting but also promote generalization of the similarity metric and of the resulting embeddings, we explore the following idea. Suppose we are given a pair $\mathbf{A}_1,\mathbf{A}_2$ of adjacency matrices both drawn independently from $f_A(\cdot)$. In this case, we would be able to use one as approximation of $\mathbf{S}^\ast \approx \mathbf{A}_1$, and the other to form the multihop similarity matrix $\mathbf{S}_\mathcal{G}(\mathbf{A}_2;\boldsymbol{\theta})$; parameters $\boldsymbol{\theta}$ can then be learned by solving
\begin{equation}\label{ok1}
\min_{\boldsymbol{\theta}\in \mathcal{S}^K}\ell\left(\mathbf{A}_1,\mathbf{S}_\mathcal{G}(\mathbf{A}_2;\boldsymbol{\theta})\right).
\end{equation}     
Since separate samples are not available, we approximate the aforementioned process by randomly extracting part of $\mathbf{A}$ and approaching \eqref{ok1} as 
\begin{equation}\label{ok2}
\min_{\boldsymbol{\theta}\in \mathcal{S}^K}\ell_\mathcal{S}\left(\mathbf{A},\mathbf{S}_\mathcal{G}(\mathbf{A}\ast\mathbf{S}^c;\boldsymbol{\theta})\right)
\end{equation}     
where $\mathcal{S} \in \{ 1, \ldots, N \}^2$ is a subset of all possible pairs of nodes with $|\mathcal{S}| = N_s$, and $\mathbf{S}^c$ is an $N\times N$ binary section matrix with $S^c_{i,j} =0$, if $\{i,j\}\in \mathcal{S}$, and $S^c_{i,j} =1$, otherwise. Furthermore, $\ell_\mathcal{S}(\cdot,\cdot)$ in \eqref{ok2} denotes cost $\ell(\cdot,\cdot)$ applied selectively only to entries of the matrix variables that belong to $\mathcal{S}$. Here, such that $ \mathcal{S} = \mathcal{S^+} \cup \mathcal{S^-}$, with $\mathcal{S^+}\in\mathcal{E}$ being as subset of the edges and $\mathcal{S^-}\in \{1, \ldots, N\}^2\setminus\mathcal{E}$ a subset of node index tuples that are not connected (non-edges). To balance the influence of existing and non-existing edges, we use subsets of equal cardinality, that is $|\mathcal{S}^+| =|\mathcal{S}^-| = N_s/2$. 

To arrive from the unsupervised similarity learning framework \eqref{ok2} to a practical method, it remains to specify two modular sub-systems: one responsible for sampling edges, and one specifying $\ell(\cdot,\cdot)$ to find $\boldsymbol{\theta}^\ast$ by solving \eqref{ok2}.

\subsection{Edge sampling }
The choice of the sampling scheme for $\mathcal{S}$ plays an important role in the overall performance of the proposed adaptive embedding framework. Ideally, edge sampling should take into account the following criteria.
 \begin{itemize}
 	\item Sample $\mathcal{S}^+$ should be representative of the graph;
 	\item Edge removal should inflict minimal perturbation;
 	\item Edge removal should avoid isolating nodes; and 
 	\item Sampling scheme should be simple and scalable.
 \end{itemize}
Aiming at a `sweet spot' of these objectives, we populate $\mathcal{S}^+$ by sampling edges according to the following procedure: first, a node $v_1$ is sampled uniformly at random from $\mathcal{V}$; then, a second node $v_2$ is sampled uniformly from the neighborhood set $\mathcal{N}_\mathcal{G}(v_1)$ of $v_1$. The selected edge is removed only if both adjacent nodes have degree greater than one. Non-edges $\mathcal{S}^-$ are obtained by uniform sampling without replacement over $\{1, \ldots, N\}^2\setminus\mathcal{E}$. The overall procedure is summarized in Algorithm \ref{alg:ES}. For $N_s\ll N$, sampling probabilities remain approximately unchanged despite the removals, since the probability of selecting the same node is relatively small. Thus, one may approximate $\Pr\{e_t = (i,j)\}\approx\Pr\{e_0 = (i,j)\}$, and assuming for simplicity that $d_i>1\forall i$, it follows that     
\begin{align}\nonumber
\Pr\{e_0 = (i,j)\} &= \Pr\{v_1 = i, v_2 = j\} + \Pr\{v_1 = j, v_2 = i\}\\\nonumber
&=\Pr\{v_2 = i|v_1 = j\}\Pr\{v_1 = j\}\\\nonumber
 &~+ \Pr\{v_2 = j|v_1 = i\}\Pr\{v_1 = i\}\\ \label{probability}
 &= \frac{1}{d_j}\frac{1}{N} + \frac{1}{d_i}\frac{1}{N} \propto \frac{d_i + d_j}{d_i d_j},
\end{align}
meaning that edge $e=(i,j)$ is removed with probability that is proportional to the harmonic mean of the degrees of the nodes that it connects.  As shown in \cite{pertubation}, the perturbation that the removal of edge $e=(i,j)$ inflicts on the spectrum of an undirected graph is proportional to $d_id_j$; that is, removing edges that connect high-degree nodes leads to higher perturbation. Thus, Algorithm \ref{alg:ES} tends to inflict minimal perturbation by sampling with probability that is inversely proportional to $d_id_j$ for $d_i,~d_j\gg 1$; this is because the  denominator of \eqref{probability} dominates its numerator for large degrees. On the other hand, for smaller $d_i$ and $d_j$, the numerator ensures relatively high probabilities for moderate-degree nodes. The combination of the two effects yields edge samples that are fairly representative of the graph, while inflicting low perturbation when removed.

\subsection{Parameter training}
Subsequently, for a given sample $\mathcal{S}$, we can obtain the corresponding optimal parameters as (cf. \eqref{ok2})
\begin{equation}\label{good}
\boldsymbol{\theta}^\ast_\mathcal{S} = \arg\min_{\boldsymbol{\theta}\in \mathcal{S}^K}\sum_{i,j\in \mathcal{S}}\ell\left( A_{i,j}, s_{\mathcal{G}^-}(v_i,v_j;\boldsymbol{\theta})  \right)
\end{equation}     
where $\mathcal{G}^- := \left( \mathcal{V}, \mathcal{E}\setminus\mathcal{S^+} \right)$ is the original graph with the randomly sampled subset  $\mathcal{S^+}$ of edges removed.

Interestingly, one way that \eqref{good} could be solved is by explicitly computing the entries of $\mathbf{S}_\mathcal{G}(\boldsymbol{\theta})$ that are in $\mathcal{S}$. This would require performing $K$ sparse matrix-vector products to obtain every column of $\mathbf{S}^k$ for $k\in \{1,\ldots,K\}$, for all the columns that contain sampled entries. In the worst case, if all nodes in the tuples of $\mathcal{S}$ correspond to different columns of $\mathbf{S}_\mathcal{G}(\boldsymbol{\theta})$, two random walks are required for every tuple, for a total of $2N_s$ random walks. This requires $\mathcal{O}\left(N_sK|\mathcal{E}|\right)$ computations, and $\mathcal{O}\left(N_sN\right)$ memory if they are to be performed concurrently or in matrix form. Since $K$ will typically be in the order of tens, these requirements will be affordable, if $N_s$ is relatively small. Nevertheless, they quickly become cumbersome for $N_s\gg K$, which may be necessary to estimate the $K$-dimensional $\boldsymbol{\theta}$.    

\begin{algorithm}[h!]
	\caption{\textsc{Adaptive Similarity Embedding}}
	\label{alg:ASE}
	\begin{algorithmic}
		\State \textbf{Input:} $\mathcal{G}$~~\textbf{Output:} $\mathbf{E}$
		\State
		\vspace{-0.1in}
		\State // Training phase
		\State $\boldsymbol{\Theta} =\emptyset$
		\While { $|\boldsymbol{\Theta}|<T_s$ }
		\State $\mathcal{G}^-$, $\mathcal{S}^+$, $\mathcal{S}^-~=$ \textsc{Sample Edges}( $\mathcal{G}$ ) 
		\State $\boldsymbol{\theta}^\ast_\mathcal{S}~=$ \textsc{Train Parameters}( $\mathcal{G}^-,\mathcal{S}^+, \mathcal{S}^-$)
		\State $\boldsymbol{\Theta} = \boldsymbol{\Theta} \cup \boldsymbol{\theta}^\ast_\mathcal{S}$
		\EndWhile
		\State $\boldsymbol{\theta}^\ast = T_s^{-1}\sum_{\boldsymbol{\theta}\in\boldsymbol{\Theta}}\boldsymbol{\theta}$
		\State
		\vspace{-0.1in}
		\State // Embedding phase
		\State $\mathbf{S} = \frac{1}{2}\left( \mathbf{I} + \mathbf{D}^{-1/2} \mathbf{A} \mathbf{D}^{-1/2} \right)$
		\vspace{0.03in}
		\State $\mathbf{S} = \mathbf{U}_d\boldsymbol{\Sigma}_d\mathbf{U}_d^T$ 
				\vspace{0.03in}
		\State $\boldsymbol{\Sigma}_d(\boldsymbol{\theta}^\ast) = \sum_{k=1}^K \theta_k^\ast \boldsymbol{\Sigma}_d^k$
		\State
		\vspace{-0.1in}
		\State \Return $\mathbf{E} = \mathbf{U}_d \sqrt{\boldsymbol{\Sigma}_d(\boldsymbol{\theta}^\ast)}$ 
	\end{algorithmic}
\end{algorithm} 
\begin{algorithm}[h!]
	\caption{\textsc{Sample Edges}}
	\label{alg:ES}
	\begin{algorithmic}
		\State \textbf{Input:} $\mathcal{G}$~~
		\textbf{Output:} $\mathcal{G}^-,\mathcal{S}^+, \mathcal{S}^-$
				\State
		\vspace{-0.1in}
		\State // Sample edges
		\State $\mathcal{S}^+=\emptyset$, $\mathcal{G}^- = \mathcal{G}$
		\While { $|\mathcal{S}^+|<N_s/2$ }
		\State Sample $v_1\sim\mathrm{Unif}\left(\mathcal{V}\right)$
		\If { $|\mathcal{N}_{\mathcal{G}^-}(v_1)|>1$ }
				\State Sample $v_2\sim\mathrm{Unif}\left(\mathcal{N}_{\mathcal{G}^-}(v_1)\right)$
				\If { $|\mathcal{N}_{\mathcal{G}^-}(v_2)|>1$ }
	  			\State $\mathcal{S}^+=\mathcal{S}^+ \cup (v_1,v_2)$ 
				\State $\mathcal{G^-} = \mathcal{G^-} \setminus (v_1,v_2)$
				\EndIf
		\EndIf
		\EndWhile						
				\State
		\vspace{-0.1in}
		\State // Sample non-edges
		\State $\mathcal{S}^-=\emptyset$
		\While { $|\mathcal{S}^-|<N_s/2$ }
		\State Sample $(v_1,v_2)\sim\mathrm{Unif}\left(\mathcal{V}\times\mathcal{V}\right)$
		\If {$(v_1,v_2)\notin \mathcal{E}$}
		\State $\mathcal{S}^-=\mathcal{S}^- \cup (v_1,v_2)$
		\EndIf
		\EndWhile						
		\State \Return $\mathcal{G^-}$, $\mathcal{S}^+$, $\mathcal{S}^-$
	\end{algorithmic}
\end{algorithm} 
\begin{algorithm}[h!]
	\caption{\textsc{Train Parameters}}
	\label{alg:TRAIN}
	\begin{algorithmic}
		\State \textbf{Input:} $\mathcal{G}$, $\mathcal{S}^+$, $\mathcal{S}^-$ \textbf{Output:}
		\vspace{0.03in} $\boldsymbol{\theta}^\ast_\mathcal{S}$
		\State $\mathbf{S} = \frac{1}{2}\left( \mathbf{I} + \mathbf{D}^{-1/2} \mathbf{A} \mathbf{D}^{-1/2} \right)$
\vspace{0.03in}
\State $\mathbf{S} = \mathbf{U}_d\boldsymbol{\Sigma}_d\mathbf{U}_d^T$
\vspace{0.03in}
\State $\mathcal{S} = \mathcal{S}^+\cup \mathcal{S}^-$ 
\State Form $\mathcal{X_S}=\{ \mathbf{x}_{(i,j)}\}_{(i,j)\in\mathcal{S}}$ as in \eqref{feats}
\vspace{0.03in}
		\vspace{0.03in}
\State \Return $\boldsymbol{\theta}^\ast_\mathcal{S}=~$\textsc{SimplexSVM}( $\mathcal{X_S},\mathcal{S}^+, \mathcal{S}^-$)
	\end{algorithmic}
\end{algorithm} 
\begin{algorithm}[h!]
	\caption{\textsc{SimplexSVM}}
	\label{alg:SVMs On Simplex}
	\begin{algorithmic}
		\State \textbf{Input:} $\mathcal{X},\mathcal{S}^+,\mathcal{S}^-$
		\textbf{Output:} $\boldsymbol{\theta}^\ast$
		\vspace{0.03in}
		\State $\boldsymbol{\theta}_0= \frac{1}{K} \mathbf{1},~t=1$
		\vspace{0.03in}
		\While { $\|\boldsymbol{\theta}_t - \boldsymbol{\theta}_{t-1}\|_\infty \geq \mathrm{tol}$ }
		\State $t=t+1,$ $\eta_t=a/\sqrt{t}$		
				\vspace{0.03in}
		\State $\mathcal{S}^+_a = \{e\in \mathcal{S}^+|~ \mathbf{x}_e^T\boldsymbol{\theta}_{t-1} \leq \epsilon \}$
				\vspace{0.03in}
\State $\mathcal{S}^-_a = \{e\in \mathcal{S}^-|~ \mathbf{x}_e^T\boldsymbol{\theta}_{t-1} \geq -\epsilon \}$
		\vspace{0.03in}
		\State $\mathbf{g}_t = \sum_{e\in\mathcal{S}^-_a}\mathbf{x}_e - \sum_{e\in\mathcal{S}^+_a}\mathbf{x}_e$
				\vspace{0.03in}
		\State $\mathbf{z}_t=(1-2\eta_t\lambda)\boldsymbol{\theta}_{t-1}-\frac{\eta_t}{N_s}\mathbf{g}_t$
		\vspace{0.03in}
				\vspace{0.03in}
		\State $\boldsymbol{\theta}_{t}=$\textsc{SimplexProj}( $\mathbf{z}_t$ )
		\EndWhile
		\State \Return $\boldsymbol{\theta}_{t}$
		\end{algorithmic}
\end{algorithm} 

Instead, we will rely on the fact that the proposed embeddings are smooth and differentiable wrt to $\boldsymbol{\theta}$ (cf. \eqref{solution2}), to develop a solution that allows for selecting arbitrarily large $N_s$, using the approximation
\begin{align}\nonumber
s_{\mathcal{G}^-}(v_i,v_j;\boldsymbol{\theta}) & \approx
s_{\mathcal{E}}(\mathbf{e}_i^-(\boldsymbol{\theta},\mathbf{e}_j^-(\boldsymbol{\theta})) \\ \nonumber
& = \left(\mathbf{e}_i^-(\boldsymbol{\theta})\right)^\top \mathbf{e}_j^-(\boldsymbol{\theta}) \\ \nonumber  
& = \left(\sqrt{\boldsymbol{\Sigma}^-_d(\boldsymbol{\theta})}~\mathbf{u}_i^-\right)^\top \sqrt{\boldsymbol{\Sigma}^-_d(\boldsymbol{\theta})}~\mathbf{u}_j^-\\ \nonumber
&  = \left( \mathbf{u}_i^-\right)^\top \boldsymbol{\Sigma}^-_d(\boldsymbol{\theta})\mathbf{u}_j^- \\ \label{approx}
& = \mathbf{x}_{i,j}^\top ~\boldsymbol{\theta}
\end{align}   
where 
\begin{equation}\label{feats}
\mathbf{x}_{i,j} = \left( \mathbf{u}_i^- \ast \mathbf{u}_j^- \right)^\top \boldsymbol{\Sigma}_d^K,
\end{equation}
and
\begin{equation*}
\boldsymbol{\Sigma}_d^K = \left[ \begin{array}{cccc}
\sigma_1~&~\sigma_1^2~&~\cdots~&~~\sigma_1^K \\
\vdots~&~\vdots~&~\ddots~&~\vdots \\
~~~\sigma_{d-1}~&~~~~\sigma_{d-1}^2~&~\cdots~&~~~\sigma_{d-1}^K\\
\sigma_d~&~\sigma_d^2~&~\cdots~&~\sigma_d^K
\end{array} \right]. 
\end{equation*}
Conveniently, $\{\mathbf{x}_{i,j}\}$s act as features over every possible pair of nodes, which when linearly combined with weights $\boldsymbol{\theta}$ to produce similarities, allow us to approach \eqref{good} using well-understood learning and optimization tools. Among the various loss functions one may fit the removed edges\footnote{In our implementation, we also provide learning mechanisms based on least-squares, logistic regression, as well as finding the best single $k$. Due to space constrains though we only  present and report results of the SVM-based approach.} using the hinge loss 
\begin{equation}
\ell(y,f) :=  \max(0,\epsilon - yf)
\end{equation}
which is suitable for real-world graphs thanks to its robustness properties \cite{svm}; note that target variables here are defined as $y_{i,j} = 2A_{i,j}-1$ so that $y_{i,j}\in \{-1,1\}$. We can then equivalently express \eqref{good} as
\begin{equation}\label{final}
\boldsymbol{\theta}^\ast_\mathcal{S} = \arg\min_{\boldsymbol{\theta}\in \mathcal{S}^K}\sum_{i,j\in \mathcal{S}}\max(0,\epsilon - y_{i,j}\mathbf{x}_{i,j}^\top ~\boldsymbol{\theta}) + \lambda \|\boldsymbol{\theta}\|_2^2
\end{equation} 
where $\lambda\geq0$ is the regularization parameter of the $\ell_2$ regularization typically used to improve the robustness and generalization capability of SVMs \cite{svm}. To solve our variant of simplex-constrained SVMs (cf. \eqref{final}), we employ the projected-gradient descent approach \cite{bertsekas} that we describe in Algorithm 4, where \textsc{SimplexProj}( $\cdot$ ) is a subroutine that implements projections onto $\mathcal{S}^K$; the latter can be performed with $\mathcal{O}(K\log K)$ complexity as described in \cite{simplex_proj}. The overall parameter learning procedure for a given sample is summarized in Algorithm 3.

In general, if runtime or computational resources allow, the sampling and training process described in the last two sections can be repeated $T_s$ times to obtain different $\{\boldsymbol{\theta}^\ast_\mathcal{S}\}$s, which can then be averaged in order to reduce their variance. In practice, this may not be necessary if $N_s$ is large enough, which will yield a near-deterministic $\boldsymbol{\theta}$. The overall proposed adaptive-similarity embedding (ASE) framework is summarized in Algorithm 1.

\subsection{Complexity} The computational complexity of ASE is dominated by the cost of performing the truncated SVD of $\mathbf{S}$ in the training as well as testing phases of Algorithm 1. Relying on the sparsity ($|\mathcal{E}|\ll N^2$) and symmetry of $\mathbf{S}$, the Lanczos algorithm followed by EVD of a tridiagonal matrix yield the truncated SVD in a very efficient manner. Provided that $d \ll N$, the decomposition can be achieved in $\mathcal{O}(|\mathcal{E}|d)$ time and using $\mathcal{O}(Nd)$ memory. Therefore, for the $T_s\geq 1$ training rounds and a single embedding round of Algorithm 1, the overall complexity is $\mathcal{O}((T_s + 1)|\mathcal{E}|d)$. 

\section{Related work}\label{sec:remarks}
Two recent embedding methods also pursue similarity matrices that combine walks of different lengths \cite{arope,attention}. Most relevant to the proposed ASE is the ``Arbitrary-Order Proximity Preserved Network Embedding'' \cite{arope} approach, where a method is proposed for obtaining the SVD of a polynomial of the adjacency matrix without having to recompute the singular vectors. 

Compared to \cite{arope}, we put forth the following contributions. First, we introduce a family of multihop similarities whose decomposition leads to embeddings that inherit the rich information contained in the spectral embeddings (cf. Section 2.3). An equally important contribution in terms of modeling is that our embeddings can be differentiated with respect to (wrt) weights $\boldsymbol{\theta}$ (cf. \eqref{approx}-\eqref{final}), whereas the embeddings in \cite{arope} are non-differentiable wrt the weights. Hence, \cite{arope} can only proceed in a ``forward'' fashion given some order proximity weights $\boldsymbol{\theta}$, whereas our approach allows for ``navigating'' the space of possible similarity functions $s(v_i,v_j;\boldsymbol{\theta})$ in a smooth fashion, meaning that $\boldsymbol{\theta}$ can be learned with simple optimization on well-defined fitting models such as logistic regression or SVMs (cf. \eqref{final}). This leads to the third main contribution, which is a means of learning ``personalized'' $\boldsymbol{\theta}$ (cf. Section 4) in an unsupervised fashion, meaning without downstream information such as node or edge labels/attributes that can guide cross-validation in high-dimensional discretized parameter grids. 

The second related embedding method presented in \cite{attention} builds on the concept of graph attention mechanisms to place weights on lengths of truncated random walks. These mechanisms are used to build a similarity matrix containing co-occurrence probabilities. The matrix is jointly decomposed by maximizing a graph-likelihood function. The model in \cite{attention} is a generalization of the ones implicitly adopted by \cite{deepwalk} and \cite{node2vec}, building on similar tools and concepts that emerge from natural language processing. Different from \cite{deepwalk, node2vec} and the proposed ASE, \cite{attention} explicitly constructs and factorizes a dense $N\times N$ similarity matrix. The detailed procedure incurs complexity that is \emph{cubic} wrt $N$, and becomes at best \emph{quadratic} after model approximations, meaning that \cite{attention} scales rather poorly beyond small graphs. 

\section{Experimental Evaluation}
\label{sec:experiments}

The present section reports extensive experimental results on a variety of real-world networks. The aim of the presented tests is twofold. First, to determine and quantify the quality of the proposed ASE embeddings for different downstream learning tasks. Second, to analyze and interpret the resulting embedding parameters for different networks. 

\noindent \textbf{Datasets.} In our experiments, we used the following real-world networks (see also Table 2). 

\begin{itemize}
\item \textbf{\texttt{ca-AstroPh}}. The Astro Physics collaboration network is from the e-print arXiv and covers scientific collaborations between co-authored papers submitted to Astro Physics category~\cite{snap}. If an author $i$ co-authored a paper with author $j$, the graph contains a undirected edge from $i$ to $j$. If the paper is co-authored by $k$ authors, this generates a completely connected (sub)graph on $k$ nodes.
	
\item \textbf{\texttt{ca-CondMat}}. Condense Matter Physics collaboration network from ArXiv \cite{snap}.

\item \textbf{\texttt{CoCit}}. A co-citation network of papers citing other papers extracted by \cite{VERSE}; labels represent conferences in which papers were published.
	
\item \textbf{\texttt{com-DBLP}}. Computer science research bibliography collaboration network \cite{snap}.

\item \textbf{\texttt{com-Amazon}}. Network collected by crawling Amazon website \cite{snap}. It is based on ``Customers Who Bought This Item Also Bought'' feature of the Amazon website. If a product $i$ is frequently co-purchased with product $j$, the graph contains an undirected edge from $i$ to $j$.

\item \textbf{\texttt{vk2016-17}}. VK is a Russian all-encompassing social network. In \cite{VERSE}, two snapshots of the network were extracted in November 2016 and May 2017, to obtain information about link appearance.
	
\item \textbf{\texttt{email-Enron}}. Enron email communication network covering all the email communication within a dataset of around half a million emails \cite{snap}. 

\item \textbf{\texttt{PPI (H.Sapiens)}}. Subgraph of the  protein-protein interaction  network  for  Homo  Sapiens. The subgraph corresponds to the graph induced by nodes for which labels (representing biological states) were obtained from the hallmark gene sets \cite{node2vec}. 
	
\item \textbf{\texttt{Wikipedia}}. This is a co-occurrence network of words
appearing in the first million bytes of the Wikipedia dump. The labels represent the Part-of-Speech (POS) tags inferred using the Stanford POS-Tagger \cite{node2vec}.
			
\item \textbf{\texttt{BlogCatalog}}. A network of social relationships of the bloggers listed on the BlogCatalog website. The labels represent blogger interests  inferred through the meta-data provided by the bloggers. 
\end{itemize}

\begin{table}[t]
	\centering
	\caption{ Network Characteristics }
	\rowcolors{2}{}{gray!7}
	\begin{tabular} {ccccc}         			
		\toprule
		Graph & $|\mathcal{V}|$ & $|\mathcal{E}|$ & $|\mathcal{Y}|$ & Density  \\
		\midrule 
				
		\texttt{PPI (H. Sapiens)} & 3,890 & 76,584 & 50 & $10^{-2}$ \\
		
		\texttt{Wikipedia} & 4,733 & 184,182 & 40 & $1.6\times10^{-2}$  \\

		\texttt{BlogCatalog} & 10,312 & 333,983 & 39 & $6.2\times10^{-3}$  \\
				
		\texttt{ca-CondMat} & 23,133 & 93,497 & - & $3.5\times10^{-4}$ \\

		\texttt{ca-AstroPh} &18,772 & 198,110 & - & $1.1\times10^{-3}$ \\

		\texttt{email-Enron} &36,692 & 183,831 & - & $2.7\times10^{-4}$ \\

		\texttt{CoCit} & 44,312 & 195,362 & 15 & $2\times10^{-4}$  \\

		\texttt{vk2016-17} & 78,593 & 2,680,542 & - & $8.7\times10^{-4}$ \\	
		
		\texttt{com-Amazon} & 334,863 & 925,872 & - & $1.7\times10^{-5}$ \\
		
		\texttt{com-DBLP} &317,080 & 1,049,866 & - & $2.1\times10^{-5}$ \\		
		
		\bottomrule
	\end{tabular}\label{tab:graphs}
\end{table}

   \begin{table*}[t]
	\centering
	\caption{ Inferred parameters and interpretation }
	\rowcolors{2}{}{gray!7}
	\begin{tabular} {ccccccccccccc}         			
		\toprule
		Graph & $\theta_1$ & $\theta_2$ & $\theta_3$ & $\theta_4$ & $\theta_5$ & $\theta_6$ & $\theta_7$ & $\theta_8$ & $\theta_9$ & $\theta_{10}$ & range & strength \\
		\midrule 
		
		\texttt{PPI (H. Sapiens)} & 0.00  &  \textbf{0.14}  & \textbf{0.31} & \textbf{0.29} & \textbf{0.21} & \textbf{0.04} & 0.00 & 0.00  & 0.00  & 0.00 & medium& medium \\
		
		\texttt{Wikipedia} & 0.00  & 0.00 & 0.00 & 0.00 & 0.00 & 0.00 & 0.00 & \textbf{0.01} & \textbf{0.37} & \textbf{0.62} & long & strong \\
		
		\texttt{BlogCatalog} & \textbf{1.00} & 0.00 & 0.00 & 0.00 & 0.00 & 0.00 & 0.00 & 0.00 & 0.00 & 0.00 & short & very strong \\
		
		\texttt{ca-CondMat} & \textbf{0.55} & \textbf{0.33} & \textbf{0.12} & 0.00 & 0.00 & 0.00 & 0.00 & 0.00 & 0.00 & 0.00 & short & strong \\
		
		\texttt{ca-AstroPh} & \textbf{0.76} & \textbf{0.24} & 0.00 & 0.00 & 0.00 & 0.00 & 0.00 & 0.00 & 0.00 & 0.00 & short & strong \\
		
		\texttt{email-Enron} & \textbf{0.24} & \textbf{0.25} & \textbf{0.18} & \textbf{0.14} & \textbf{0.1} & \textbf{0.06} & \textbf{0.02} & 0.00 & 0.00 & 0.00 & medium & weak \\
		
		\texttt{CoCit} & \textbf{0.61} & \textbf{0.33} & \textbf{0.06} & 0.00 & 0.00 & 0.00 & 0.00 & 0.00 & 0.00 & 0.00 & short & strong \\		
		
		\texttt{vk2016-17} & \textbf{0.71} & \textbf{0.29} & 0.00 & 0.00 & 0.00 & 0.00 & 0.00 & 0.00 & 0.00 & 0.00 & short & strong\\	
		
		\texttt{com-Amazon} & \textbf{0.10} & \textbf{0.10} & \textbf{0.10} & \textbf{0.10} & \textbf{0.09} & \textbf{0.09} & \textbf{0.09} & \textbf{0.09} & \textbf{0.09} & \textbf{0.09} & short & very weak \\
		
		\texttt{com-DBLP} & \textbf{0.11} & \textbf{0.10} & \textbf{0.10} & \textbf{0.09} & \textbf{0.09} & \textbf{0.09} & \textbf{0.09} & \textbf{0.09} & \textbf{0.09} & \textbf{0.08} & short & very weak \\		
		
		\bottomrule
	\end{tabular}\label{tab:thetas}
\end{table*}

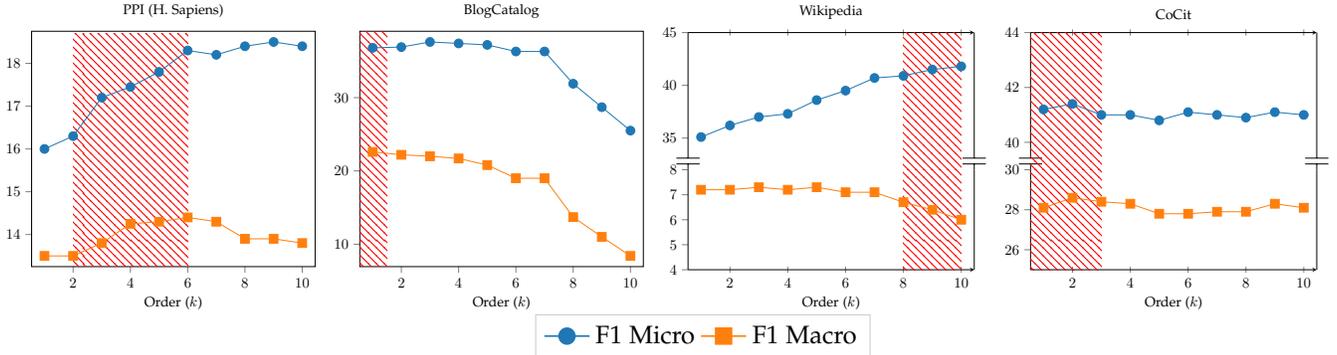
\begin{figure*}[t!]  
	\centering
\begin{tikzpicture}[scale=0.55]

\definecolor{color1}{rgb}{1,0.498039215686275,0.0549019607843137}
\definecolor{color0}{rgb}{0.12156862745098,0.466666666666667,0.705882352941177}

\begin{axis}[
legend cell align={left},
legend entries={{F1 Micro},{F1 Macro}},
legend style={at={(0.03,0.97)}, anchor=north west, draw=white!80.0!black},
legend columns = 2,
legend to name = {named5},
tick align=outside,
tick pos=left,
title={PPI (H. Sapiens)},
x grid style={lightgray!92.02614379084967!black},
xlabel={Order ($k$)},
xmin=0.55, xmax=10.45,
y grid style={lightgray!92.02614379084967!black},
ymin=13.25, ymax=18.75
]

\draw[pattern=north west lines, pattern color=red, draw = none] (2,13.25) rectangle (6,18.7);

\addplot [semithick, color0, mark=*, mark size=3, mark options={solid}]
table [row sep=\\]{%
1	16 \\
2	16.3 \\
3	17.2 \\
4	17.45 \\
5	17.8 \\
6	18.3 \\
7	18.2 \\
8	18.4 \\
9	18.5 \\
10	18.4 \\
};
\addplot [semithick, color1, mark=square*, mark size=3, mark options={solid}]
table [row sep=\\]{%
1	13.5 \\
2	13.5 \\
3	13.8 \\
4	14.25 \\
5	14.3 \\
6	14.4 \\
7	14.3 \\
8	13.9 \\
9	13.9 \\
10	13.8 \\
};

%

\end{axis}

\end{tikzpicture}	
\begin{tikzpicture}[scale=0.55]

\definecolor{color1}{rgb}{1,0.498039215686275,0.0549019607843137}
\definecolor{color0}{rgb}{0.12156862745098,0.466666666666667,0.705882352941177}

\begin{axis}[
tick align=outside,
tick pos=left,
title={BlogCatalog},
x grid style={lightgray!92.02614379084967!black},
xlabel={Order ($k$)},
xmin=0.55, xmax=10.45,
y grid style={lightgray!92.02614379084967!black},
ymin=6.94, ymax=39.06
]

\draw[pattern=north west lines, pattern color=red, draw = none] (0.5,6.94) rectangle (1.5,39.0);

\addplot [semithick, color0, mark=*, mark size=3, mark options={solid}]
table [row sep=\\]{%
1	36.8 \\
2	36.9 \\
3	37.6 \\
4	37.4 \\
5	37.2 \\
6	36.3 \\
7	36.3 \\
8	31.9 \\
9	28.7 \\
10	25.5 \\
};
\addplot [semithick, color1, mark=square*, mark size=3, mark options={solid}]
table [row sep=\\]{%
1	22.6 \\
2	22.2 \\
3	22 \\
4	21.7 \\
5	20.8 \\
6	19 \\
7	19 \\
8	13.7 \\
9	11 \\
10	8.4 \\
};
\end{axis}

\end{tikzpicture}
\begin{tikzpicture}[scale=0.55]

\definecolor{color1}{rgb}{1,0.498039215686275,0.0549019607843137}
\definecolor{color0}{rgb}{0.12156862745098,0.466666666666667,0.705882352941177}

\begin{groupplot}[
group style={
	group name=my fancy plots,
	group size=1 by 2,
	xticklabels at=edge bottom,
	vertical sep=0pt
},
tick align=outside,
tick pos=left,
x grid style={lightgray!92.02614379084967!black},
xmin=0.55, xmax=10.45,
y grid style={lightgray!92.02614379084967!black},
]


\nextgroupplot[ymin=32.0,ymax=45.0,
title={Wikipedia},
axis x line=top, 
axis y discontinuity=parallel,
width=8.50cm,
height=4.9cm]
\draw[pattern=north west lines, pattern color=red, draw = none] (8.0,32.0) rectangle (10.0,45.0);
\addplot [semithick, color0, mark=*, mark size=3, mark options={solid}]
table [row sep=\\]{%
1	35.1 \\
2	36.2 \\
3	37 \\
4	37.3 \\
5	38.6 \\
6	39.5 \\
7	40.7 \\
8	40.9 \\
9	41.5 \\
10	41.8 \\
};   

\nextgroupplot[ymin=4.0,ymax=8.0,
axis x line=bottom,
xlabel={Order ($k$)},
width=8.50cm,
height=4.0cm]
\draw[pattern=north west lines, pattern color=red, draw = none] (8.0,4.0) rectangle (10.0,8.0);

\addplot [semithick, color1, mark=square*, mark size=3, mark options={solid}]
table [row sep=\\]{%
1	7.2 \\
2	7.2 \\
3	7.3 \\
4	7.2 \\
5	7.3 \\
6	7.1 \\
7	7.1 \\
8	6.7 \\
9	6.4 \\
10	6 \\
};     
\end{groupplot}

\end{tikzpicture}		
\begin{tikzpicture}[scale=0.55]

\definecolor{color1}{rgb}{1,0.498039215686275,0.0549019607843137}
\definecolor{color0}{rgb}{0.12156862745098,0.466666666666667,0.705882352941177}

\begin{groupplot}[
group style={
	group name=my fancy plots,
	group size=1 by 2,
	xticklabels at=edge bottom,
	vertical sep=0pt
},
tick align=outside,
tick pos=left,
x grid style={lightgray!92.02614379084967!black},
xmin=0.55, xmax=10.45,
y grid style={lightgray!92.02614379084967!black},
]


\nextgroupplot[ymin=39.0,ymax=44.0,
title={CoCit},
axis x line=top, 
axis y discontinuity=parallel,
width=8.50cm,
height=4.9cm]
\draw[pattern=north west lines, pattern color=red, draw = none] (0.55,32.0) rectangle (3.0,45.0);
\addplot [semithick, color0, mark=*, mark size=3, mark options={solid}]
table [row sep=\\]{%
1	41.2 \\
2	41.4 \\
3	41.0 \\
4	41.0 \\
5	40.8 \\
6	41.1 \\
7	41.0 \\
8	40.9 \\
9	41.1 \\
10	41.0 \\
};   

\nextgroupplot[ymin=25.0,ymax=30.0,
axis x line=bottom,
xlabel={Order ($k$)},
width=8.50cm,
height=4.0cm]
\draw[pattern=north west lines, pattern color=red, draw = none] (0.55,25.0) rectangle (3.0,30.0);

\addplot [semithick, color1, mark=square*, mark size=3, mark options={solid}]
table [row sep=\\]{%
1	28.1 \\
2	28.6 \\
3	28.4 \\
4	28.3 \\
5	27.8 \\
6	27.8 \\
7	27.9 \\
8	27.9 \\
9	28.3 \\
10	28.1 \\
};     
\end{groupplot}

\end{tikzpicture}
	\phantom{ppppp}\ref{named5}			
	\caption{ Micro and Macro $F_1$ scores for the four labeled graphs, when the ``pure'' $k-$order $\mathbf{S}^k$ is used for embedding, given as a function of $k$. Red shade denotes the corresponding $k$'s where ASE assigned non-zero $\boldsymbol{\theta}_k$'s; see also Table 2. }  \label{fig:single}
\end{figure*}

\noindent \textbf{Methods.} Experiments were run using the following \emph{unsupervised} and \emph{scalable} embedding methods.

\begin{itemize}
	
\item \textbf{ASE}. Our proposed adaptive similarity embedding. Based on observations made in Sections 3, and to retain optimization stability, we set the maximum number of steps to $K=10$. We also use the default SVM regularizer ($\lambda=1$). To have a single learning round with learned parameters having small enough variance, we sampled with $N_s/2 = 1,000$. We made our implementation of ASE freely available~\footnote{https://github.com/DimBer/ASE-project}.  
	
\item \textbf{VERSE} \cite{VERSE}. This is a scalable framework for generating node embeddings according to a similarity function by minimizing a KL-divergence-objective via stochastic optimization. We used the default version with similarity (PPR with $\alpha=0.85$), as suggested and implemented by the authors.\footnote{https://github.com/xgfs/verse}
	
\item \textbf{Deepwalk} \cite{deepwalk}. This approach learns an embedding by sampling random walks from each node, and applying word2vec-based learning on those walks. We use the default parameters proposed in \cite{deepwalk}, i.e., walk length $t= 80$, number of walks per node $\gamma= 80$, window size $w= 10$, and the scalable C++ implementation\footnote{https://github.com/xgfs/deepwalk-c} provided in \cite{VERSE}.
	
\item \textbf{HOPE} \cite{hope}. This SVD-based approach approximates high-order proximities and leverages directed edges. We report the results	obtained with the default parameters, i.e, Katz similarity as the similarity measure with $\beta$ inversely proportional to the spectral radius. 

\item \textbf{AROPE} \cite{arope}. An approach for fast computation of thin SVD of different polynomials of $\mathbf{A}$. We used the official Python implementation \footnote{\url{https://github.com/ZW-ZHANG/AROPE}} to produce the embeddings. We selected the polynomial (hyper) parameters of AROPE using a set of validation edges that was sampled similarily to ASE (Algorithm 2). We consider proximity orders in the range $[1,10]$, and perform grid search over the different proximity weights as suggested in \cite{arope}.
	
\item \textbf{LINE} \cite{LINE}. This approach learns a $d$-dimensional embedding
in two steps, both using adjacency similarity. First, it learns $d/2$ dimensions using first-order proximity; then, it learns another $d/2$ features using second-order proximity.  Last, the two halves are normalized and concatenated. We obtained a copy of the code\footnote{https://github.com/tangjianpku/LINE}, and
run experiments with $T= 10^{10}$ samples (although $T= 10^9$ yielded the same accuracy for smaller graphs), and $s = 5$ negative samples, as described in the paper.

\item \textbf{Spectral}. This approach relies on the first $d$ eigenvectors of $\mathbf{D}^{-1/2}\mathbf{A}\mathbf{D}^{-1/2}$. The baseline was developed for clustering \cite{spectral}, and has also been run as a benchmark for node embeddings \cite{node2vec}. In our case, spectral embedding is of particular interest since it can be obtained by column-wise normalization of the embeddings generated by the proposed method.
\end{itemize}

We excluded comparisons with Node2vec \cite{node2vec} because they use cross-validation on node labels for hyper-parameter selection. Thus comparing Node2vec to methods such as LINE, Deepwalk, HOPE, VERSE, and EMB that all operate with \emph{fixed} hyperparameters in a fully \emph{unsupervised} manner would be unfair. We also excluded comparisons with GraRep \cite{grarep} and M-NMF \cite{netmf} due to their limited scalability ($\mathcal{O}(N^2d)$ computational and $\mathcal{O}(N^2)$ memory complexity). \\

\noindent \textbf{Evaluation methodology}. Our experiment setting follows the one in \cite{VERSE}. All methods are set to embed nodes to dimension $d=100$. Using the resulting embeddings as feature vectors, we evaluated their performance in terms of node classification and link prediction accuracy, and clustering quality. All experiments were repeated 10 times and reported are the averaged results. \\

\begin{figure*}[t!]  
	\vspace{0.2in}
	\centering
\begin{tikzpicture}[scale=0.55]

\definecolor{color3}{rgb}{0.83921568627451,0.152941176470588,0.156862745098039}
\definecolor{color1}{rgb}{1,0.498039215686275,0.0549019607843137}
\definecolor{color2}{rgb}{0.172549019607843,0.627450980392157,0.172549019607843}
\definecolor{color4}{rgb}{0.580392156862745,0.403921568627451,0.741176470588235}
\definecolor{color0}{rgb}{0.12156862745098,0.466666666666667,0.705882352941177}
\definecolor{color5}{rgb}{0.549019607843137,0.337254901960784,0.294117647058824}
\definecolor{color6}{rgb}{0.749019607843137,0.137254901960784,0.594117647058824}

\begin{axis}[
legend cell align={left},
legend entries={{ASE},{VERSE},{Deepwalk},{LINE},{HOPE},{AROPE},{Spectral}},
legend style={at={(0.97,0.03)}, anchor=south east, draw=white!80.0!black},
legend columns = 7,
legend to name = {named3},
tick align=outside,
tick pos=left,
x grid style={lightgray!92.02614379084967!black},
xlabel={Label rate},
ylabel = {F1 Micro},
xmin=0.0, xmax=0.22,
y grid style={lightgray!92.02614379084967!black},
ymin=0.1000085170935933847, ymax=0.203633448732827,
title =  { PPI (H. Sapiens)  }
]
\addplot [line width=2.0pt, color0]
table [row sep=\\]{%
0.02	0.126 \\
0.04	0.145 \\
0.06	0.158 \\
0.08	0.168 \\
0.1	    0.175 \\
0.12	0.179 \\
0.14	0.182 \\
0.16    0.187 \\
0.18	0.193 \\
0.2	    0.198 \\
};
\addplot [line width=1.0pt, color1, mark=*, mark size=2, mark options={solid}]
table [row sep=\\]{%
0.02	0.112 \\
0.04	0.132 \\
0.06	0.147 \\
0.08	0.154 \\
0.1	    0.16 \\
0.12	0.164 \\
0.14	0.172 \\
0.16    0.177 \\
0.18	0.177 \\
0.2	    0.181 \\
};
\addplot [line width=1.0pt, color2, mark=+, mark size=3, mark options={solid}]
table [row sep=\\]{%
0.02	0.116 \\
0.04	0.141 \\
0.06	0.151 \\
0.08	0.16 \\
0.1	    0.167 \\
0.12	0.169 \\
0.14	0.174 \\
0.16    0.182 \\
0.18	0.184 \\
0.2	    0.187 \\
};
\addplot [line width=1.0pt, color3, mark=diamond, mark size=3, mark options={solid}]
table [row sep=\\]{%
0.02	0.112 \\
0.04	0.137 \\
0.06	0.154 \\
0.08	0.16 \\
0.1	    0.166 \\
0.12	0.176 \\
0.14	0.18 \\
0.16    0.186 \\
0.18	0.191 \\
0.2	    0.195 \\
};
\addplot [line width=1.0pt, color4, mark=triangle*, mark size=3, mark options={solid,rotate=90}]
table [row sep=\\]{%
0.02	0.101 \\
0.04	0.104 \\
0.06	0.116 \\
0.08	0.124 \\
0.1	    0.126 \\
0.12	0.134 \\
0.14	0.14 \\
0.16    0.142 \\
0.18	0.143 \\
0.2	    0.146 \\
};

\addplot [line width=1.0pt, color6, mark=square, mark size=3, mark options={solid,rotate=90}]
table [row sep=\\]{%
	0.02 0.09 \\
	0.04 0.11 \\
	0.06 0.115 \\
	0.08 0.13 \\
	0.10 0.132 \\
	0.12 0.14 \\
	0.14 0.142 \\
	0.16 0.149 \\
	0.18 0.152 \\
	0.20 0.157 \\
};

\addplot [line width=1.0pt,dashed, color5, mark=dot, mark size=3, mark options={solid,rotate=90}]
table [row sep=\\]{%
	0.02 0.038 \\
	0.04 0.035 \\
	0.06 0.034 \\
	0.08 0.033 \\
	0.10 0.034 \\
	0.12 0.033 \\
	0.14 0.033 \\
	0.16 0.034 \\
	0.18 0.033 \\
	0.20 0.033 \\
};

\end{axis}

\end{tikzpicture}
\begin{tikzpicture}[scale=0.55]

\definecolor{color3}{rgb}{0.83921568627451,0.152941176470588,0.156862745098039}
\definecolor{color1}{rgb}{1,0.498039215686275,0.0549019607843137}
\definecolor{color2}{rgb}{0.172549019607843,0.627450980392157,0.172549019607843}
\definecolor{color4}{rgb}{0.580392156862745,0.403921568627451,0.741176470588235}
\definecolor{color0}{rgb}{0.12156862745098,0.466666666666667,0.705882352941177}
\definecolor{color5}{rgb}{0.549019607843137,0.337254901960784,0.294117647058824}

\begin{axis}[
tick align=outside,
tick pos=left,
x grid style={lightgray!92.02614379084967!black},
xlabel={Label rate},
xmin=0.0, xmax=0.22,
y grid style={lightgray!92.02614379084967!black},
ymin=0.27, ymax=0.395,
title =  { BlogCatalog  }
]
\addplot [line width=2.0pt, color0]
table [row sep=\\]{%
	0.02	0.299 \\
	0.04	0.331 \\
	0.06	0.346 \\
	0.08	0.358 \\
	0.1	    0.368 \\
	0.12	0.373 \\
	0.14	0.38 \\
	0.16    0.384 \\
	0.18	0.387 \\
	0.2	    0.389 \\
};
\addplot [line width=1.0pt, color1, mark=*, mark size=2, mark options={solid}]
table [row sep=\\]{%
	0.02	0.277 \\
	0.04	0.311 \\
	0.06	0.329\\
	0.08	0.337 \\
	0.1	    0.348 \\
	0.12	0.359 \\
	0.14	0.366 \\
	0.16    0.37 \\
	0.18	0.372 \\
	0.2	    0.376 \\
};
\addplot [line width=1.0pt, color2, mark=+, mark size=3, mark options={solid}]
table [row sep=\\]{%
	0.02	0.297 \\
	0.04	0.331 \\
	0.06	0.35 \\
	0.08	0.361 \\
	0.1	    0.364 \\
	0.12	0.373 \\
	0.14	0.383 \\
	0.16    0.384 \\
	0.18	0.386 \\
	0.2	    0.389 \\
};
\addplot [line width=1.0pt, color3, mark=diamond, mark size=3, mark options={solid}]
table [row sep=\\]{%
	0.02	0.292 \\
0.04	0.324 \\
0.06	0.344 \\
0.08	0.355 \\
0.1	    0.36 \\
0.12	0.366 \\
0.14	0.371 \\
0.16    0.376 \\
0.18	0.378 \\
0.2	    0.38 \\		
};
\addplot [line width=1.0pt, color4, mark=triangle*, mark size=3, mark options={solid,rotate=90}]
table [row sep=\\]{%
	0.02	0.126 \\
0.04	0.145 \\
0.06	0.158 \\
0.08	0.168 \\
0.1	    0.175 \\
0.12	0.179 \\
0.14	0.182 \\
0.16    0.187 \\
0.18	0.193 \\
0.2	    0.198 \\
};
\end{axis}

\end{tikzpicture}
\begin{tikzpicture}[scale=0.55]

\definecolor{color3}{rgb}{0.83921568627451,0.152941176470588,0.156862745098039}
\definecolor{color1}{rgb}{1,0.498039215686275,0.0549019607843137}
\definecolor{color2}{rgb}{0.172549019607843,0.627450980392157,0.172549019607843}
\definecolor{color4}{rgb}{0.580392156862745,0.403921568627451,0.741176470588235}
\definecolor{color0}{rgb}{0.12156862745098,0.466666666666667,0.705882352941177}
\definecolor{color5}{rgb}{0.549019607843137,0.337254901960784,0.294117647058824}

\begin{axis}[
tick align=outside,
tick pos=left,
x grid style={lightgray!92.02614379084967!black},
xlabel={Label rate},
xmin=0.0, xmax=0.22,
y grid style={lightgray!92.02614379084967!black},
ymin=0.335170935933847, ymax=0.4570633448732827,
title =  { Wikipedia }
]
\addplot [line width=2.0pt, color0]
table [row sep=\\]{%
	0.02	0.403 \\
	0.04	0.4053 \\
	0.06	0.411 \\
	0.08	0.413 \\
	0.1	    0.413 \\
	0.12	0.421 \\
	0.14	0.425 \\
	0.16    0.426 \\
	0.18	0.427 \\
	0.2	    0.4285 \\
};
\addplot [line width=1.0pt, color1, mark=*, mark size=2, mark options={solid}]
table [row sep=\\]{%
	0.02	0.35 \\
	0.04	0.341 \\
	0.06	0.342 \\
	0.08	0.343 \\
	0.1	    0.342 \\
	0.12	0.349 \\
	0.14	0.355 \\
	0.16    0.361 \\
	0.18	0.365 \\
	0.2	    0.364 \\
};
\addplot [line width=1.0pt, color2, mark=+, mark size=3, mark options={solid}]
table [row sep=\\]{%
	0.02	0.35 \\
	0.04	0.346 \\
	0.06	0.341 \\
	0.08	0.343 \\
	0.1	    0.348 \\
	0.12	0.355 \\
	0.14	0.356 \\
	0.16    0.362 \\
	0.18	0.363 \\
	0.2	    0.363 \\
};
\addplot [line width=1.0pt, color3, mark=diamond, mark size=3, mark options={solid}]
table [row sep=\\]{%
	0.02	0.39 \\
	0.04	0.385 \\
	0.06	0.393 \\
	0.08	0.388 \\
	0.1	    0.39 \\
	0.12	0.389 \\
	0.14	0.394 \\
	0.16    0.392 \\
	0.18	0.395 \\
	0.2	    0.396 \\
};
\addplot [line width=1.0pt, color4, mark=triangle*, mark size=3, mark options={solid,rotate=90}]
table [row sep=\\]{%
	0.02	0.408 \\
	0.04	0.417 \\
	0.06	0.4256 \\
	0.08	0.4286 \\
	0.1	    0.433 \\
	0.12	0.434 \\
	0.14	0.436 \\
	0.16    0.441 \\
	0.18	0.442 \\
	0.2	    0.445 \\
};

\addplot [line width=1.0pt, color6, mark=square, mark size=3, mark options={solid,rotate=90}]
table [row sep=\\]{%
	0.02	0.41 \\
	0.04	0.412 \\
	0.06	0.416 \\
	0.08	0.422 \\
	0.1	    0.426 \\
	0.12	0.43 \\
	0.14	0.434 \\
	0.16    0.438 \\
	0.18	0.441 \\
	0.2	    0.444 \\
};

\addplot [line width=1.0pt,dashed, color5, mark=dot, mark size=3, mark options={solid,rotate=90}]
table [row sep=\\]{%
	0.02 0.415 \\
	0.04 0.406 \\
	0.06 0.405 \\
	0.08 0.402 \\
	0.10 0.404 \\
	0.12 0.407 \\
	0.14 0.407 \\
	0.16 0.405 \\
	0.18 0.406 \\
	0.20 0.408 \\
};

\end{axis}

\end{tikzpicture}
\begin{tikzpicture}[scale=0.55]

\definecolor{color3}{rgb}{0.83921568627451,0.152941176470588,0.156862745098039}
\definecolor{color1}{rgb}{1,0.498039215686275,0.0549019607843137}
\definecolor{color2}{rgb}{0.172549019607843,0.627450980392157,0.172549019607843}
\definecolor{color4}{rgb}{0.580392156862745,0.403921568627451,0.741176470588235}
\definecolor{color0}{rgb}{0.12156862745098,0.466666666666667,0.705882352941177}
\definecolor{color5}{rgb}{0.549019607843137,0.337254901960784,0.294117647058824}

\begin{axis}[
tick align=outside,
tick pos=left,
x grid style={lightgray!92.02614379084967!black},
xlabel={Label rate},
xmin=0.0, xmax=0.055,
y grid style={lightgray!92.02614379084967!black},
ymin=0.34, ymax=0.44,
title =  { CoCit  }
]
\addplot [line width=2.0pt, color0]
table [row sep=\\]{%
0.005  0.378 \\
0.01  0.395 \\
0.02   0.414 \\
0.03   0.419 \\
0.04   0.428 \\
0.05   0.434 \\
};
\addplot [line width=1.0pt, color1, mark=*, mark size=2, mark options={solid}]
table [row sep=\\]{%
0.005  0.369 \\
0.01   0.387 \\
0.02   0.404 \\
0.03   0.413 \\
0.04  0.42 \\
0.05  0.427  \\
};
\addplot [line width=1.0pt, color2, mark=+, mark size=3, mark options={solid}]
table [row sep=\\]{%
0.005  0.356 \\
0.01   0.377 \\
0.02  0.395 \\
0.03  0.406 \\
0.04  0.414 \\
0.05  0.419 \\
};
\addplot [line width=1.0pt, color3, mark=diamond, mark size=3, mark options={solid}]
table [row sep=\\]{%
0.005 0.342 \\
0.01  0.372 \\
0.02  0.396 \\
0.03  0.408 \\
0.04  0.412 \\
0.05  0.415 \\
};
\addplot [line width=1.0pt, color4, mark=triangle*, mark size=3, mark options={solid,rotate=90}]
table [row sep=\\]{%
0.005 0. \\
0.01  0. \\
0.02  0. \\
0.03  0. \\
0.04  0. \\
0.05  0. \\
};

\addplot [line width=1.0pt,dashed, color5, mark=dot, mark size=3, mark options={solid,rotate=90}]
table [row sep=\\]{%
0.005 0. \\
0.01  0. \\
0.02  0. \\
0.03  0. \\
0.04  0. \\
0.05  0. \\
};

\end{axis}

\end{tikzpicture}
\begin{tikzpicture}[scale=0.55]

\definecolor{color3}{rgb}{0.83921568627451,0.152941176470588,0.156862745098039}
\definecolor{color1}{rgb}{1,0.498039215686275,0.0549019607843137}
\definecolor{color2}{rgb}{0.172549019607843,0.627450980392157,0.172549019607843}
\definecolor{color4}{rgb}{0.580392156862745,0.403921568627451,0.741176470588235}
\definecolor{color0}{rgb}{0.12156862745098,0.466666666666667,0.705882352941177}
\definecolor{color5}{rgb}{0.549019607843137,0.337254901960784,0.294117647058824}

\begin{axis}[
tick align=outside,
tick pos=left,
x grid style={lightgray!92.02614379084967!black},
xlabel={Label rate},
ylabel = {F1 Macro},
xmin=0.0, xmax=0.22,
y grid style={lightgray!92.02614379084967!black},
ymin=0.05170935933847, ymax=0.1709633448732827,
]
\addplot [line width=2.0pt, color0]
table [row sep=\\]{%
	0.02	0.084 \\
	0.04	0.108 \\
	0.06	0.123 \\
	0.08	0.133 \\
	0.1	    0.142 \\
	0.12	0.146 \\
	0.14	0.149 \\
	0.16    0.154 \\
	0.18	0.16 \\
	0.2	    0.166 \\
};
\addplot [line width=1.0pt, color1, mark=*, mark size=2, mark options={solid}]
table [row sep=\\]{%
	0.02	0.077 \\
0.04	0.099 \\
0.06	0.112 \\
0.08	0.124 \\
0.1	    0.13 \\
0.12	0.135 \\
0.14	0.141 \\
0.16    0.145 \\
0.18	0.149 \\
0.2	    0.151 \\
};
\addplot [line width=1.0pt, color2, mark=+, mark size=3, mark options={solid}]
table [row sep=\\]{%
	0.02	0.08 \\
0.04	0.103 \\
0.06	0.117 \\
0.08	0.127 \\
0.1	    0.136 \\
0.12	0.1375 \\
0.14	0.145 \\
0.16    0.149 \\
0.18	0.153 \\
0.2	    0.156 \\
};
\addplot [line width=1.0pt, color3, mark=diamond, mark size=3, mark options={solid}]
table [row sep=\\]{%
	0.02	0.066 \\
0.04	0.086 \\
0.06	0.103 \\
0.08	0.109 \\
0.1	    0.116 \\
0.12	0.126 \\
0.14	0.129 \\
0.16    0.137 \\
0.18	0.143 \\
0.2	    0.144 \\
};

\addplot [line width=1.0pt, color6, mark=square, mark size=3, mark options={solid,rotate=90}]
table [row sep=\\]{%
	0.02 0.05 \\
	0.04 0.065 \\
	0.06 0.069 \\
	0.08 0.07 \\
	0.10 0.085 \\
	0.12 0.09 \\
	0.14 0.1 \\
	0.16 0.102 \\
	0.18 0.11 \\
	0.20 0.115 \\
};

\addplot [line width=1.0pt, color4, mark=triangle*, mark size=3, mark options={solid,rotate=90}]
table [row sep=\\]{%
	0.02	0.057 \\
0.04	0.068 \\
0.06	0.077 \\
0.08	0.083 \\
0.1	    0.091 \\
0.12	0.095 \\
0.14	0.1 \\
0.16    0.099 \\
0.18	0.104 \\
0.2	    0.105 \\
};
\end{axis}

\end{tikzpicture}
\begin{tikzpicture}[scale=0.55]

\definecolor{color3}{rgb}{0.83921568627451,0.152941176470588,0.156862745098039}
\definecolor{color1}{rgb}{1,0.498039215686275,0.0549019607843137}
\definecolor{color2}{rgb}{0.172549019607843,0.627450980392157,0.172549019607843}
\definecolor{color4}{rgb}{0.580392156862745,0.403921568627451,0.741176470588235}
\definecolor{color0}{rgb}{0.12156862745098,0.466666666666667,0.705882352941177}
\definecolor{color5}{rgb}{0.549019607843137,0.337254901960784,0.294117647058824}

\begin{axis}[
tick align=outside,
tick pos=left,
x grid style={lightgray!92.02614379084967!black},
xlabel={Label rate},
xmin=0.0, xmax=0.22,
y grid style={lightgray!92.02614379084967!black},
ymin=0.11, ymax=0.242,
]
\addplot [line width=2.0pt, color0]
table [row sep=\\]{%
	0.02	0.174 \\
	0.04	0.196 \\
	0.06	0.21 \\
	0.08	0.215 \\
	0.1	    0.224 \\
	0.12	0.228 \\
	0.14	0.231 \\
	0.16    0.237 \\
	0.18	0.238 \\
	0.2	    0.24 \\
};
\addplot [line width=1.0pt, color1, mark=*, mark size=2, mark options={solid}]
table [row sep=\\]{%
	0.02	0.129 \\
	0.04	0.161 \\
	0.06	0.184 \\
	0.08	0.192 \\
	0.1	    0.207 \\
	0.12	0.216 \\
	0.14	0.22 \\
	0.16    0.225 \\
	0.18	0.229 \\
	0.2	    0.235 \\
};
\addplot [line width=1.0pt, color2, mark=+, mark size=3, mark options={solid}]
table [row sep=\\]{%
	0.02	0.14 \\
	0.04	0.17 \\
	0.06	0.193 \\
	0.08	0.202 \\
	0.1	    0.213 \\
	0.12	0.221 \\
	0.14	0.227 \\
	0.16    0.23 \\
	0.18	0.234 \\
	0.2	    0.238 \\
};
\addplot [line width=1.0pt, color3, mark=diamond, mark size=3, mark options={solid}]
table [row sep=\\]{%
0.02	0.118 \\
0.04	0.143 \\
0.06	0.162 \\
0.08	0.176 \\
0.1	    0.179 \\
0.12	0.185 \\
0.14	0.192 \\
0.16    0.197 \\
0.18	0.201 \\
0.2	    0.204 \\
};
\addplot [line width=1.0pt, color4, mark=triangle*, mark size=3, mark options={solid,rotate=90}]
table [row sep=\\]{%
	0.02	0.025 \\
0.04	0.026 \\
0.06	0.056 \\
0.08	0.09 \\
0.1	    0.096 \\
0.12	0.097 \\
0.14	0.1 \\
0.16    0.101 \\
0.18	0.103 \\
0.2	    0.108 \\	
};

\end{axis}

\end{tikzpicture}
	\hspace{-0.1cm}
\begin{tikzpicture}[scale=0.55]

\definecolor{color3}{rgb}{0.83921568627451,0.152941176470588,0.156862745098039}
\definecolor{color1}{rgb}{1,0.498039215686275,0.0549019607843137}
\definecolor{color2}{rgb}{0.172549019607843,0.627450980392157,0.172549019607843}
\definecolor{color4}{rgb}{0.580392156862745,0.403921568627451,0.741176470588235}
\definecolor{color0}{rgb}{0.12156862745098,0.466666666666667,0.705882352941177}
\definecolor{color5}{rgb}{0.549019607843137,0.337254901960784,0.294117647058824}

\begin{axis}[
tick align=outside,
tick pos=left,
x grid style={lightgray!92.02614379084967!black},
xlabel={Label rate},
xmin=0.0, xmax=0.22,
y grid style={lightgray!92.02614379084967!black},
ymin=0.030170935933847, ymax=0.07209633448732827,
]
\addplot [line width=2.0pt, color0]
table [row sep=\\]{%
	0.02	0.046 \\
	0.04	0.052 \\
	0.06	0.058 \\
	0.08	0.061 \\
	0.1	    0.062 \\
	0.12	0.067 \\
	0.14	0.068 \\
	0.16    0.068 \\
	0.18	0.07 \\
	0.2	    0.071 \\
};
\addplot [line width=1.0pt, color1, mark=*, mark size=2, mark options={solid}]
table [row sep=\\]{%
	0.02	0.045 \\
	0.04	0.046 \\
	0.06	0.048 \\
	0.08	0.049 \\
	0.1	    0.048 \\
	0.12	0.049 \\
	0.14	0.048 \\
	0.16    0.05 \\
	0.18	0.05 \\
	0.2	    0.049 \\
};
\addplot [line width=1.0pt, color2, mark=+, mark size=3, mark options={solid}]
table [row sep=\\]{%
	0.02	0.045 \\
	0.04	0.047 \\
	0.06	0.046 \\
	0.08	0.046 \\
	0.1	    0.047 \\
	0.12	0.048 \\
	0.14	0.047 \\
	0.16    0.048 \\
	0.18	0.048 \\
	0.2	    0.046 \\
};
\addplot [line width=1.0pt, color3, mark=diamond, mark size=3, mark options={solid}]
table [row sep=\\]{%
	0.02	0.041 \\
	0.04	0.043 \\
	0.06	0.045 \\
	0.08	0.044 \\
	0.1	    0.045 \\
	0.12	0.044 \\
	0.14	0.045 \\
	0.16    0.044 \\
	0.18	0.0445 \\
	0.2	    0.044 \\
};
\addplot [line width=1.0pt, color4, mark=triangle*, mark size=3, mark options={solid,rotate=90}]
table [row sep=\\]{%
	0.02	0.047 \\
	0.04	0.055 \\
	0.06	0.058 \\
	0.08	0.06 \\
	0.1	    0.061 \\
	0.12	0.062 \\
	0.14	0.063 \\
	0.16    0.065 \\
	0.18	0.066 \\
	0.2	    0.067 \\
};

\addplot [line width=1.0pt, color6, mark=square, mark size=3, mark options={solid,rotate=90}]
table [row sep=\\]{%
	0.02	0.04 \\
	0.04	0.041 \\
	0.06	0.044 \\
	0.08	0.049 \\
	0.1	    0.05 \\
	0.12	0.052 \\
	0.14	0.057 \\
	0.16    0.058 \\
	0.18	0.06 \\
	0.2	    0.063 \\
};

\addplot [line width=1.0pt,dashed, color5, mark=dot, mark size=3, mark options={solid,rotate=90}]
table [row sep=\\]{%
	0.02 0.038 \\
	0.04 0.035 \\
	0.06 0.034 \\
	0.08 0.033 \\
	0.10 0.034 \\
	0.12 0.033 \\
	0.14 0.033 \\
	0.16 0.034 \\
	0.18 0.033 \\
	0.20 0.033 \\
};

\end{axis}

\end{tikzpicture}
	\hspace{-0.2cm}
\begin{tikzpicture}[scale=0.55]

\definecolor{color3}{rgb}{0.83921568627451,0.152941176470588,0.156862745098039}
\definecolor{color1}{rgb}{1,0.498039215686275,0.0549019607843137}
\definecolor{color2}{rgb}{0.172549019607843,0.627450980392157,0.172549019607843}
\definecolor{color4}{rgb}{0.580392156862745,0.403921568627451,0.741176470588235}
\definecolor{color0}{rgb}{0.12156862745098,0.466666666666667,0.705882352941177}
\definecolor{color5}{rgb}{0.549019607843137,0.337254901960784,0.294117647058824}

\begin{axis}[
tick align=outside,
tick pos=left,
x grid style={lightgray!92.02614379084967!black},
xlabel={Label rate},
xmin=0.0, xmax=0.055,
y grid style={lightgray!92.02614379084967!black},
ymin=0.214170935933847, ymax=0.315,
]
\addplot [line width=2.0pt, color0]
table [row sep=\\]{%
0.005 0.255 \\
0.01  0.272 \\
0.02  0.286 \\
0.03  0.292 \\
0.04  0.295 \\
0.05  0.295 \\
};
\addplot [line width=1.0pt, color1, mark=*, mark size=2, mark options={solid}]
table [row sep=\\]{%
0.005 0.256 \\
0.01  0.272 \\
0.02  0.29 \\
0.03  0.3 \\
0.04  0.301 \\
0.05  0.31 \\
};
\addplot [line width=1.0pt, color2, mark=+, mark size=3, mark options={solid}]
table [row sep=\\]{%
0.005 0.249 \\
0.01  0.27 \\
0.02  0.284 \\
0.03  0.292 \\
0.04  0.295 \\
0.05  0.3 \\
};
\addplot [line width=1.0pt, color3, mark=diamond, mark size=3, mark options={solid}]
table [row sep=\\]{%
0.005 0.22 \\
0.01  0.255 \\
0.02  0.277 \\
0.03  0.289 \\
0.04  0.293 \\
0.05  0.294 \\
};
\addplot [line width=1.0pt, color4, mark=triangle*, mark size=3, mark options={solid,rotate=90}]
table [row sep=\\]{%
0.005 0. \\
0.01  0. \\
0.02  0. \\
0.03  0. \\
0.04  0. \\
0.05  0. \\
};

\addplot [line width=1.0pt,dashed, color5, mark=dot, mark size=3, mark options={solid,rotate=90}]
table [row sep=\\]{%
0.005 0. \\
0.01  0. \\
0.02  0. \\
0.03  0. \\
0.04  0. \\
0.05  0. \\
};

\end{axis}

\end{tikzpicture}
	\phantom{ppppp}\ref{named3}
	\caption{Micro (upper row) and Macro (lower row) $F_1$ scores that different embeddings + logistic regression yield on labeled graphs, as a function of the labeling rated (percentage of training data)} \label{fig:class}
	\vspace{-0.35cm}
\end{figure*}
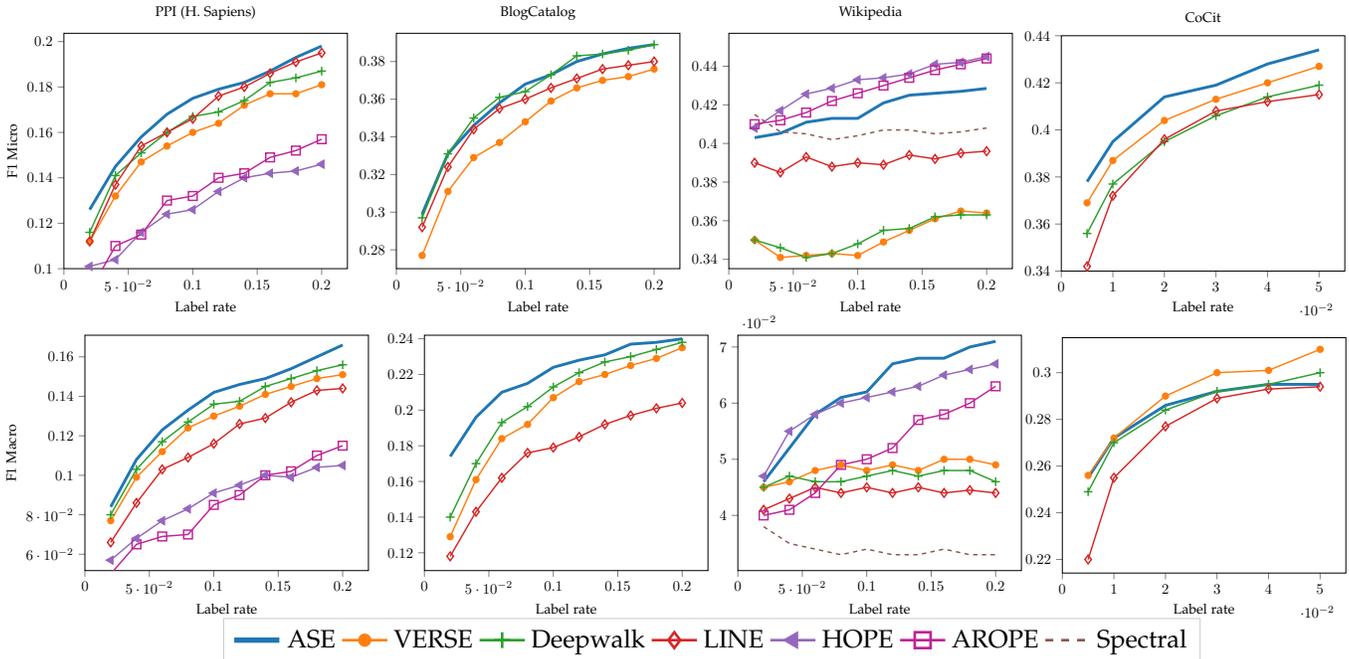

\noindent \textbf{Interpretation of results}. One interesting aspect of the proposed ASE method, is that the inferred parameters $\boldsymbol{\theta}^\ast$ from the first phase of Algorithm 1 can be used to characterise the underlying similarity structure of the graph, and the way nodes ``interact'' over different path lengths (short, medium, and long range). The ``strength'' of interactions is inferred by how uniform the coefficients of $\boldsymbol{\theta}^\ast$ are, and depend on the value of $\lambda$. Since the default value was $\lambda=1$ for all graphs, the results can be interpreted as relative interaction strengths between them. The resulting $\{\boldsymbol{\theta}^\ast\}$s for all graphs are listed in Table 3. 

It can be immediately observed that the type of node interactions varies significantly across different graphs, with similar behavior for graphs that belong to the same domain. Specifically, \texttt{ca-CondMat, ca-AstroPh}, and \texttt{CoCit} that belong to the citation/co-authorship domain all show relatively strong interactions of short range. \texttt{BlogCatalog} shows very strong short-range similarities of only one-hop neighborhood interactions among bloggers. On the other hand, the \texttt{Wikipedia} word co-occurrence network shows a strong tendency for long-range interactions; while other graphs, such as the \texttt{PPI} protein interaction network stay on the medium range.\\

\noindent \textbf{Node classification}. 
Graphs with labeled nodes are frequently used to measure the ability of embedding methods to produce features suitable for classification. For each experiment, nodes were randomly split to a training set and a test set. Similar to other works, and to cope with multi-label targets, we fed the training features and labels into the one-vs-the-rest configuration of logistic regression classifier provided by the \texttt{sklearn} Python library. In the testing phase, we sorted the predicted class probabilities for each node in decreasing order, and extracted the top-$k_i$ ranking labels, were $k_i$ is the true number of labels of node $v_i$. We then computed the Micro- and Macro-averaged $F_1$ scores \cite{manning2008ir} of the predicted labels.     

Apart from comparisons with alternative embedding methods, node classification can reveal whether available node labels (metadata) are distributed in a manner that matches the node relations/interactions that are inferred by ASE. To reveal this information, we obtain embeddings for every $k\in \{1,\ldots,10\}$ by ignoring the training phase and ``forcing'' $\boldsymbol{\theta}^\ast=\mathbf{e}_k$ (i.e., 1 at the $k$-th entry and $0$ elsewhere) in Algorithm \ref{alg:ASE}, and then using each embedding for classification with $10\%$ labeling rate. Figure  \ref{fig:single} plots Micro and Macro $F_1$ for all labeled graphs as a function of $k$, while red shade is placed on the hops where the \emph{unsupervised} ASE parameters $\boldsymbol{\theta}^\ast$ are non-zero (cf. Table 1). As seen in Fig. \ref{fig:single}, the accuracy on the four labeled graphs evolves with $k$ in a markedly different manner. Nevertheless, ASE identifies the trends and tends to assign non-zero weights to hops that yield a desirable trade-off between Micro and Macro $F_1$. Bearing in mind that ASE does \emph{not} use labels for training or validation, this is rather remarkable considering the fact that $\boldsymbol{\theta}^\ast$ depends only on the graph.

\begin{figure*}[t!]  
	\centering
\begin{tikzpicture}[scale=0.65]

\definecolor{color3}{rgb}{0.83921568627451,0.152941176470588,0.156862745098039}
\definecolor{color1}{rgb}{1,0.498039215686275,0.0549019607843137}
\definecolor{color2}{rgb}{0.172549019607843,0.627450980392157,0.172549019607843}
\definecolor{color4}{rgb}{0.580392156862745,0.403921568627451,0.741176470588235}
\definecolor{color0}{rgb}{0.12156862745098,0.466666666666667,0.705882352941177}
\definecolor{color5}{rgb}{0.549019607843137,0.337254901960784,0.294117647058824}
\definecolor{color6}{rgb}{0.749019607843137,0.137254901960784,0.594117647058824}

\begin{semilogyaxis}[
legend cell align={left},
legend entries={{ASE},{VERSE},{Deepwalk},{LINE},{HOPE},{AROPE},{Spectral}},
legend style={at={(0.97,0.03)}, anchor=south east, draw=white!80.0!black},
legend columns = 7,
legend to name = {named4},
tick align=outside,
tick pos=left,
x grid style={lightgray!92.02614379084967!black},
xlabel={$\#$ clusters},
ylabel = { Mean Conductance },
xmin=3 , xmax=101,
y grid style={lightgray!92.02614379084967!black},
ymin=0.0, ymax=0.43,
title =  { ca-AstroPh }
]
\addplot [line width=2.0pt, color0]
table [row sep=\\]{%
4 0.066 \\
8 0.063 \\ 
12 0.067 \\
16 0.069 \\
20 0.073 \\
24 0.074 \\
28 0.073 \\
32 0.080 \\
36 0.082 \\
40 0.084 \\
60 0.092 \\
80 0.105 \\
100 0.115 \\
};
\addplot [line width=1.0pt, color1, mark=*, mark size=2, mark options={solid}]
table [row sep=\\]{%
4 0.136 \\
8 0.221 \\
12 0.257 \\ 
16 0.283 \\
20 0.295 \\
24 0.303 \\
28 0.314 \\
32 0.321 \\
36 0.329 \\
40 0.337 \\
60 0.358 \\
80 0.372 \\
100 0.382 \\
};
\addplot [line width=1.0pt, color2, mark=+, mark size=3, mark options={solid}]
table [row sep=\\]{%
4 0.147 \\
8 0.160 \\
12 0.160 \\
16 0.175 \\
20 0.187 \\
24 0.180 \\
28 0.192 \\
32 0.192 \\
36 0.205 \\
40 0.197 \\
60 0.208 \\
80 0.226 \\
100 0.225 \\
};
\addplot [line width=1.0pt, color3, mark=diamond, mark size=3, mark options={solid}]
table [row sep=\\]{%
4 0.261  \\
8 0.286 \\
12 0.318 \\
16 0.332 \\
20 0.327 \\
24 0.337 \\
28 0.339 \\
32 0.345 \\
36 0.346 \\
40 0.348 \\
60 0.354 \\
80 0.349 \\
100 0.360 \\
};
\addplot [line width=1.0pt, color4, mark=triangle*, mark size=3, mark options={solid,rotate=90}]
table [row sep=\\]{%
4 0.246 \\
8 0.294 \\
12 0.296 \\
16 0.318 \\
20 0.333 \\
24 0.331 \\
28 0.355 \\
32 0.370 \\
36 0.388 \\
40 0.385 \\
60 0.50 \\
80 0.603 \\
100 0.665 \\
};

\addplot [line width=1.0pt, color6, mark=square, mark size=3, mark options={solid,rotate=90}]
table [row sep=\\]{%
	4 0.32 \\
	8 0.35 \\
	12 0.37 \\
	16 0.38 \\
	20 0.39 \\
	24 0.40 \\
	28 0.395 \\
	32 0.405 \\
	36 0.40 \\
	40 0.395 \\
	60 0.45 \\
	80 0.50 \\
	100 0.53 \\
};

\addplot [line width=1.0pt,dashed, color5, mark=dot, mark size=3, mark options={solid,rotate=90}]
table [row sep=\\]{%
4 0.099 \\
8 0.102 \\
12 0.110 \\
16 0.109 \\
20 0.101 \\
24 0.099 \\
28 0.105 \\
32 0.109 \\
36 0.102 \\
40 0.108 \\
60 0.115 \\
80 0.120 \\
100 0.125 \\
};

\end{semilogyaxis}

\end{tikzpicture}		
\begin{tikzpicture}[scale=0.65]

\definecolor{color3}{rgb}{0.83921568627451,0.152941176470588,0.156862745098039}
\definecolor{color1}{rgb}{1,0.498039215686275,0.0549019607843137}
\definecolor{color2}{rgb}{0.172549019607843,0.627450980392157,0.172549019607843}
\definecolor{color4}{rgb}{0.580392156862745,0.403921568627451,0.741176470588235}
\definecolor{color0}{rgb}{0.12156862745098,0.466666666666667,0.705882352941177}
\definecolor{color5}{rgb}{0.549019607843137,0.337254901960784,0.294117647058824}

\begin{semilogyaxis}[
tick align=outside,
tick pos=left,
x grid style={lightgray!92.02614379084967!black},
xlabel={$\#$ clusters},
ylabel = { Mean Conductance },
xmin=3 , xmax=101,
y grid style={lightgray!92.02614379084967!black},
ymin=0.0, ymax=0.3,
title =  { ca-CondMat }
]
\addplot [line width=2.0pt, color0]
table [row sep=\\]{%
4 0.030 \\
8 0.045 \\
12 0.049 \\
16 0.053 \\
20 0.056 \\
24 0.058 \\
28 0.058 \\
32 0.060 \\
36 0.064 \\
40 0.066 \\
60 0.074 \\
80 0.087 \\
100 0.098 \\
};
\addplot [line width=1.0pt, color1, mark=*, mark size=2, mark options={solid}]
table [row sep=\\]{%
4 0.102 \\
8 0.172 \\
12 0.207 \\
16 0.224 \\
20 0.235 \\
24 0.244 \\
28 0.251 \\
32 0.254 \\
36 0.258 \\
40 0.260 \\
60 0.267 \\
80 0.275 \\
100 0.281 \\
};
\addplot [line width=1.0pt, color2, mark=+, mark size=3, mark options={solid}]
table [row sep=\\]{%
4 0.116 \\
8 0.157 \\
12 0.168 \\
16 0.175 \\
20 0.181 \\
24 0.187 \\
28 0.182 \\
32 0.188 \\
36 0.197 \\
40 0.189 \\
60 0.188 \\
80 0.195 \\
100 0.202 \\
};
\addplot [line width=1.0pt, color3, mark=diamond, mark size=3, mark options={solid}]
table [row sep=\\]{%
4 0.198 \\
8 0.228 \\
12 0.239 \\
16 0.249 \\
20 0.252 \\
24 0.260 \\
28 0.263 \\
32 0.264 \\
36 0.265 \\
40 0.270 \\
60 0.277 \\
80 0.288 \\
100 0.291 \\
};
\addplot [line width=1.0pt, color4, mark=triangle*, mark size=3, mark options={solid,rotate=90}]
table [row sep=\\]{%
4 0.297 \\
8 0.368 \\
12 0.379 \\
16 0.411 \\
20 0.423 \\
24 0.435 \\
28 0.442 \\
32 0.460 \\
36 0.471 \\
40 0.482 \\
};

\addplot [line width=1.0pt, color4, mark=triangle*, mark size=3, mark options={solid,rotate=90}]
table [row sep=\\]{%
	4 0.31 \\
	8 0.38 \\
	12 0.45 \\
	16 0.47 \\
	20 0.44 \\
	24 0.47 \\
	28 0.48 \\
	32 0.45 \\
	36 0.46 \\
	40 0.47 \\
};

\addplot [line width=1.0pt,dashed, color5, mark=dot, mark size=3, mark options={solid,rotate=90}]
table [row sep=\\]{%
4 0.044 \\
8 0.050 \\
12 0.061 \\
16 0.059 \\
20 0.062 \\
24 0.061 \\
28 0.066 \\
32 0.067 \\
36 0.066 \\
40 0.071 \\
60 0.074 \\
80 0.096 \\
100 0.097 \\
};

\end{semilogyaxis}

\end{tikzpicture}
\begin{tikzpicture}[scale=0.65]

\definecolor{color3}{rgb}{0.83921568627451,0.152941176470588,0.156862745098039}
\definecolor{color1}{rgb}{1,0.498039215686275,0.0549019607843137}
\definecolor{color2}{rgb}{0.172549019607843,0.627450980392157,0.172549019607843}
\definecolor{color4}{rgb}{0.580392156862745,0.403921568627451,0.741176470588235}
\definecolor{color0}{rgb}{0.12156862745098,0.466666666666667,0.705882352941177}
\definecolor{color5}{rgb}{0.549019607843137,0.337254901960784,0.294117647058824}

\begin{semilogyaxis}[
tick align=outside,
tick pos=left,
x grid style={lightgray!92.02614379084967!black},
xlabel={$\#$ clusters},
ylabel = { Mean Conductance },
xmin=3 , xmax=101,
y grid style={lightgray!92.02614379084967!black},
ymin=0.0, ymax=0.5,
title =  { Enron }
]
\addplot [line width=2.0pt, color0]
table [row sep=\\]{%
4 0.026 \\
8 0.028 \\
12 0.033 \\
16 0.035 \\
20 0.033 \\
24 0.034 \\
28 0.035 \\
32 0.037 \\
36 0.038 \\
40 0.039 \\
60 0.042 \\
80 0.060 \\
100 0.099 \\
};
\addplot [line width=1.0pt, color1, mark=*, mark size=2, mark options={solid}]
table [row sep=\\]{%
4 0.105 \\
8 0.192 \\
12 0.247 \\
16 0.278 \\
20 0.295 \\
24 0.318 \\
28 0.322 \\
32 0.339 \\
36 0.348 \\
40 0.350 \\
60 0.424 \\
80 0.429 \\
100 0.441 \\
};
\addplot [line width=1.0pt, color2, mark=+, mark size=3, mark options={solid}]
table [row sep=\\]{%
4 0.126 \\
8 0.176 \\
12 0.202 \\
16 0.209 \\
20 0.204 \\
24 0.222 \\
28 0.207 \\
32 0.220 \\
36 0.218 \\
40 0.223 \\
60 0.240 \\
80 0.245 \\
100 0.233 \\
};
\addplot [line width=1.0pt, color3, mark=diamond, mark size=3, mark options={solid}]
table [row sep=\\]{%
4 0.189 \\
8 0.244 \\
12 0.305 \\
16 0.304 \\
20 0.344 \\
24 0.348 \\
28 0.366 \\
32 0.363 \\
36 0.374 \\
40 0.398 \\
60 0.404 \\
80 0.429 \\
100 0.442 \\
};
\addplot [line width=1.0pt, color4, mark=triangle*, mark size=3, mark options={solid,rotate=90}]
table [row sep=\\]{%
4 0.892 \\
8 0.839 \\
12 0.795 \\
16 0.813 \\
20 0.833 \\
24 0.837 \\
28 0.839 \\
32 0.815 \\
36 0.815 \\
40 0.835 \\
};

\addplot [line width=1.0pt, color6, mark=square, mark size=3, mark options={solid,rotate=90}]
table [row sep=\\]{%
	4 0.892 \\
	8 0.839 \\
	12 0.795 \\
	16 0.813 \\
	20 0.833 \\
	24 0.837 \\
	28 0.839 \\
	32 0.815 \\
	36 0.815 \\
	40 0.835 \\
};

\addplot [line width=1.0pt,dashed, color5, mark=dot, mark size=3, mark options={solid,rotate=90}]
table [row sep=\\]{%
4 0.034 \\
8 0.039 \\
12 0.036 \\
16 0.040 \\
20 0.044 \\
24 0.041 \\
28 0.039 \\
32 0.042 \\
36 0.042 \\
40 0.042 \\
60 0.047 \\
80 0.065 \\
100 0.11 \\
};

\end{semilogyaxis}

\end{tikzpicture}
\begin{tikzpicture}[scale=0.65]

\definecolor{color3}{rgb}{0.83921568627451,0.152941176470588,0.156862745098039}
\definecolor{color1}{rgb}{1,0.498039215686275,0.0549019607843137}
\definecolor{color2}{rgb}{0.172549019607843,0.627450980392157,0.172549019607843}
\definecolor{color4}{rgb}{0.580392156862745,0.403921568627451,0.741176470588235}
\definecolor{color0}{rgb}{0.12156862745098,0.466666666666667,0.705882352941177}
\definecolor{color5}{rgb}{0.549019607843137,0.337254901960784,0.294117647058824}

\begin{semilogyaxis}[
tick align=outside,
tick pos=left,
x grid style={lightgray!92.02614379084967!black},
xlabel={$\#$ clusters},
ylabel = { Mean Conductance },
xmin=3 , xmax=101,
y grid style={lightgray!92.02614379084967!black},
ymin=0.0, ymax=0.98,
title =  { BlogCatalog }
]
\addplot [line width=2.0pt, color0]
table [row sep=\\]{%
4 0.659 \\
8 0.668 \\
12 0.710 \\
16 0.720 \\
20 0.726 \\
24 0.739 \\
28 0.746 \\
32 0.755 \\
36 0.764 \\
40 0.770 \\
60 0.796 \\
80 0.823 \\
100 0.855 \\
};
\addplot [line width=1.0pt, color1, mark=*, mark size=2, mark options={solid}]
table [row sep=\\]{%
4 0.507 \\
8 0.732 \\
12 0.766 \\
16 0.784 \\
20 0.801 \\
24 0.813 \\
28 0.825 \\
32 0.834 \\
36 0.842 \\
40 0.847 \\
60 0.871 \\
80 0.883 \\
100 0.893 \\
};
\addplot [line width=1.0pt, color2, mark=+, mark size=3, mark options={solid}]
table [row sep=\\]{%
4 0.493 \\
8 0.638 \\
12 0.698 \\
16 0.733 \\
20 0.754 \\
24 0.770 \\
28 0.784 \\
32 0.794 \\
36 0.804 \\
40 0.813 \\
60 0.840 \\
80 0.860 \\
100 0.874 \\
};
\addplot [line width=1.0pt, color3, mark=diamond, mark size=3, mark options={solid}]
table [row sep=\\]{%
4 0.523 \\
8 0.658 \\
12 0.723 \\
16 0.757 \\
20 0.781 \\
24 0.801 \\
28 0.815 \\
32 0.827 \\
36 0.839 \\
40 0.849 \\
60 0.883 \\
80 0.906 \\
100 0.920 \\
};
\addplot [line width=1.0pt, color4, mark=triangle*, mark size=3, mark options={solid,rotate=90}]
table [row sep=\\]{%
4 0.777 \\
8 0.867 \\
12 0.909 \\
16 0.930 \\
20 0.939 \\
24 0.947 \\
28 0.951 \\
32 0.953 \\
36 0.957 \\
40 0.959 \\
60 0.967 \\
80 0.963 \\
100 0.967 \\
};

\addplot [line width=1.0pt, color6, mark=square, mark size=3, mark options={solid,rotate=90}]
table [row sep=\\]{%
	4 0.73 \\
	8 0.86 \\
	12 0.90 \\
	16 0.92 \\
	20 0.94 \\
	24 0.95 \\
	28 0.955 \\
	32 0.956 \\
	36 0.957 \\
	40 0.955 \\
	60 0.956 \\
	80 0.964 \\
	100 0.968 \\
};

\addplot [line width=1.0pt,dashed, color5, mark=dot, mark size=3, mark options={solid,rotate=90}]
table [row sep=\\]{%
4 0.708 \\
8 0.724 \\
12 0.728 \\
16 0.729 \\
20 0.758 \\
24 0.750 \\
28 0.765 \\
32 0.767 \\
36 0.771 \\
40 0.778 \\
60 0.801 \\
80 0.823 \\
100 0.855 \\
};
\end{semilogyaxis}

\end{tikzpicture}		
\begin{tikzpicture}[scale=0.65]

\definecolor{color3}{rgb}{0.83921568627451,0.152941176470588,0.156862745098039}
\definecolor{color1}{rgb}{1,0.498039215686275,0.0549019607843137}
\definecolor{color2}{rgb}{0.172549019607843,0.627450980392157,0.172549019607843}
\definecolor{color4}{rgb}{0.580392156862745,0.403921568627451,0.741176470588235}
\definecolor{color0}{rgb}{0.12156862745098,0.466666666666667,0.705882352941177}
\definecolor{color5}{rgb}{0.549019607843137,0.337254901960784,0.294117647058824}

\begin{semilogyaxis}[
tick align=outside,
tick pos=left,
x grid style={lightgray!92.02614379084967!black},
xlabel={$\#$ clusters},
ylabel = { Mean Conductance },
xmin=3 , xmax=101,
y grid style={lightgray!92.02614379084967!black},
ymin=0.0, ymax=0.46,
title =  { com-Amazon }
]
\addplot [line width=2.0pt, color0]
table [row sep=\\]{%
4 0.020 \\
8 0.016 \\
12 0.021 \\
16 0.022 \\
20 0.018 \\
24 0.025 \\
28 0.022 \\
32 0.026 \\
36 0.024 \\
40 0.026 \\
60 0.029 \\
80 0.030 \\
100 0.043 \\
};
\addplot [line width=1.0pt, color1, mark=*, mark size=2, mark options={solid}]
table [row sep=\\]{%
4 0.054 \\
8 0.070 \\
12 0.083 \\
16 0.084 \\
20 0.089 \\
24 0.093 \\
28 0.094 \\
32 0.098 \\
36 0.101 \\
40 0.104 \\
60 0.109 \\
80 0.111 \\
100 0.115 \\
};
\addplot [line width=1.0pt, color2, mark=+, mark size=3, mark options={solid}]
table [row sep=\\]{%
	4	0.076 \\
	8	0.079 \\
	12	0.094 \\
	16	0.092 \\
	20	    0.095 \\
	24	0.102 \\
	28	0.103 \\
	32    0.105 \\
	36	0.107 \\
	40	    0.115 \\
60 0.104 \\
80 0.099 \\
100 0.095 \\
};
\addplot [line width=1.0pt, color3, mark=diamond, mark size=3, mark options={solid}]
table [row sep=\\]{%
	4	0.431 \\
	8	0.376 \\
	12	0.356 \\
	16	0.344 \\
	20	    0.343 \\
	24	0.334 \\
	28	0.332 \\
	32    0.327 \\
	36	0.329 \\
	40	    0.32 \\
	60 0.338 \\
	80 0.354 \\
	100 0.372 \\
};
\addplot [line width=1.0pt, color4, mark=triangle*, mark size=3, mark options={solid,rotate=90}]
table [row sep=\\]{%
	4	0.047 \\
	6	0.055 \\
	12	0.058 \\
	16	0.06 \\
	20	    0.064 \\
	24	0.062 \\
	28	0.063 \\
	32    0.065 \\
	36	0.066 \\
	40	    0.067 \\
	60	    0.072 \\
	80	    0.074 \\
	100	    0.08 \\	
};

\addplot [line width=1.0pt, color6, mark=square, mark size=3, mark options={solid,rotate=90}]
table [row sep=\\]{%
	4	0.6 \\
};

\addplot [line width=1.0pt,dashed, color5, mark=dot, mark size=3, mark options={solid,rotate=90}]
table [row sep=\\]{%
4 0.007 \\
8 0.009 \\
12 0.015 \\
16 0.021 \\
20 0.032 \\
24 0.034 \\
28 0.029 \\
32 0.027 \\
36 0.027 \\
40 0.03 \\
60 0.029 \\
80 0.032 \\
100 0.045 \\
};

\end{semilogyaxis}

\end{tikzpicture}
\begin{tikzpicture}[scale=0.65]

\definecolor{color3}{rgb}{0.83921568627451,0.152941176470588,0.156862745098039}
\definecolor{color1}{rgb}{1,0.498039215686275,0.0549019607843137}
\definecolor{color2}{rgb}{0.172549019607843,0.627450980392157,0.172549019607843}
\definecolor{color4}{rgb}{0.580392156862745,0.403921568627451,0.741176470588235}
\definecolor{color0}{rgb}{0.12156862745098,0.466666666666667,0.705882352941177}
\definecolor{color5}{rgb}{0.549019607843137,0.337254901960784,0.294117647058824}

\begin{semilogyaxis}[
tick align=outside,
tick pos=left,
x grid style={lightgray!92.02614379084967!black},
xlabel={$\#$ clusters},
ylabel = { Mean Conductance },
xmin=3 , xmax=101,
y grid style={lightgray!92.02614379084967!black},
ymin=0.0, ymax=0.66,
title =  { com-DBLP }
]
\addplot [line width=2.0pt, color0]
table [row sep=\\]{%
4 0.014 \\
8 0.014 \\
12 0.019 \\
16 0.021 \\
20 0.020 \\
24 0.018 \\
28 0.020 \\
32 0.021 \\
36 0.022 \\
40 0.026 \\
60 0.028 \\
80 0.036 \\
100 0.053 \\
};
\addplot [line width=1.0pt, color1, mark=*, mark size=2, mark options={solid}]
table [row sep=\\]{%
4 0.107 \\
8 0.14 \\
12 0.155 \\
16 0.171 \\
20 0.179 \\
24 0.19 \\
28 0.194 \\
32 0.197 \\
36 0.199 \\
40 0.201 \\
60 0.210 \\
80 0.221 \\
100 0.224 \\
};
\addplot [line width=1.0pt, color2, mark=+, mark size=3, mark options={solid}]
table [row sep=\\]{%
4 0.109 \\
8 0.135 \\
12 0.150 \\
16 0.162 \\
20 0.171 \\
24 0.178 \\
28 0.185 \\
32 0.184 \\
36 0.187 \\
40 0.191 \\
60 0.187  \\
80 0.184  \\
100 0.174  \\
};
\addplot [line width=1.0pt, color3, mark=diamond, mark size=3, mark options={solid}]
table [row sep=\\]{%
4 0.61  \\
8 0.55 \\
12 0.56 \\
16 0.56 \\
20 0.57 \\
24 0.58 \\
28 0.56 \\
32 0.55 \\
36 0.56 \\
40 0.575 \\
60 0.55 \\
80 0.568 \\
100 0.575 \\
};
\addplot [line width=1.0pt, color4, mark=triangle*, mark size=3, mark options={solid,rotate=90}]
table [row sep=\\]{%
4 0.067 \\
8 0.091 \\
12 0.095 \\
16 0.098 \\
20 0.112 \\
24 0.147 \\
28 0.165 \\
32 0.184 \\
36 0.204 \\
40 0.220 \\
60 0.329 \\
80 0.432 \\
100 0.501 \\
};

\addplot [line width=1.0pt, color6, mark= square, mark size=3, mark options={solid,rotate=90}]
table [row sep=\\]{%
	4 0.11 \\
	8 0.15 \\
	12 0.16 \\
	16 0.17 \\
	20 0.18 \\
	24 0.19 \\
	28 0.17 \\
	32 0.19 \\
	36 0.21 \\
	40 0.215 \\
	60 0.23 \\
	80 0.286 \\
	100 0.344 \\
};

\addplot [line width=1.0pt,dashed, color5, mark=dot, mark size=3, mark options={solid,rotate=90}]
table [row sep=\\]{%
4 0.014 \\
8 0.02 \\
12 0.021 \\
16 0.025 \\
20 0.022 \\
24 0.021 \\
28 0.023 \\
32 0.022 \\
36 0.027 \\
40 0.025 \\
60 0.031 \\
80 0.037 \\
100 0.056 \\
};

\end{semilogyaxis}

\end{tikzpicture}
	\phantom{ppppp}\ref{named4}
		\vspace{-0.2cm}
	\caption{Average conductance of different embeddings used by kmeans for clustering, as a function of number of clusters. } \label{fig:cluster}
	\vspace{-0.3cm}
\end{figure*}
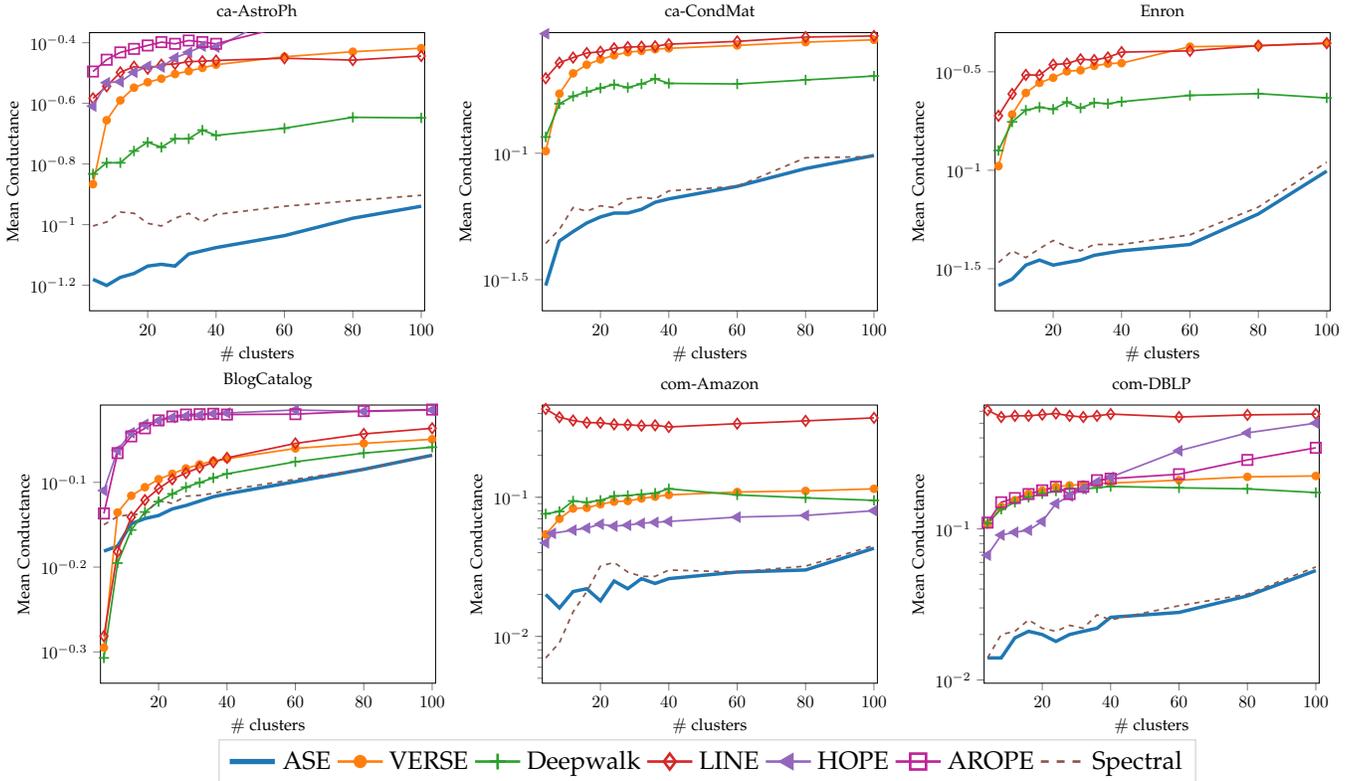

We also compared the classification accuracy of ASE embeddings with those of the alternative embedding approaches, with results plotted in Fig. \ref{fig:class}. The plots for some method-graph pairs are not discernible  when values are too low. While the relative performance of any given method varies from graph to graph, ASE adapts to each graph and yields consistently reliable embeddings, with accuracy that in most cases reaches or surpasses that of state-of-the-art methods, especially in terms of Macro $F_1$. The two exceptions are the Macro $F_1$ in \texttt{CoCit}, and Micro $F_1$ in \texttt{Wikipedia}, where VERSE and HOPE are correspondingly more accurate. Interestingly, HOPE achieving high Micro $F_1$ and low Macro $F_1$ in \texttt{Wikipedia} is in agreement with the findings in Fig. \ref{fig:single}, combined with the fact that HOPE focuses on longer paths.\\
 \begin{table}[h]
 	\centering
 	\caption{ Link Prediction Accuracy on \texttt{vk2016-17} }
 	\rowcolors{2}{}{gray!7}
 	\begin{tabular} { c c c c c c c }         			
 		\toprule
 		VERSE &  ASE  & LINE & Deepwalk & AROPE & HOPE & Spectral \\
 		\midrule 
 		0.79 & 0.75 & 0.74 & 0.69  & 0.65 & 0.62  & 0.60 \\		
 		\bottomrule
 	\end{tabular}\label{tab:link_pred}
 \end{table}

\noindent \textbf{Link prediction}. Link prediction is the task of estimating the probability that a link
between two unconnected nodes will appear in the future. We repeat the experiment performed in \cite{VERSE} on the \texttt{vk2016-17} social network. For every possible edge, we build a feature vector as the Hadamard product between the embedded vectors of its two adjacent nodes. Using the two time instances of \texttt{vk2016-17}, we predict whether a new friendship link appears between November
2016 and May 2017, using $50\%$ of the new links for training and $50\%$
for testing. To train the binary logistic regression classifier, we also  randomly sample non-existing
edges as negative examples. The link prediction accuracy for different embeddings is reported in Table 3. While for this experiment ASE does not reach the accuracy of VERSE, it provides the second most accurate link prediction, far surpassing the also SVD-based HOPE and spectral embeddings.
\\

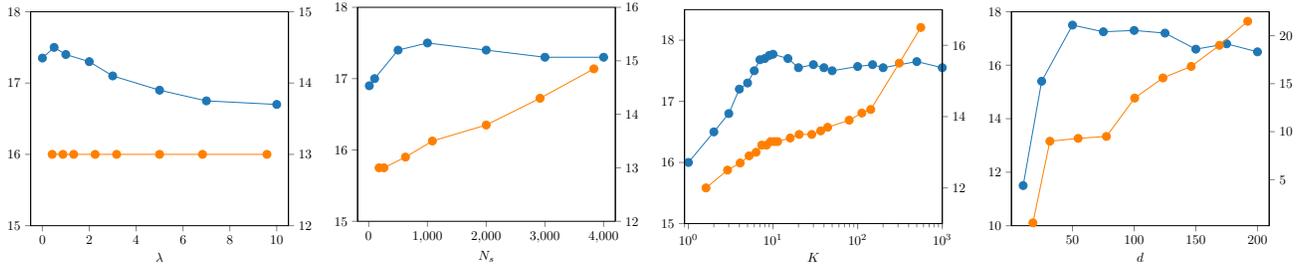
\begin{figure*}[t!]  
	\centering
\begin{tikzpicture}[scale=0.5]

\definecolor{color0}{rgb}{0.12156862745098,0.466666666666667,0.705882352941177}

\begin{axis}[
tick align=outside,
tick pos=left,
x grid style={white!69.01960784313725!black},
xlabel={$\lambda$},
xmin=-0.5, xmax=10.5,
y grid style={white!69.01960784313725!black},
ymin=15.0, ymax=18.0
]
\addplot [semithick, color0, mark=*, mark size=3, mark options={solid}, forget plot]
table [row sep=\\]{%
0	17.35 \\
0.5	17.5 \\
1	17.4 \\
2	17.3 \\
3	17.1 \\
5	16.9 \\
7	16.75 \\
10	16.7 \\
};
\end{axis}

\begin{axis}
[
tick align=outside,
tick pos=right,
axis x line=none,
y grid style={white!69.01960784313725!black},
ymin=12.0, ymax=15.0
]
\addplot [semithick, orange, mark=*, mark size=3, mark options={solid}, forget plot]
table [row sep=\\]{%
	0	13.0 \\
	0.5	13.0 \\
	1	13.0 \\
	2	13.0 \\
	3	13.0 \\
	5	13.0 \\
	7	13.0 \\
	10	13.0 \\
};
\end{axis}

\end{tikzpicture}
	\vspace{0.3cm}	
\begin{tikzpicture}[scale=0.5]

\definecolor{color0}{rgb}{0.12156862745098,0.466666666666667,0.705882352941177}

\begin{axis}[
tick align=outside,
tick pos=left,
x grid style={white!69.01960784313725!black},
xlabel={$N_s$},
xmin=-189.5, xmax=4199.5,
y grid style={white!69.01960784313725!black},
ymin=15.0, ymax=18.0
]
\addplot [semithick, color0, mark=*, mark size=3, mark options={solid}, forget plot]
table [row sep=\\]{%
10	16.9 \\
100	17 \\
500	17.4 \\
1000	17.5 \\
2000	17.4 \\
3000	17.3 \\
4000	17.3 \\
};
\end{axis}

\begin{axis}
[
tick align=outside,
tick pos=right,
axis x line=none,
y grid style={white!69.01960784313725!black},
ymin=12.0, ymax=16.0
]
\addplot [semithick, orange, mark=*, mark size=3, mark options={solid}, forget plot]
table [row sep=\\]{%
10	13.0 \\
100	13.0 \\
500	13.2 \\
1000	13.5 \\
2000	13.8 \\
3000	14.3 \\
4000	14.85 \\
};
\end{axis}

\end{tikzpicture}
	\vspace{0.3cm}	
\begin{tikzpicture}[scale=0.5]

\definecolor{color0}{rgb}{0.12156862745098,0.466666666666667,0.705882352941177}

\begin{semilogxaxis}[
tick align=outside,
tick pos=left,
x grid style={white!69.01960784313725!black},
xlabel={$K$},
xmin=0.9, xmax=1000.5,
y grid style={white!69.01960784313725!black},
ymin=15.0, ymax=18.5
]
\addplot [semithick, color0, mark=*, mark size=3, mark options={solid}, forget plot]
table [row sep=\\]{%
1	16.0 \\
2	16.5 \\
3	16.8 \\
4	17.2 \\
5	17.3 \\
6	17.5 \\
7	17.68 \\
8	17.7 \\
9	17.75 \\
10  17.77 \\
15  17.7 \\
20  17.55 \\
30  17.6 \\
40  17.55 \\
50  17.5 \\
100  17.57 \\
150  17.6 \\
200  17.55 \\
500  17.65 \\
1000  17.55 \\
};
\end{semilogxaxis}

\begin{semilogxaxis}
[
tick align=outside,
tick pos=right,
axis x line=none,
y grid style={white!69.01960784313725!black},
ymin=11.0, ymax=17.0
]
\addplot [semithick, orange, mark=*, mark size=3, mark options={solid}, forget plot]
table [row sep=\\]{%
1	12.0 \\
2	12.5 \\
3	12.7 \\
4	12.9 \\
5	13.0 \\
6	13.2 \\
7	13.2 \\
8	13.3 \\
9	13.3 \\
10  13.3 \\
15  13.4 \\
20  13.5 \\
30  13.5 \\
40  13.6 \\
50  13.7 \\
100  13.9 \\
150  14.1 \\
200  14.2 \\
500  15.5 \\
1000  16.5 \\
};
\end{semilogxaxis}

\end{tikzpicture}
		\vspace{0.3cm}		
\begin{tikzpicture}[scale=0.5]

\definecolor{color0}{rgb}{0.12156862745098,0.466666666666667,0.705882352941177}

\begin{axis}[
tick align=outside,
tick pos=left,
x grid style={white!69.01960784313725!black},
xlabel={$d$},
xmin=0.5, xmax=209.5,
y grid style={white!69.01960784313725!black},
ymin=10.0, ymax=18.0
]
\addplot [semithick, color0, mark=*, mark size=3, mark options={solid}, forget plot]
table [row sep=\\]{%
10	11.5 \\
25	15.4 \\
50	17.5 \\
75	17.25 \\
100	17.3 \\
125	17.2 \\
150	16.6 \\
175	16.8 \\
200	16.5 \\
};
\end{axis}

\begin{axis}
[
tick align=outside,
tick pos=right,
axis x line=none,
y grid style={white!69.01960784313725!black},
ymin=0.2, ymax=22.5
]
\addplot [semithick, orange, mark=*, mark size=3, mark options={solid}, forget plot]
table [row sep=\\]{%
10	0.5 \\
25	9.0 \\
50	9.3 \\
75	9.5 \\
100	13.5 \\
125	15.6 \\
150	16.8 \\
175	19.0 \\
200	21.5 \\
};
\end{axis}

\end{tikzpicture}		
			\vspace{-0.9 cm}
	\caption{ Sensitivity (\textcolor{blue}{F-1 Micro} on left axes, and \textcolor{orange}{Runtime} on right axes) of ASE on \texttt{PPI} graphs wrt various parameters. } \label{fig:sense}
		\vspace{-0.3 cm}
\end{figure*}
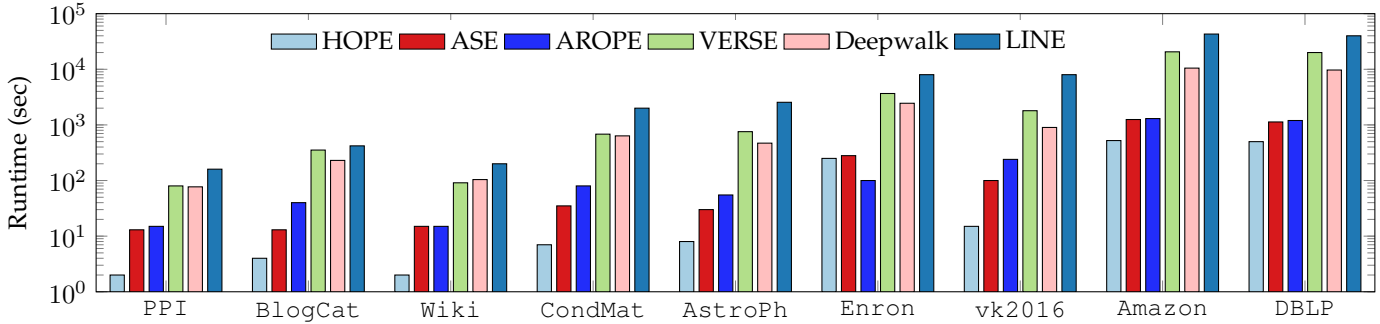
\begin{figure*}[t!]  
	\centering
%
%
\definecolor{mycolor1}{RGB}{215,25,28}%
\definecolor{mycolor2}{RGB}{166,206,227}%
\definecolor{mycolor3}{RGB}{31,120,180}
\definecolor{mycolor4}{RGB}{178,223,138}
\definecolor{mycolor5}{RGB}{34,45,250}

\begin{tikzpicture}

\begin{axis}[%
width=17 cm,
height=3.7 cm,
at={(0\figW,0\figH)},
scale only axis,
bar shift auto,
log origin=infty,
xmin=0.509090909090909,
xmax=9.49090909090909,
xtick={1,2,3,4,5,6,7,8,9},
xticklabels={{\texttt{PPI}},{\texttt{BlogCat}},{\texttt{Wiki}},{\texttt{CondMat}},{\texttt{AstroPh}}, {\texttt{Enron}},{\texttt{vk2016}},{\texttt{Amazon}},{\texttt{DBLP}}},
ymode=log,
ymin=1,
ymax=100000,
yminorticks=true,
ylabel style={font=\color{white!15!black}},
ylabel={Runtime (sec) },
ylabel near ticks,
axis background/.style={fill=white},
 legend columns=6,
legend style={at={(0.13,0.82)}, anchor=south west, legend cell align=left, align=left, draw=none}
]

\addplot[ybar, bar width=0.1, fill=mycolor2, draw=black, area legend] table[row sep=crcr] {%
	1	2 \\
	2	4 \\
	3	2\\
	4	7 \\
	5	8 \\
	6 	250 \\
	7 	15 \\
	8  523	\\
	9  500	\\
};
\addlegendentry{HOPE}

\addplot[ybar, bar width=0.1, fill=mycolor1, draw=black, area legend] table[row sep=crcr] {%
	1	13\\
	2	13\\
	3	15\\
	4	35\\
	5	30\\
	6	280\\
	7	100\\
	8	1250\\
	9	1130\\
};
\addlegendentry{ASE}

\addplot[ybar, bar width=0.1, fill=mycolor5, draw=black, area legend] table[row sep=crcr] {%
	1	15\\
	2	40\\
	3	15\\
	4	80\\
	5	55\\
	6	100\\
	7	240\\
	8	1300\\
	9	1200\\
};
\addlegendentry{AROPE}

\addplot[ybar, bar width=0.1, fill=mycolor4, draw=black, area legend] table[row sep=crcr] {%
1	80\\
2	353\\
3	91\\
4	682\\
5	756\\
6	3670\\
7	1800\\
8	20600\\
9	20000\\
};
\addlegendentry{VERSE}

\addplot[ybar, bar width=0.1, fill=pink, draw=black, area legend] table[row sep=crcr] {%
	1	77\\
	2	230\\
	3	104\\
	4	635\\
	5	470\\
	6	2450\\
	7	900\\
	8	10500 \\
	9	9700 \\
};
\addlegendentry{Deepwalk}

\addplot[ybar, bar width=0.1, fill=mycolor3, draw=black, area legend] table[row sep=crcr] {%
	1	160\\
	2	420\\
	3	200\\
	4	2000\\
	5	2550\\
	6	8000\\
	7	8000\\
	8	43000\\
	9	40000\\
};
\addlegendentry{LINE}

\end{axis}
\end{tikzpicture}%
	\caption{ Runtime of various embedding methods across different graphs } \label{fig:runtime}
	\vspace{-0.4 cm}
\end{figure*}
\noindent \textbf{Node clustering}. Finally, the embedded vectors were used to cluster the nodes into different communities, using the \texttt{sklearn} library K-means with the default K-means++ initialization \cite{kmeans++}. We evaluate the quality of node clustering with conductance, a well-known metric for measuring the goodness of a  community \cite{leskovec2009community}; conductance is minimized for large, well connected communities that are also well separated from the rest of the graph. Each plot in Fig. \ref{fig:cluster} gives the average conductance across communities, as a function of the total number of clusters. Results indicate that the proposed ASE as well as the spectral clustering benchmark yield much lower conductance compared to other embeddings. Apparently, since ASE builds on the same basis of eigenvectors used by normalized spectral clustering, it inherits the property of the latter to approximately minimize the normalized-cut metric \cite{spectral}, which is very similar to conductance. A closer look at the resulting clusters, reveals that clustering beased on VERSE, Deepwalk, LINE, and HOPE splits graphs into very large communities of roughly equal size, cutting a large number of edges in the process. This is an indication that these methods are  subject to a \emph{resolution limit}, which is the inability to detect well-separated communities that are below a certain size \cite{resolution}. On the other hand, Spectral and the proposed ASE separate the graph into a large-core component, and many smaller well-separated communities, a structure that many large-scale information networks have been observed to have \cite{leskovec2009community}. Indeed, the conductance gap is smaller for \texttt{BlogCatalog}, which is relatively small and with less pronounced communities.\\

\noindent \textbf{Parameter sensitivity}. We also present results after varying ASE parameters and measured embedding runtime for \texttt{PPI} as well as classification Micro $F_1$ accuracy with $10\%$ labeling rate. The aim is to assess the sensitivity of ASE wrt its basic parameters. The plot on the left shows how increasing $\lambda$ (cf. \eqref{final}) may decrease accuracy by forcing the entries of $\boldsymbol{\theta}^\ast$ to be close to uniform, thus losing the benefits of graph-specific adaptation. Regarding the number of sampled edges $N_s$, results (middle plot) indicate relative robustness of ASE embeddings, given a minimum number of samples. As expected, sampling a large number of edges may cause noticeable perturbation on the graph (even using the minimally-perturbing Algorithm 2); this may be causing a slight decrease in accuracy. Sensitivity is also measured wrt $K$ (i.e., the maximum walk length considered in the optimization). As expected, the accuracy increases sharply with $K$ for the first few steps, and then plateaus as higher order coefficients of \texttt{PPI} take zero values (c.f., Table 3) and do not affect the results. Finally, the plot on the left depicts accuracy across a range of embedding dimensions $d$. \\

\noindent \textbf{Runtime}. Finally, we compared different embedding methods in terms of runtime. Results for all graphs are reported in Fig. \ref{fig:runtime}. All experiments were run on a personal workstation with a quad-core i5 processor, and 16 GB of RAM. For our proposed ASE, we provide a  light-weight yet highly portable implementation~\footnote{https://github.com/DimBer/ASE-project/tree/master/portable} that uses the SVDLIBC library \cite{svdlibc} for sparse SVD. We also developed a more scalable implementation~\footnote{https://github.com/DimBer/ASE-project/tree/master/slepc\textunderscore based} that relies on (and requires installation of) the SLEPc package \cite{slepc}; this scalable version can perform large-scale sparse SVD on multiple processes and distributed memory environments using the message-passing interface (MPI) \cite{mpi}. We used the high-performance implementation for the five larger graphs, and the portable one for the five smaller ones. Evidently, ASE and HOPE that are SVD-based are orders of magnitudes faster than VERSE, Deepwalk, and LINE. The main factor that slows the latter down seems to be the large number of stochastic optimization iterations that these methods must perform to reach accurate embeddings. Nevertheless, it should be noted that sampling based methods enjoy nearly-full parallelization and could thus benefit more from highly   
multi-threaded environments. On the other hand, methods that rely on SVD (and EVD) can greatly benefit from decades of research on how to efficiently perform these decompositions, and a suite of stable and highly optimized software tools.

\section{Conclusions and Future work}
\label{sec:conclusions}
We presented a scalable node embedding framework that is based on factorizing an adaptive node similarity matrix. The model is carefully studied, interpreted, and numerically evaluated using stochastic block models, with an algorithmic scheme proposed for training the model parameters efficiently and without supervision.

The novel framework opens up several interesting future research directions. For instance, one can explore larger families of node similarity metrics that can be learned using the graph. Furthermore, it would be interesting to assess the performance of different randomized edge sampling methods, and generalize the notion of adaptive-similarity to heterogeneous and multi-layered graph embedding, as well as to edge embedding.

	\vspace{-1cm}

\begin{IEEEbiography}
	[{\includegraphics[width=1in,height=1.25in,clip,keepaspectratio]{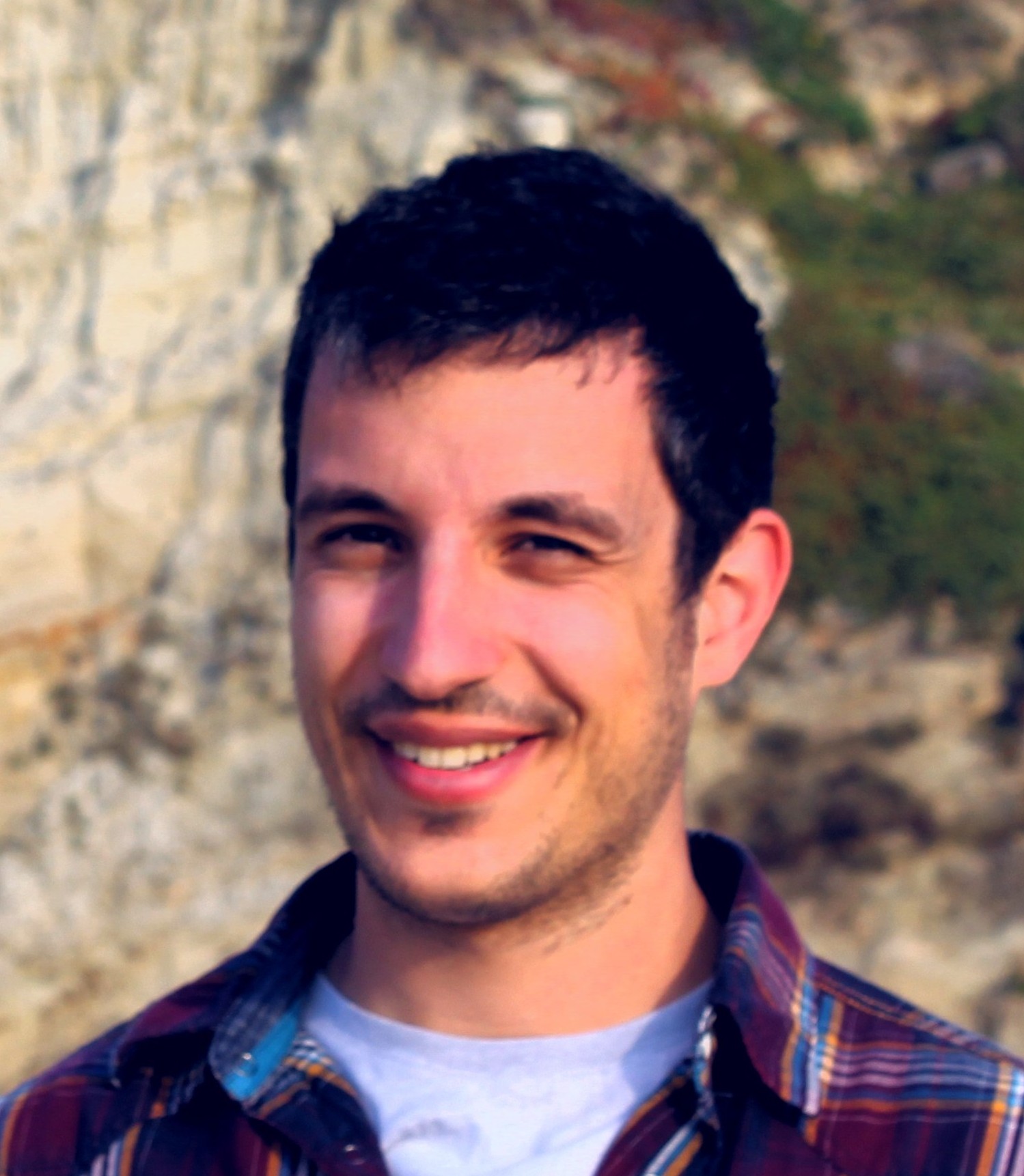}}]{Dimitris Berberidis
		(S'15)} received the Diploma in
	electrical and computer engineering (ECE) from the
	University of Patras, Patras, Greece, in 2012; and the
	M.Sc. as well as Ph.D. degrees in ECE from the University of Minnesota,
	Minneapolis, MN, USA. His research interests
	lie in the areas of statistical signal processing, focusing on sketching and tracking of large-scale processes, and in machine learning, focusing on the developpement of algorithms for scalable learning over graphs, including semi-supervised classification, and node embedding.  
\end{IEEEbiography}

\vspace{-1cm}

\begin{IEEEbiography}
	[{\includegraphics[width=1in,height=1.25in,clip,keepaspectratio]{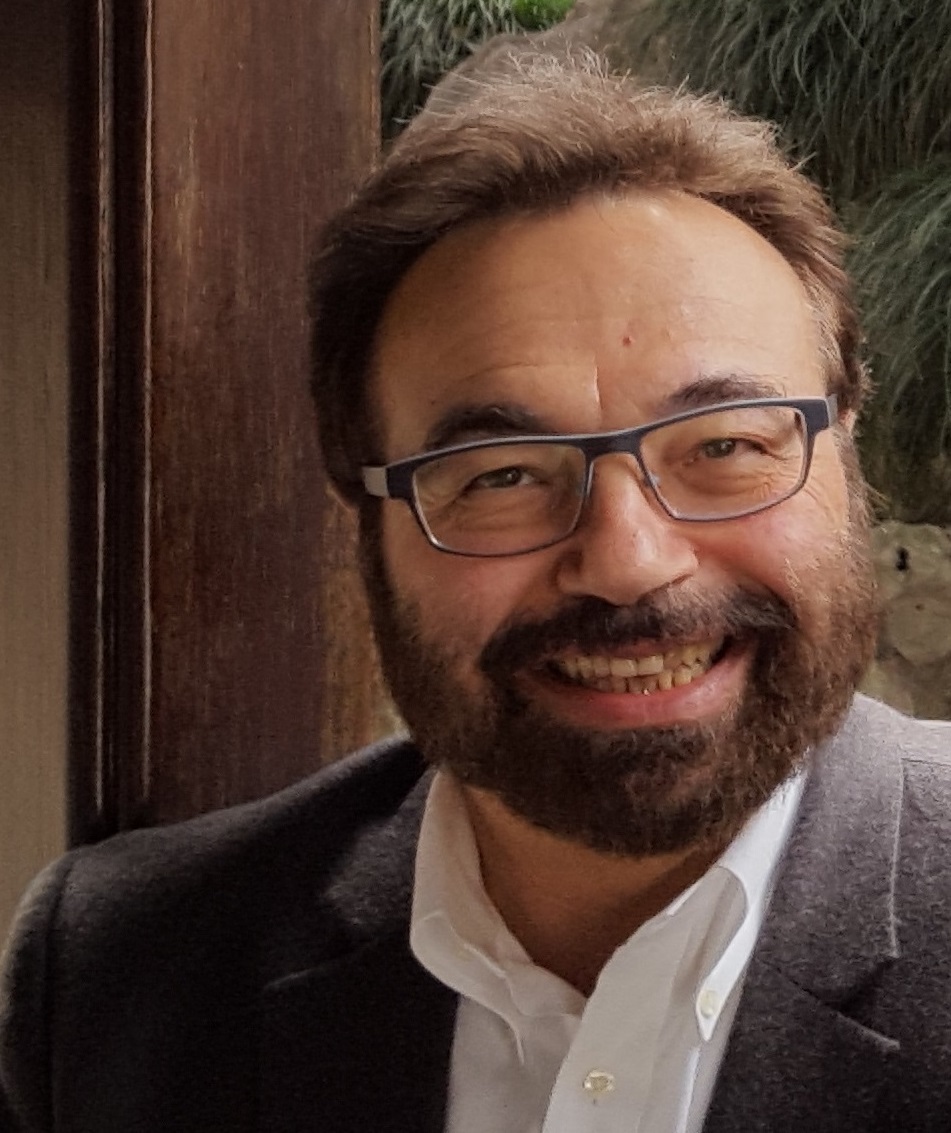}}]
	{Georgios B. Giannakis
		(Fellow'97)} received his Diploma in Electrical 
	Engr. from the Ntl. Tech. Univ. of Athens, Greece, 1981. From 
	1982 to 1986 he was with the Univ. of Southern California (USC), 
	where he received his MSc. in Electrical Engineering, 1983, MSc. 
	in Mathematics, 1986, and Ph.D. in Electrical Engr., 1986. He 
	was with the University of Virginia from 1987 to 1998, and since 
	1999 he has been a professor with the Univ. of Minnesota, where 
	he holds an Endowed Chair in Wireless Telecommunications, a 
	University of Minnesota McKnight Presidential Chair in ECE, 
	and serves as director of the Digital Technology Center. 
	
	His general interests span the areas of communications, networking 
	and statistical learning - subjects on which he has published 
	more than 450 journal papers, 750 conference papers, 25 book 
	chapters, two edited books and two research monographs (h-index 142). 
	Current research focuses on learning from Big Data, wireless 
	cognitive radios, and network science with applications to 
	social, brain, and power networks with renewables. He is the 
	(co-) inventor of 32 patents issued, and the (co-) recipient of 
	9 best journal paper awards from the IEEE Signal Processing (SP) and 
	Communications Societies, including the G. Marconi Prize Paper Award 
	in Wireless Communications. He also received Technical Achievement 
	Awards from the SP Society (2000), from EURASIP (2005), a Young 
	Faculty Teaching Award, the G. W. Taylor Award for Distinguished 
	Research from the University of Minnesota, and the IEEE Fourier 
	Technical Field Award (inaugural recipient in 2015). He is a 
	Fellow of EURASIP, and has served the IEEE in a number of posts, 
	including that of a Distinguished Lecturer for the IEEE-SP Society. 
\end{IEEEbiography}

\end{document}